\documentclass[11pt]{article}
\usepackage[letterpaper, left = 1in, top = 0.9in, right = 1in, bottom = 0.7in, nohead, includefoot, verbose, ignoremp]{geometry}
\pdfoutput=1
\usepackage{float}
\usepackage{amsfonts}
\usepackage{bbm}
\usepackage{amsmath}
\usepackage{graphicx}
\usepackage{natbib}
\usepackage{xy}\xyoption{all} \xyoption{poly} \xyoption{knot}
\usepackage{verbatim}
\usepackage{authblk}
\usepackage[colorlinks = true,
            linkcolor = blue,
            urlcolor  = blue,
            citecolor = blue,
            anchorcolor = blue]{hyperref}

\usepackage[labelformat=simple]{subcaption}

\newcommand{\indep}{\rotatebox[origin=c]{90}{$\models$}}
\newcommand{\sech}{\mbox{sech}}

\author{Chris Glynn} 
\author{Surya T. Tokdar}
\author{David L. Banks}
\affil{Statistical Science, Duke University}
\author{Brian Howard}
\affil{Sciome, LLC}

\title{Bayesian Analysis of Dynamic Linear Topic Models}
\date{\today}

\begin{document}
\maketitle

\begin{abstract}
In dynamic topic modeling, the proportional contribution of a topic to a document depends on the temporal dynamics of that topic's overall prevalence in the corpus.  We extend the Dynamic Topic Model of \citet{blei2006dynamic} by explicitly modeling document-level topic proportions with covariates and dynamic structure that includes polynomial trends and periodicity.   A Markov Chain Monte Carlo (MCMC) algorithm that utilizes Polya-Gamma data augmentation is developed for posterior inference.  Conditional independencies in the model and sampling are made explicit, and our MCMC algorithm is parallelized where possible to allow for inference in large corpora.  To address computational bottlenecks associated with Polya-Gamma sampling, we appeal to the Central Limit Theorem to develop a Gaussian approximation to the Polya-Gamma random variable.  This approximation is fast and reliable for parameter values relevant in the text-mining domain.  Our model and inference algorithm are validated with multiple simulation examples, and we consider the application of modeling trends in PubMed abstracts.  We demonstrate that sharing information across documents is critical for accurately estimating document-specific topic proportions.   We also show that explicitly modeling polynomial and periodic behavior improves our ability to predict topic prevalence at future time points.   
\end{abstract}

\section{Introduction}

Text data is ubiquitous. Newspapers, blogs, emails, tweets, and countless other expressions of written language are central to daily communication. These various forms of text documents both disseminate and preserve information, ideas, and creative expression.  In many cases, the time at which a document is created is an important piece of metadata.  Collections of documents, called corpora, exhibit themes which vary with time.  Modeling the dynamics in a corpus is critical when knowledge accumulates sequentially, such as in bodies of academic literature and news articles.

In this paper, we make four contributions to the dynamic topic modeling literature.  The first contribution is a mathematically principled framework for modeling complex dynamic behavior in corpora.  Existing dynamic topic models \citep{blei2006dynamic, wang2008continuous} are unable to explicitly model polynomial time trends or periodicity in the marginal probabilities of topics in the corpus as a whole.  We develop a model to explicitly account for periodicity and polynomial growth.    

The second contribution is an MCMC algorithm which allows us to quantify uncertainty in the exact posterior distribution of both topics themselves and document-specific topic proportions.  The MCMC algorithm and model also allow us to share information across documents.  Inference for the Dynamic Topic Model (DTM) of \citet{blei2006dynamic} relies on a variational approximation to the posterior.  Because the variational approximation endows each document with its own independent Dirichlet distribution for topic proportions, there is no mechanism for borrowing information across documents.  Furthermore, it is not possible to infer dynamic trends in topic proportions globally in the corpus.  Inferring the time-varying trends in topics globally is an important aspect of the MCMC and model we develop.  We find that borrowing information across documents is critical in order to accurately estimate the document-specific topic proportions and the global topic trends.  

The third contribution is a framework for assessing MCMC convergence in dynamic topic models.  We adapt ideas from \citet{gelman1992} to consider the within chain and across chain variability in total variation distance between topics.  These convergence diagnostics ensure that our estimated topics and document-specific topic proportions are reproducible.  

The fourth contribution is the foundation of a model-based method for choosing the number of topics in a corpus.  Choosing the number of topics in a topic model is an open problem.  We demonstrate that the time-varying marginal probabilities of topics, when paired with the uncertainty in the posterior distributions of topics themselves, can be useful for choosing the number of topics necessary to model the data.
  
\begin{figure}[h!]
\centering
\caption{Left: PubMed trends for Polio and Autism as generated by \citet{Medline}.  Middle and Right: Google search trends for the terms Tweet and Christmas \citep{GoogleTrend} }
\begin{subfigure}{.3\textwidth}
  \centering
  \includegraphics[width=1\textwidth]{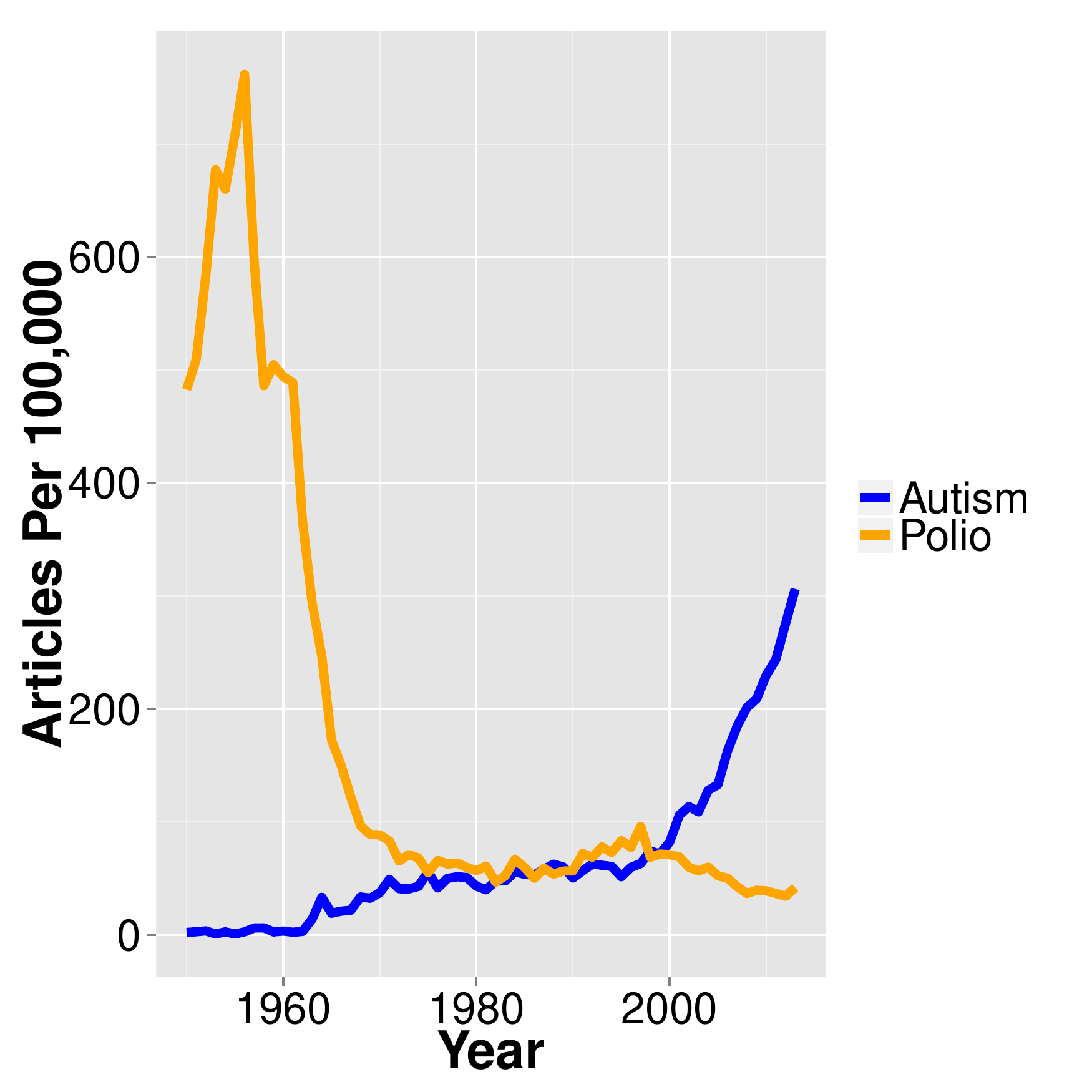}
  \caption{PubMed trends}
  \label{subfig:ML_Trends}
\end{subfigure}%
\begin{subfigure}{.3\textwidth}
  \centering
  \includegraphics[width=1\textwidth]{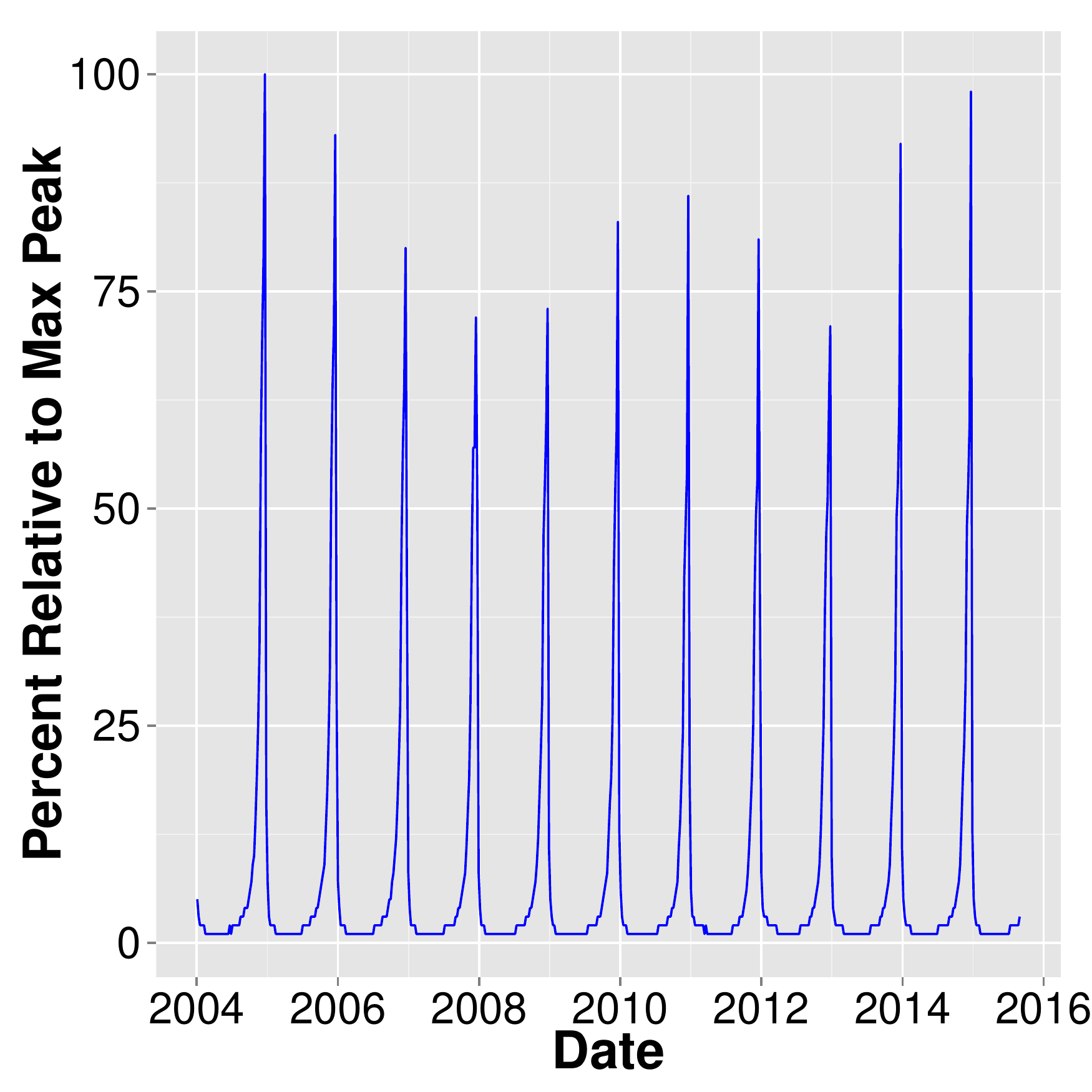}
  \caption{Christmas}
  \label{subfig:Christmas_Trend}
\end{subfigure}
\begin{subfigure}{.3\textwidth}
  \centering
  \includegraphics[width=1\textwidth]{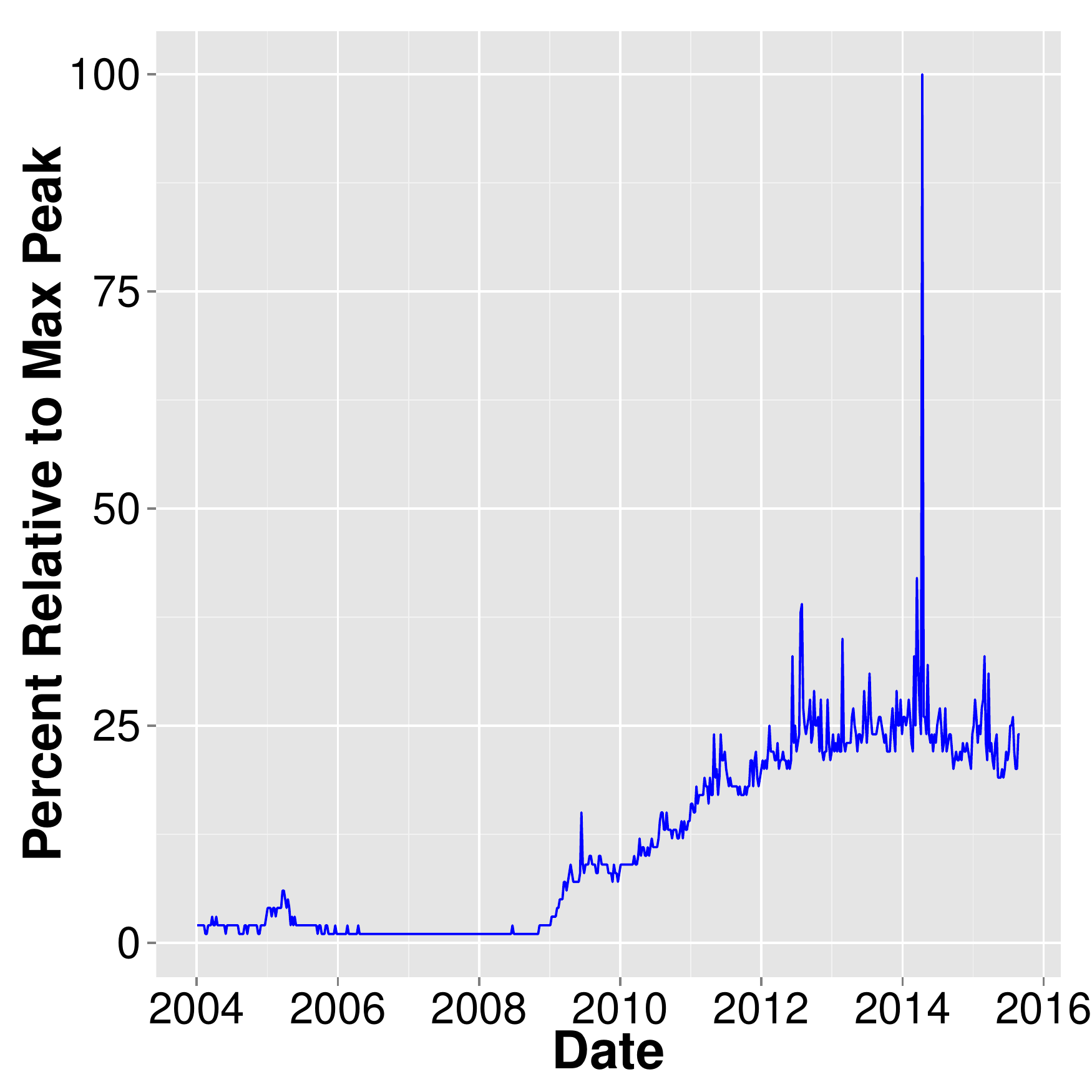}
  \caption{Tweet}
  \label{subfig:Tweet_Trend}
\end{subfigure}
\label{fig:Trends}
\end{figure}

Our modeling and computational framework is motivated by observed dynamic features in corpora ranging from PubMed abstracts to Google searches.  Each corpus presents its own modeling challenge.  The PubMed corpus, a collection of 25 million abstracts in biomedical and life sciences, calls for a model which allows the marginal probability of topics to rapidly grow and decay. In collections of academic articles, open problems are solved and new questions emerge.  As an example, many PubMed articles in the early 1950s focused on the disease Polio.  After the vaccine for Polio was discovered in 1952, the scientific interest in the topic declined.  Today, many researchers are focused on understanding the causes of Autism.  Figure \ref{subfig:ML_Trends} illustrates this changing interest in Polio and Autism research in the PubMed database.        

While some semantic themes rapidly rise or fall in their importance, others naturally re-occur.  Figure \ref{subfig:Christmas_Trend} shows the periodicity associated with Google searches for the term Christmas \citep{GoogleTrend}.  Periodic interest in a topic is natural for semantic themes such as holidays, sporting events, weather, and many other calendar driven events.  

The way and frequency with which words are used can also change.  Prior to 2006, the word tweet primarily referred to the vocal call of a bird.  In 2015, it is likely that a tweet refers to a 140-character message published on the website Twitter.  Because of changes in social norms, communication, and technology, tweet is more important in the cultural lexicon than it was a decade ago.  As a result, it is used more frequently.  This is observed in Figure \ref{subfig:Tweet_Trend}.  The rapid emergence and decay of a topic's prevalence, themes which re-occur, and the evolution of word use all motivate the need for a modeling and computational framework which is capable of accomodating significant structural changes in a corpus over time.

Probabilistic models for text documents have been widely written about in the Machine Learning literature.  Topic models, as they are often called, are essentially hierarchical Bayesian statistical models with a multinomial likelihood.  While there are several varieties of topic models, the primary differences amongst them stem from how the probabilities of the multinomial likelihood and the document topic proportions are modeled.    

In the foundational work of \citet{blei2003}, the multinomial likelihood probabilities and the document topic proportions are modeled with Dirichlet distributions -- leading to the name Latent Dirichlet Allocation (LDA).  A fundamental assumption of LDA is that documents are exhangeable in time; however, exchangeability in time is not an appropriate assumption for corpora where information acccumulates sequentially, such as in news articles and bodies of academic literature.  In such corpora, the order in which documents are written clearly matters. 

Another popular choice for modeling these probability simplices is to place a Gaussian distribution on the natural parameters of the multinomial distribution.   One advantage of the logistic-Normal  model for the probability simplex is that it can be extended to one that evolves over time.  The Dynamic Topic Model (DTM) of \citet{blei2006dynamic} models the evolution of the probability distribution for both individual vocabulary terms and topics themselves.  In the DTM, the natural parameters for both the multinomial likelihood and the document topic proportions are modeled with random walk state-space models.   

There are other approaches to modeling temporal dynamics in text documents.  The continuous-time DTM (cDTM) of \citet{wang2008continuous} is a continuous version of the DTM which models the natural parameters with Brownian motion.  This allows for more granular time discretization and eases the computation associated with finer time scales.  In the Topics Over Time model of \citet{wang2006topics}, the topics themselves are static; however, time-stamps on the documents are used to enhance learning of these static topics. 

In this paper, we extend the DTM to allow for more complex dynamic behavior and correlation in document topic proportions.  Rather than posit a random walk state-space model for the topic probability in a specific document, we model the topic proportions with a Dynamic Linear Model \citep{westharrison}.  We call this extension the Dynamic Linear Topic Model (DLTM).  The DLTM allows the marginal probability of topics to exhibit periodicity, locally linear trends, and a rich set of other dynamic behavior.  

In addition, the DLTM offers a natural framework for incorporating document-specific covariates such as author or publisher.  Some documents are inherently similar, and covariates encode this similarity.  By including them in the model, we induce correlation among documents which share covariates.  While \citet{rosenzvi2004author} consider an author-topic model and \citet{mimno2008gibbs} introduce a Markov Random Field prior to induce correlation amongst related documents, we are not aware of a topic model that is both dynamic and which induces correlation amongst documents through the inclusion of covariates.  

It is worth noting that the DTM is a special case of the DLTM.  The distinction between DTM and DLTM is the way in which we model the dynamics of marginal topic probabilities.  The dynamics for word frequency within a specific topic are preserved from the DTM.

The DLTM model class requires intensive computation for inference and prediction. \citet{chen_nips_2013} have demonstrated the utility of a Gibbs sampling algorithm with Polya-Gamma data augmentation \citep{PSW2013} for a static logistic-Normal topic model.  \citet{Windle2013} bring Polya-Gamma data augmentation to dynamic models for count data.  We demonstrate that Polya-Gamma data augmentation is a method that allows for fully Bayesian posterior inference in the DLTM. 

One challenge of the Polya-Gamma data augmentation scheme is that sampling such random variates can be slow for parameter values pertinent to text analysis.  We design a fast, approximate Polya-Gamma sampler that dramatically reduces the amount of time required for sampling in such cases and makes posterior inference feasible for large collections of documents.  Without this approximate sampler, Polya-Gamma data augmentation is infeasible in this application due to the high computational cost of sampling Polya-Gamma random variates.  

While our Gibbs sampling algorithm is an admittedly slower method of inference than the variational Kalman filter of \citet{blei2006dynamic}, our goal is not to compete with existing methods on speed.  Rather, our aim is to develop an inference algorithm for an extended model class and to gain a fundamental understanding of the uncertainty inherent in high-dimensional probability models for categorical data.

The remainder of the paper is structured as follows:  Section \ref{sec:Model} describes the model; Section \ref{sec:Prior} constructs and examines the implications of prior distributions for topics and topic proportions; Section \ref{sec:Inference} details the Markov Chain Monte Carlo algorithm for posterior sampling; Section \ref{sec:PG_Approx} develops an approximate sampling algorithm for Polya-Gamma random variates; Section \ref{sec:SimStudy} examines the performance of our computational algorithm on a synthetic data set; Section \ref{sec:CaseStudy} demonstrates how to model locally linear trends in the document topic proportions for a sample of abstracts from PubMed; Section \ref{sec:Conclusion} concludes.

\section{Model}
\label{sec:Model}
In our analysis, we suppose that there are $K$ topics in the corpus and that there is a known vocabulary of length $V$ which neither expands nor contracts with time.  Each element of the vocabulary is a term, and these terms are indexed by $v \in \{1, \ldots, V\}$.  At each time point, $t\in \{1, \ldots, T\}$, the corpus contains $D_t$ documents with $d \in \{1, \ldots, D_t\}$ indexing the documents.  

A document itself, $W_{d,t}$, is a vector where each entry in $W_{d,t}$ corresponds to a word in the document.  The entries of document $W_{d,t}$ are denoted by $w_{n,d,t}$, which corresponds to the $n^{th}$ word in the $d^{th}$ document at time $t$.  Each document has its own length of $N_{d,t}$ words, and the words within document $W_{d,t}$ are exchangeable (i.e. the index set $n\in\{1, \ldots, N_{d,t} \}$ can be permuted freely).    

Each document is observed at a single time point.  Documents themselves do not evolve over time.  Only topics and the global topic proportions evolve in time.  Despite the static nature of a document, we index documents with $t$ to make explicit the membership of each document in a specific time-slice.  

The latent topic associated with the $n^{th}$ word in the $d^{th}$ document at time $t$ is denoted by $z_{n,d,t}$.  Conditional on a latent topic variable, $z_{n,d,t}$, a word in the document, $w_{n,d,t}$, is sampled from a multinomial distribution over the vocabulary.

\begin{align*}
Pr(w_{n,d,t} = v | z_{n,d,t} = k) &= \frac{ e^{ \beta_{k,v,t} } }{\sum_{j = 1}^V e^{\beta_{k,j,t}} }
\end{align*}

The $\beta_{k,v,t}$ parameter is the natural parameter associated with the $v^{th}$ term of the $k^{th}$ topic.  Formally, this $k^{th}$ topic is a probability distribution over the $V$ terms in the vocabulary at time $t$. For the purpose of identifiability, we fix $\beta_{k,V,t} = 0$ for each topic $k$ and time $t$.

Following \citet{blei2006dynamic}, the evolution in time of the natural parameter $\beta_{k,v,t}$ is modeled with a random walk.  

\begin{align*}
\beta_{k,t} &= \beta_{k,t-1} + \nu_{k,t}, \hspace{1cm} \nu_{k,t} \sim N_{V}(0, \sigma^2 I) \\
\beta_{k,0} &\sim N(m_{k,0}, \sigma^2_{k,0} )
\end{align*}
The error terms $\nu_{k,t}$ are mutually independent: $\nu_{k,t} \indep \nu_{k,t'}$ for $t \neq t'$, and $\nu_{k,t} \indep \nu_{k',t}$ for $k \neq k'$.  

The word-specific latent topic variable, $z_{n,d,t}$, is sampled from its own multinomial distribution conditional on the set of natural parameters $\eta_{d,\cdot,t}$.  \begin{align*}
Pr(z_{n,d,t} = k | \eta_{d,\cdot,t} ) &= \frac{e^{\eta_{d,k,t} } }{\sum_{j=1}^K e^{\eta_{d,j,t} } }.  
\end{align*}

Throughout the paper, when an index is omitted and replaced with $\cdot$, this notation signifies the collection of all elements of the omitted index. As an example: $\eta_{d,\cdot, t} = \{\eta_{d,1,t}, \eta_{d,2,t}, \ldots, \eta_{d,K,t} \}$.   For the purpose of identifiability, we fix $\eta_{d,K,t} = 0$.

Thus far, the model described is identical to that of the Dynamic Topic Model.  Where our model deviates from the DTM is in how we model $\eta_{d,k,t}$.  For each $k \in \{1,\ldots, K \}$, we model the vector $\eta_{\cdot, k,t} = \{ \eta_{1,k,t}, \eta_{2,k,t}, \ldots, \eta_{D_t,k,t} \}$ with a Dynamic Linear Model \citep{westharrison}.  It is with this DLM that we incorporate periodic and polynomial behavior, covariates, and more broadly, an extensive set of features for temporal dependence in topic proportions:  

\begin{align*}
\eta_{\cdot, k,t} &= F_{k,t} \alpha_{k,t} + \epsilon_{k,t} \hspace{1cm} \epsilon_{k,t} \sim N_{D_t}(0, a^2 I_{D_t}) \\
\alpha_{k,t} &= G_{k,t} \alpha_{k,t-1} + \xi_{k,t}, \hspace{1cm} \xi_{k,t} \sim N_p(0,\delta^2 I_p ) \\
\alpha_{k,0} &\sim N(m_{k,0}, C_{k,0}).
\end{align*}

The error terms are mutually independent: $\epsilon_{k,t} \indep \epsilon_{k',t}$ for $k \neq k'$ and $\epsilon_{k,t} \indep \epsilon_{k,t'}$ for $t \neq t'$.  This independence statement implies the $K$ distinct DLMs are mutually independent as well.  The integer constant $p$ is the dimension of the underlying state-vector, $\alpha_{k,t}$.  The $F_{k,t} = \left( F_{1,k,t}', \ldots, F_{D_t, k, t}' \right)'$ is a known $D_t \times p$ time-varying design-matrix of document covariates and model component terms corresponding to seasonality, trend, etc.  The $G_{k,t}$ is a known $p\times p$  system matrix.  We previously noted that the DTM is a special case of the DLTM.  The DTM can be recovered by fixing $p=1$ and $F_{k,t} = G_{k,t} = 1$.

\begin{figure}[h!]
\caption{This is the graphical model representation of DLTM.  Conditional independencies are made explicit in this representation.  These conditional independencies will be important for parallel sampling in posterior inference.}
\label{fig:Graph}
$$\vcenter{\xymatrix{
\beta_{K,\cdot,1} \ar[r] \ar@/^-3pc/[ddd]  & \beta_{K,\cdot,2} \ar[r] \ar@/^-3pc/[ddd]  & \ar[r]  \cdots \ar[r] &  \beta_{K,\cdot,t-1} \ar[r] \ar@/^-3pc/[ddd]  &  \beta_{K,\cdot,t} \ar@/^-3pc/[ddd]  \\
\vdots &  \vdots & \vdots & \vdots & \vdots \\
\beta_{1,\cdot,1} \ar[r] \ar[d] & \beta_{1,\cdot,2} \ar[r] \ar[d] & \ar[r]  \cdots \ar[r] &  \beta_{1,\cdot,t-1} \ar[r] \ar[d] &  \beta_{1,\cdot,t} \ar[d] \\
W_{\cdot,1}  & W_{\cdot,2} & \cdots & W_{\cdot,t-1} & W_{\cdot,t} \\
Z_{\cdot,1} \ar[u] & Z_{\cdot,2} \ar[u] & \cdots & Z_{\cdot,t-1} \ar[u] & Z_{\cdot,t} \ar[u] \\
\eta_{\cdot,1,1} \ar@/^0pc/[u]  & \eta_{\cdot,1,2} \ar@/^0pc/[u]  & \cdots & \eta_{\cdot,1,t-1}\ar@/^0pc/[u]  &\eta_{\cdot,1,t} \ar@/^0pc/[u]  \\
\alpha_{1,1} \ar[u]\ar[r] & \alpha_{1,2} \ar[u]\ar[r]  &\ar[r] \cdots \ar[r] & \alpha_{1,t-1} \ar[u]\ar[r] &\alpha_{1,t} \ar[u] \\
\vdots &  \vdots & \vdots & \vdots & \vdots \\
\eta_{\cdot,K-1,1} \ar@/^4pc/[uuuu] & \eta_{\cdot,K-1,2} \ar@/^4pc/[uuuu] & \cdots & \eta_{\cdot,K-1,t-1} \ar@/^4pc/[uuuu] & \eta_{\cdot,K-1,t} \ar@/^4pc/[uuuu]\\
\alpha_{K-1,1} \ar[u]\ar[r] & \alpha_{K-1,2} \ar[u]\ar[r]  &\ar[r] \cdots \ar[r] & \alpha_{K-1,t-1} \ar[u]\ar[r] &\alpha_{K-1,t} \ar[u]
} } $$
\end{figure}
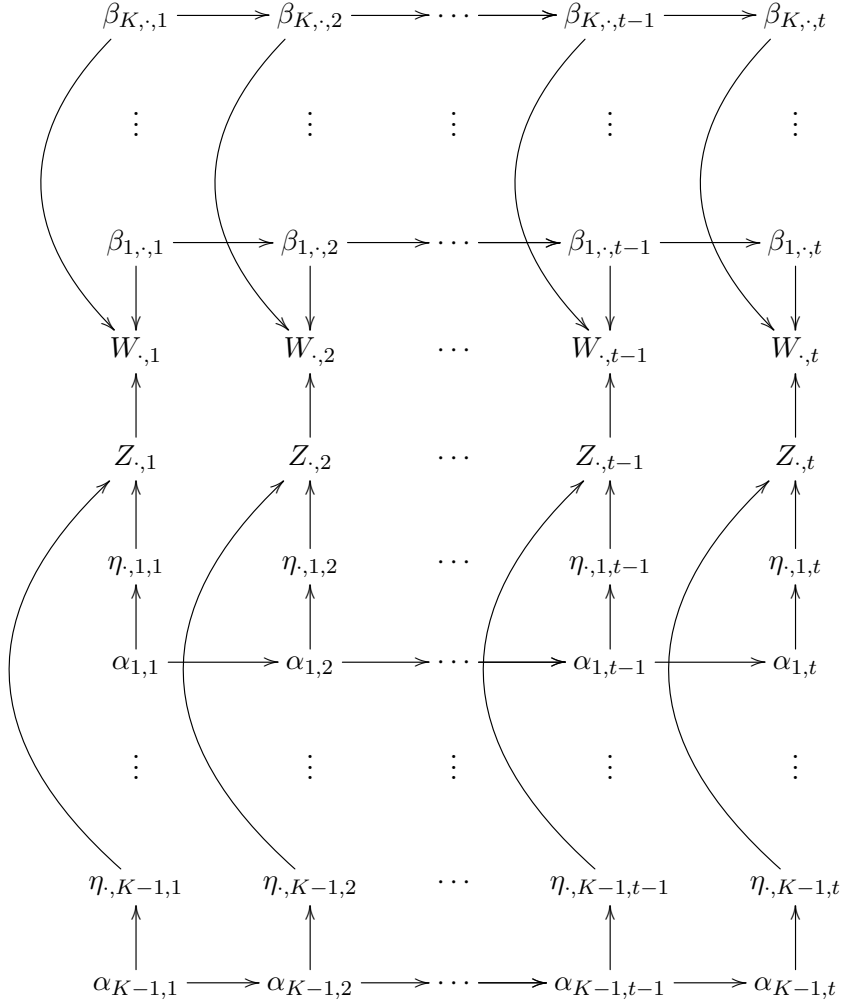

\subsection{Likelihood}
\label{sec:Likelihood}

The likelihood of an entire corpus can be computed by taking advantage of the conditional independencies encoded in the graphical representation of the DLTM, as presented in Figure \ref{fig:Graph}. For succinct notation, we let $W_{\cdot,t} = \{ W_{1,t}, W_{2,t}, \ldots, W_{D_t, t} \}$ and $W_{\cdot,1:T} = \{ W_{\cdot,1}, \ldots, W_{\cdot,T} \}$.

The formula for the likelihood is
\begin{align*}
&p(W_{\cdot,1:T} | Z_{\cdot,1:T}, \alpha_{\cdot,1:T}, \beta_{\cdot,\cdot,1:T} ) = \prod_{t=1}^T p(W_{\cdot,t} | Z_{\cdot,t}, \alpha_{\cdot,t}, \beta_{\cdot,\cdot,t})\\
&=\prod_{t=1}^T \prod_{d=1}^{D_t} p(W_{d,t} | Z_{d,t}, \alpha_{\cdot,t}, \beta_{\cdot,\cdot,t}) = \prod_{t=1}^T \prod_{d=1}^{D_t} \prod_{n=1}^{N_{d,t}} p(w_{n,d,t} | z_{n,d,t}, \alpha_{\cdot,t}, \beta_{\cdot,\cdot,t}) \\
&\propto \prod_{t=1}^T \prod_{d=1}^{D_t} \prod_{n=1}^{N_{d,t}} \left( \frac{e^{\beta_{z_{n,d,t},1,t} } }{\sum_{j=1}^{V} e^{\beta_{z_{n,d,t},j,t} } }  \right)^{\mathbbm{1}_{\{w_{n,d,t} = 1 \} } } \ldots\left( \frac{e^{\beta_{z_{n,d,t},V,t} } }{\sum_{j=1}^{V} e^{\beta_{z_{n,d,t},j,t} } }  \right)^{\mathbbm{1}_{\{w_{n,d,t} = V \} } }
\end{align*}

It is also useful to examine the likelihood contribution from a specific topic.  The objective is to demonstrate that the multinomial likelihood can be reparameterized to one which is proportional to a binomial likelihood if we condition on a specific topic.  

This proportionality to the binomial likelihood is of interest from a computational perspective.  Recent work on inference for Bayesian logistic models provides a useful data augmentation scheme.  If we can reduce the problem to inference in a logistic model by conditioning, the inference algorithm is straightforward. For a full derivation of the likelihood conditioning and reparameterization strategy, refer to Appendix \ref{App:Likelihood}.

If we condition on $z_{n,d,t} = k$, the conditional likelihood is proproportional to:
\begin{align*}  
\ell(\beta_{k,t} | z_{d,n,t} = k) & \propto \left( \frac{e^{\beta_{k,1,t} } }{\sum_{j=1}^{V} e^{\beta_{k,j,t} } } \right)^{y_{k,1,t} } \ldots \left( \frac{e^{\beta_{k,V,t} } }{\sum_{j=1}^{V} e^{\beta_{k,j,t} } } \right)^{y_{k,V,t} }    
\end{align*}
where $y_{k,v,t} = \sum_{d=1}^{D_t} \sum_{n=1}^{N_{d,t}} \mathbbm{1}_{ \{w_{n,d,t} = v \} } \mathbbm{1}_{\{z_{n,d,t} = k \} }$. Informally, $y_{k,v,t}$ is the number of times vocabulary term $v$ is assigned to topic $k$ across all documents at time $t$.
We reparameterize the likelihood following the strategy of \citet{holmes2006}.

\begin{align*}  
    \ell(\beta_{k,t} | Z_t=k) & \propto \left(\frac{e^{\gamma_{k,v,t} } }{1 + e^{\gamma_{k,v,t} } } \right)^{y_{k,v,t} } \left(\frac{1}{1+e^{\gamma_{k,v,t} } } \right)^{n_{k,t}^y - y_{k,v,t}}
\end{align*}
where $\gamma_{k,v,t} = \beta_{k,v,t} - \log \sum_{j \neq v} e^{\beta_{k,j,t} }$ and $n_{k,t}^y = \sum_{j=1}^V y_{k,j,t}$ is the total number of words assigned to topic $k$ at time $t$.  

Note that the form of the conditional likelihood is now proportional to the binomial likelhood.  This allows us to proceed with a Gibbs sampling algorithm using Polya-Gamma data augmentation as outlined in Section \ref{sec:Inference}.

\section{Prior Distributions}
\label{sec:Prior}
Eliciting priors for $\beta_{k,v,t}$, $\alpha_{k,t}$, and $\eta_{d,k,t}$ is equivalent to placing prior distributions on probability simplices of two different dimensions: the $K$ different $V-1$ dimensional simplices for vocabulary terms, and the $D_t$ different $K-1$ dimensional simplices for document topic proportions at time $t$.  We will consider each in turn.

For each $\beta_{k,v,0}$, we assume a diffuse Gaussian prior: $\beta_{k,v,0} \sim N(0,1)$.  By centering this prior at zero, we do not favor any particular vocabulary term as being a keyword in the topic.  We allow the data to inform us as to which words are keywords.  The innovation variance of the $\beta_{k,v,t}$ process is $\sigma^2 = .01$.  

\begin{figure}[h!]
\centering
\caption{Priors for vocabulary term probabilities.  Left: Prior for $P(w_{n,d,1} =v | z_{n,d,1}=k )$.  Right:  $E[P(w_{n,d,1}=v | z_{n,d,1}=k ) | \beta_{k,v,1}]$ } 
\begin{subfigure}{.4\textwidth}
  \centering
  \includegraphics[width=1\textwidth]{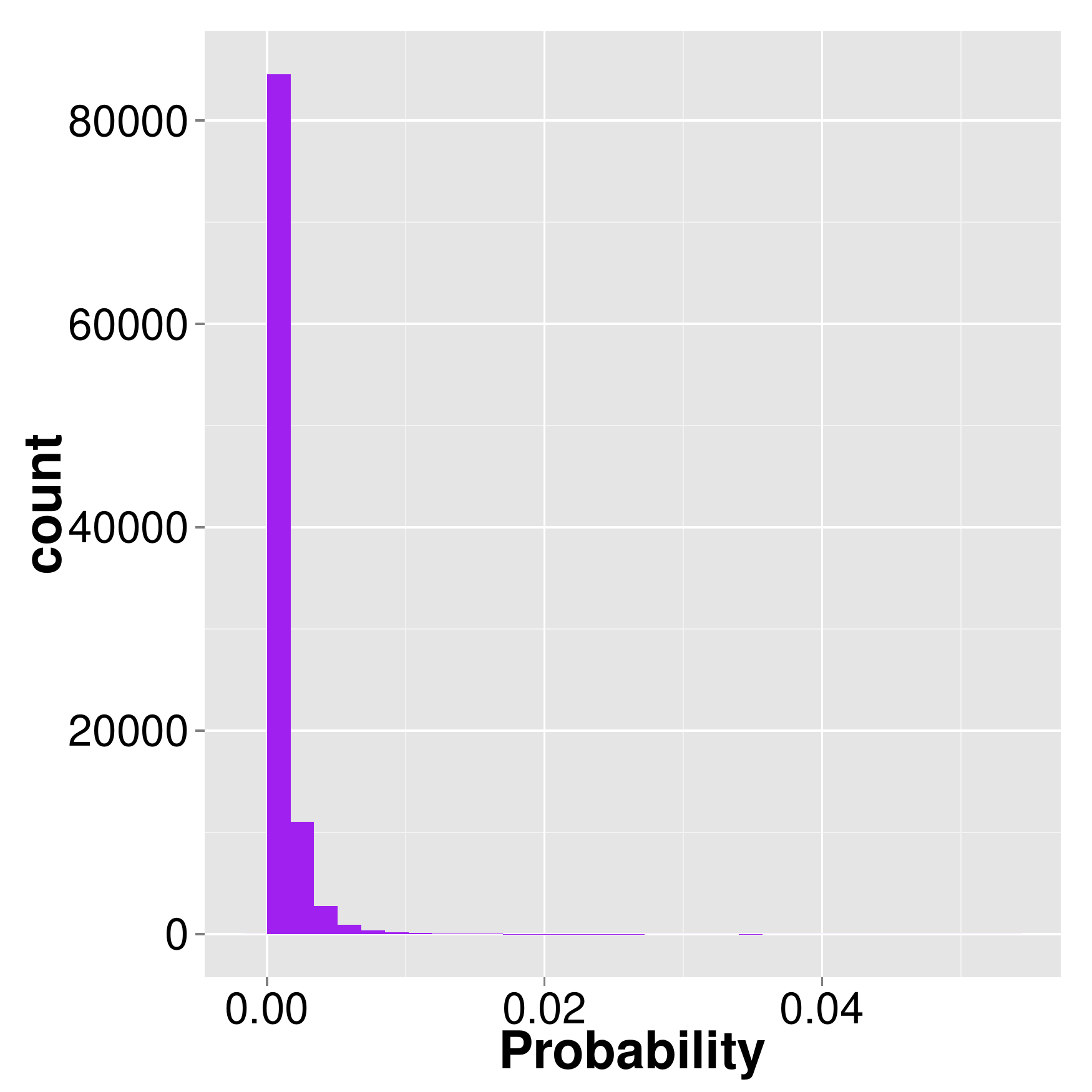}
  \caption{ }
  \label{subfig:Prob_V_Prior}
\end{subfigure}
\begin{subfigure}{.4\textwidth}
  \centering
  \includegraphics[width=1\textwidth]{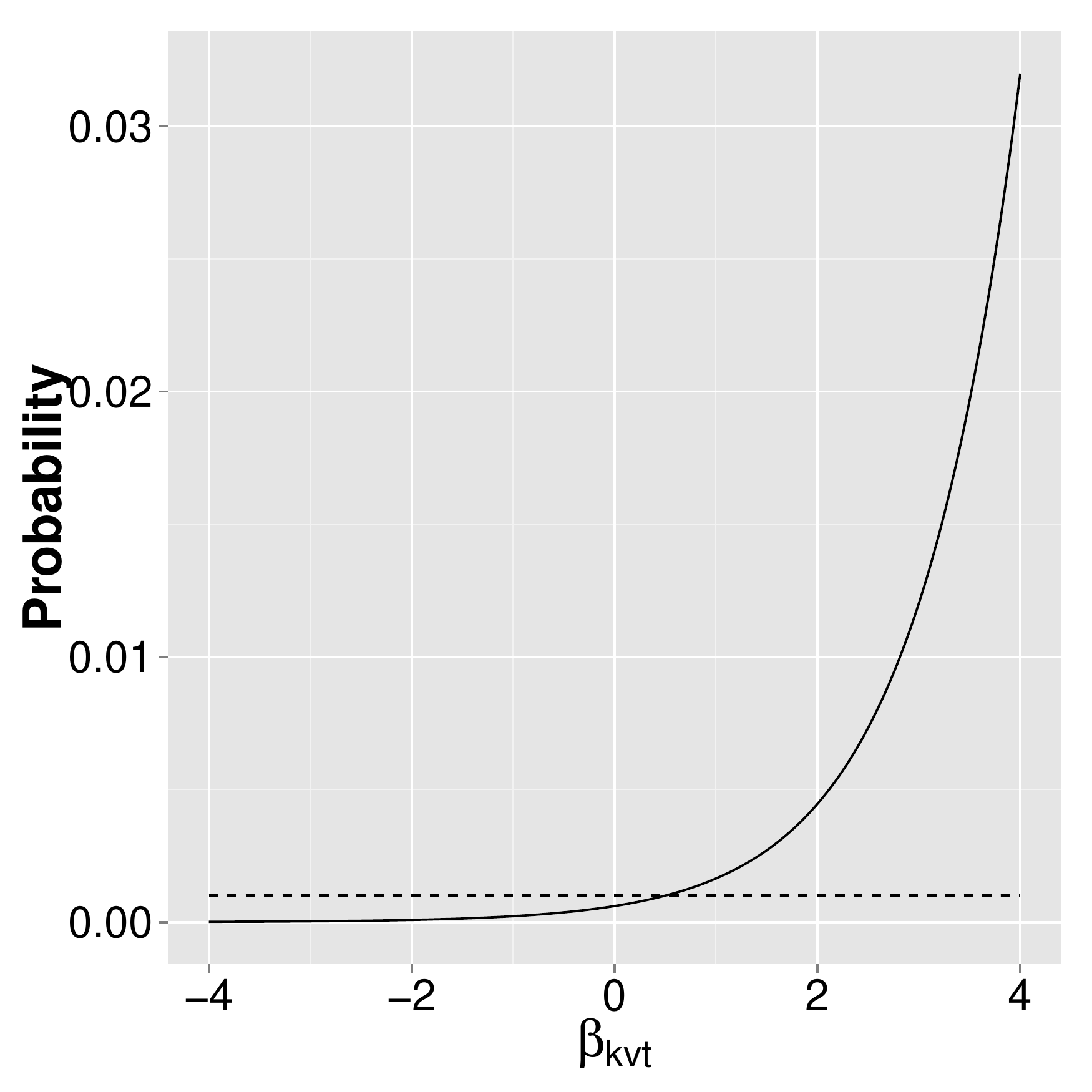}
  \caption{}
  \label{subfig:Exp_Prob_V}
\end{subfigure}
\end{figure}

The uncertainty in $\beta_{k,v,0}$ itself is not enough to determine the uncertainty in the probability distribution for the probability of the $v^{th}$ term.  To fully assess the uncertainty in the prior distribution for the probability of the $v^{th}$ term, it is necessary to consider the prior uncertainty for the remaining $V-1$ terms.  The histogram in Figure \ref{subfig:Prob_V_Prior} shows samples from the prior distribution for the probability of vocabulary term $v$ when there are $V=1000$ terms in the vocabulary.

If we fix the value of $\beta_{k,v,1}$ but preserve the uncertainty in the remaining $V-1$ terms, we can get a sense of how different values of $\beta_{k,v,1}$ impact the probability of the $v^{th}$ term at time $t=1$.  The solid black line in Figure \ref{subfig:Exp_Prob_V} represents the expected value of the prior probability for the $v^{th}$ term conditional on the value of $\beta_{k,v,1}$ on the $x$-axis.  The dashed black line in the same figure represents the naive probability $\frac{1}{V}$.

\begin{figure}[h!]
\centering
\caption{Left: Prior for Topic overlap.  Right: Simulated trajectories for $\alpha_{k,t}$ process} 
\begin{subfigure}{.4\textwidth}
  \centering
  \includegraphics[width=1\textwidth]{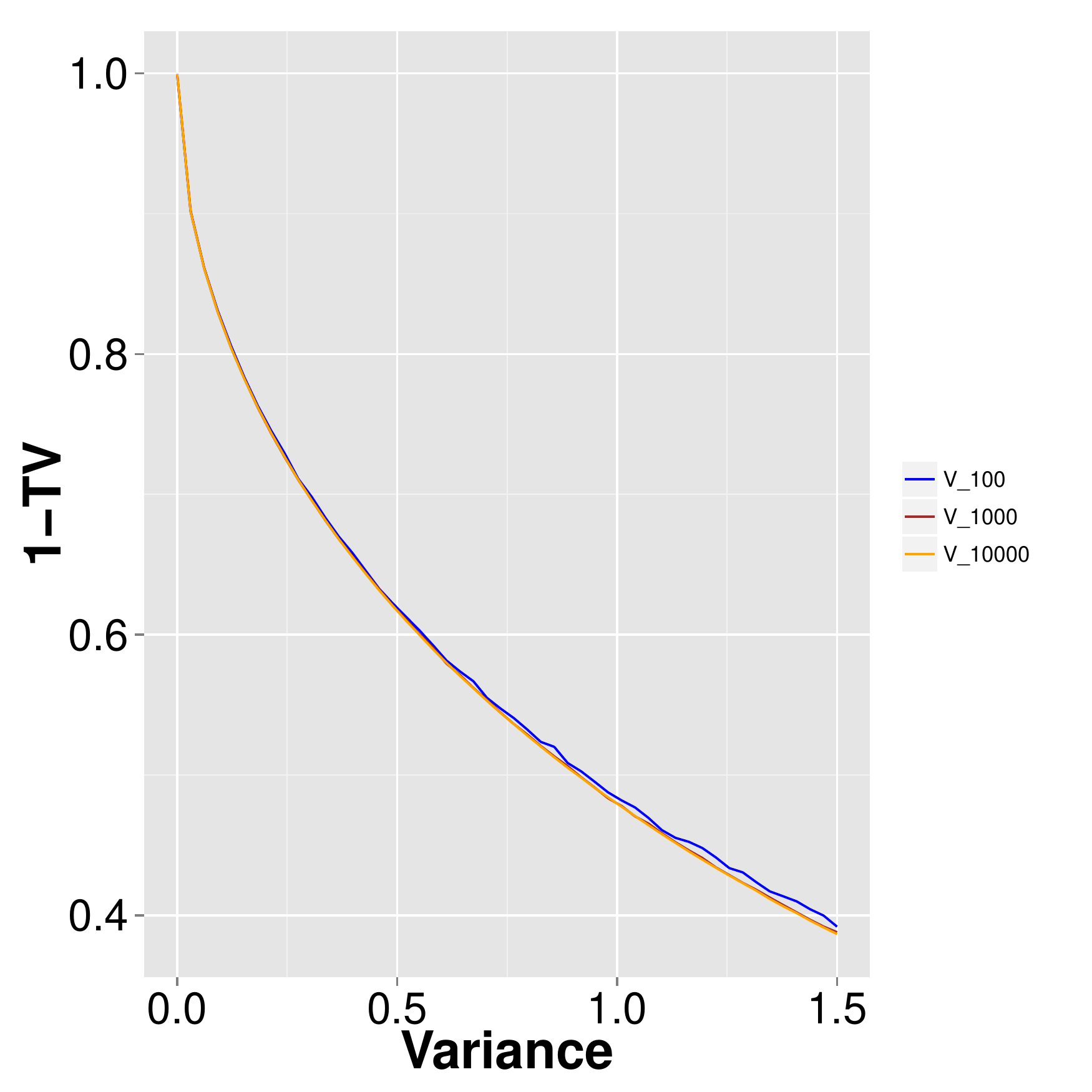}
  \caption{Topic Overlap}
  \label{subfig:Topic_Overlap}
\end{subfigure}%
\begin{subfigure}{.4\textwidth}
  \centering
  \includegraphics[width=1\textwidth]{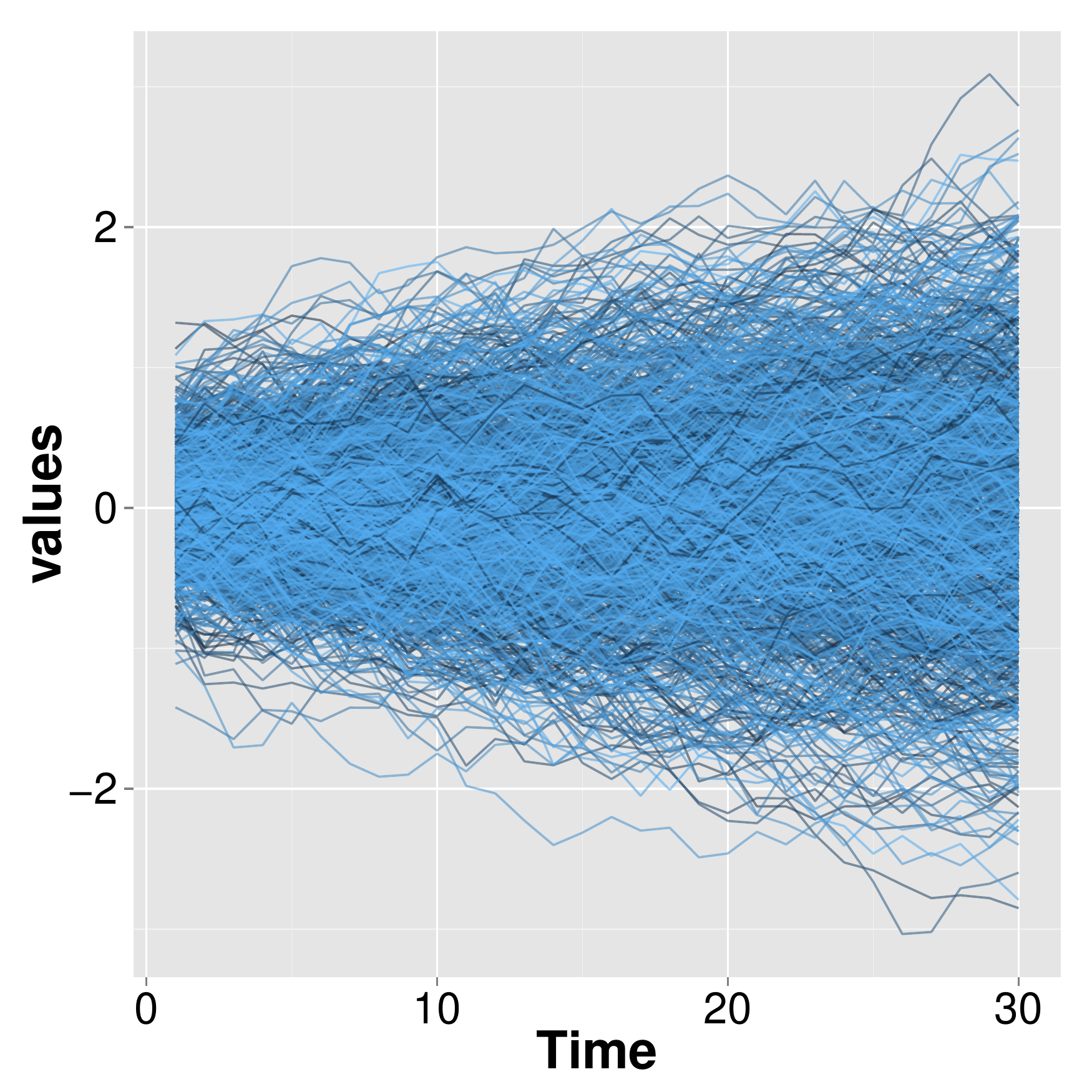}
  \caption{ $\alpha_{k,t}$ trajectories}
  \label{subfig:Alpha_Trajectory}
\end{subfigure}
\end{figure}

While it is important to consider the prior uncertainty of each parameter individually, it is equally important to consider the aggregated uncertainty on the simplex itself.  We examine the uncertainty on the simplex by computing the expected overlap between two topics repeatedly sampled from the prior.  We define the overlap in topics to be the complement of the total variation distance (TV) between the topics: $1-TV(Topic_1,Topic_2)$.  This complement in total variation distance is a good measure of how similar topics sampled from the prior are.  Topics with overlap close to 1 are nearly identical in distribution.  Topics with overlap close to zero are very different in their respective distributions.  Figure \ref{subfig:Topic_Overlap} plots the expected topic overlap for topics with $V=100$, $V=1000$, and $V=10000$ against a range of choices for the variance $\sigma^{2}_{k,v,0}$.  Note that for $\sigma^2_{k,v,0}=1$ the expected topic overlap is about $\frac{1}{2}$.  Our prior belief is that there are $K$ distinct topics in the corpus, and our choice of prior variance is consistent with this belief.  Figure \ref{subfig:Topic_Overlap} also demonstrates that $\beta_{k,v,0} \sim N(0,1)$ is a reasonable prior for a wide range of vocabulary sizes.

In the remainder of the discussion for priors, we fix our attention to the special case when the DTM and the DLTM are equivalent.  We assume that $\alpha_{k,0} \sim N(0,0.1)$ and that the variance for the innovations in the $\alpha_{k,t}$ process is $\delta^2 = 0.025$.  Figure \ref{subfig:Alpha_Trajectory} plots sampled trajectories of $\alpha_{k,t}$ under this prior.  Since the prior for each $\alpha_{k,t}$ is centered at zero, this specification favors the notion that documents are roughly equal in their topic proportions; however, the uncertainty in the $\alpha_{k,t}$ process allows for reasonable deviations from zero which enables the data to inform us that some topics are more (less) prevalent than others.

\begin{figure}[h!]
\centering
\caption{Priors for $\eta_{k,1}$ and topic proportions}
\begin{subfigure}{.3\textwidth}
  \centering
  \includegraphics[width=1\textwidth]{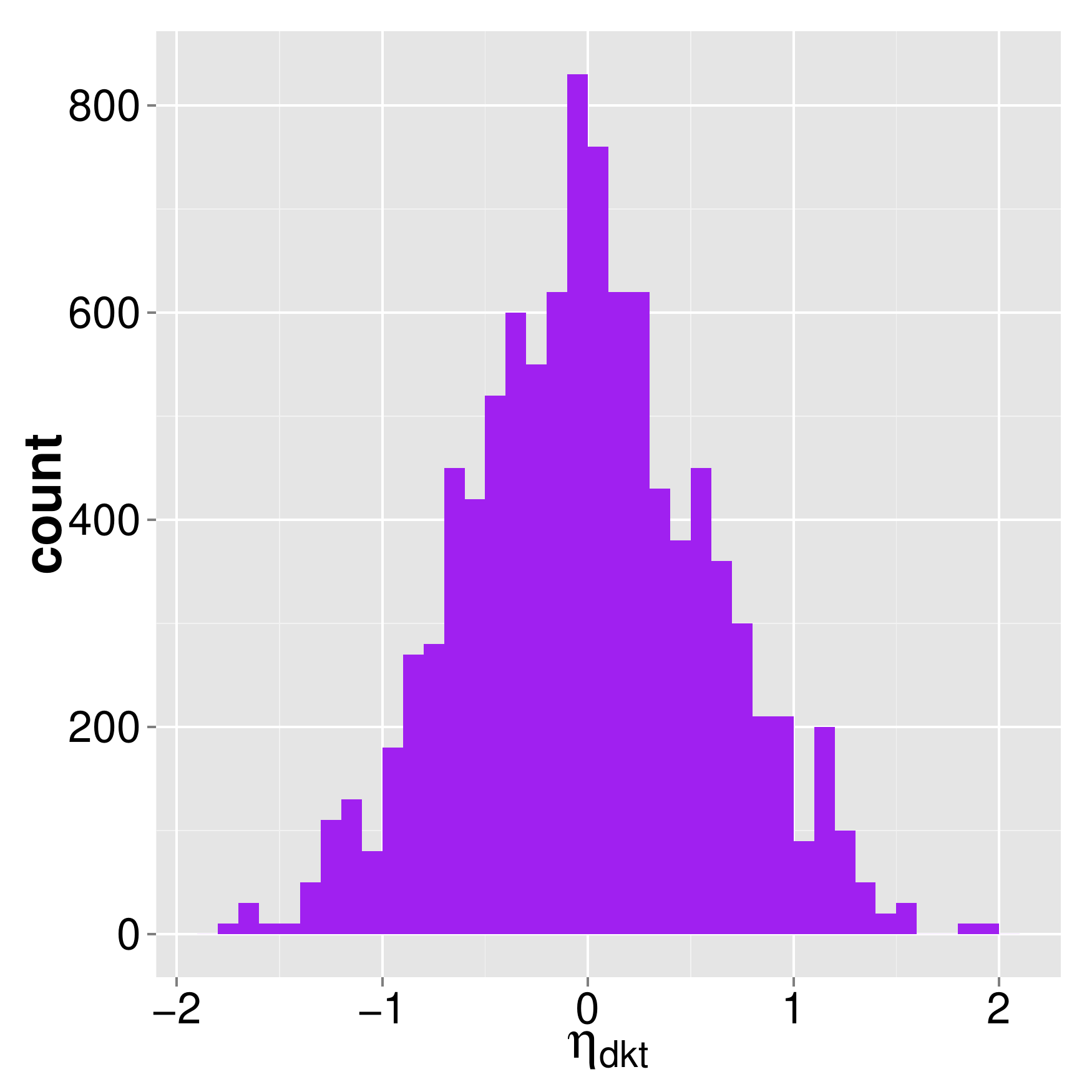}
  \caption{Prior for $\eta_{d,k,1}$}
  \label{subfig:Eta_Hist}
\end{subfigure}%
\begin{subfigure}{.3\textwidth}
  \centering
  \includegraphics[width=1\textwidth]{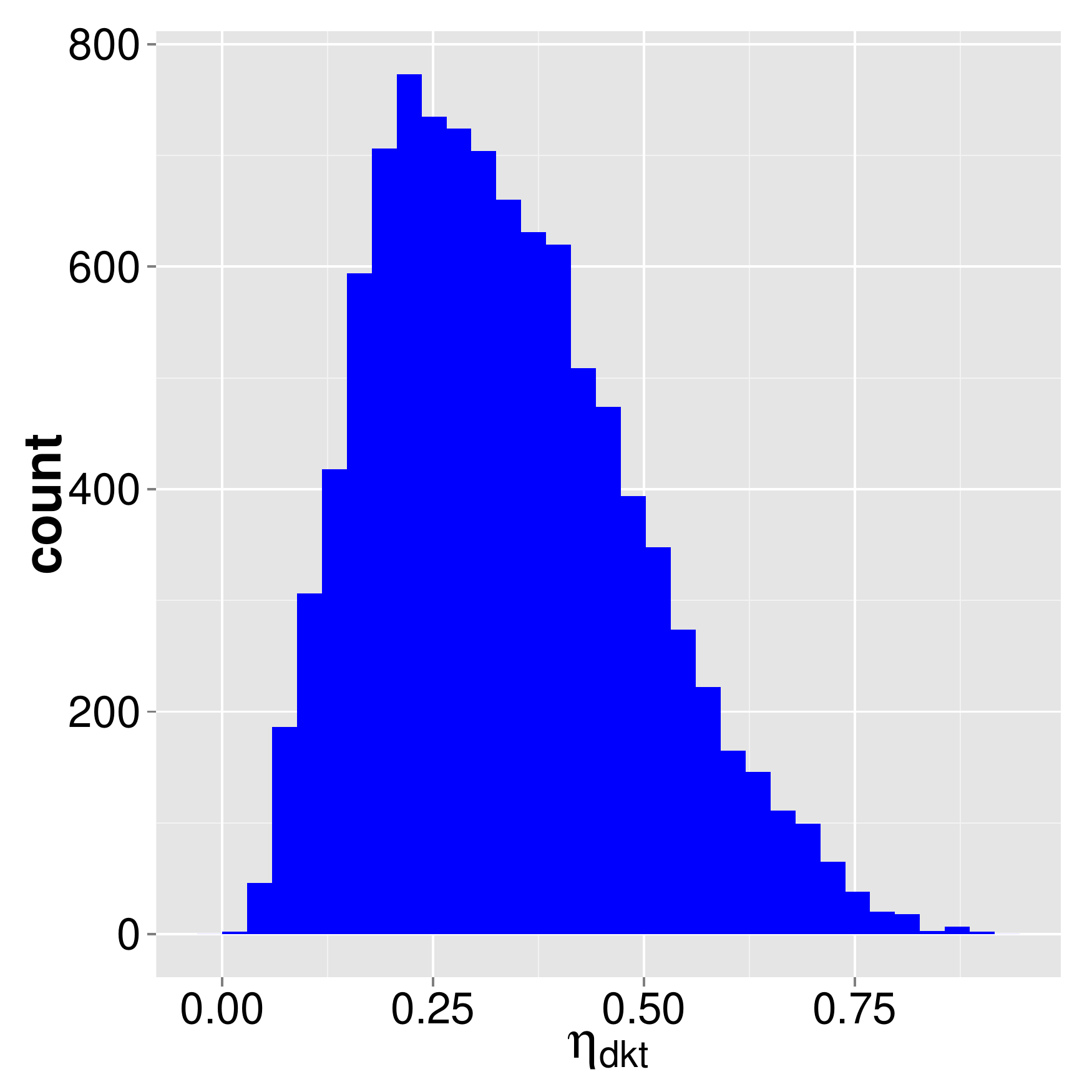}
  \caption{Prior for $P( z_{n,d,1}= k )$ }
  \label{subfig:Prob_Eta_Hist}
\end{subfigure}
\begin{subfigure}{.3\textwidth}
  \centering
  \includegraphics[width=1\textwidth]{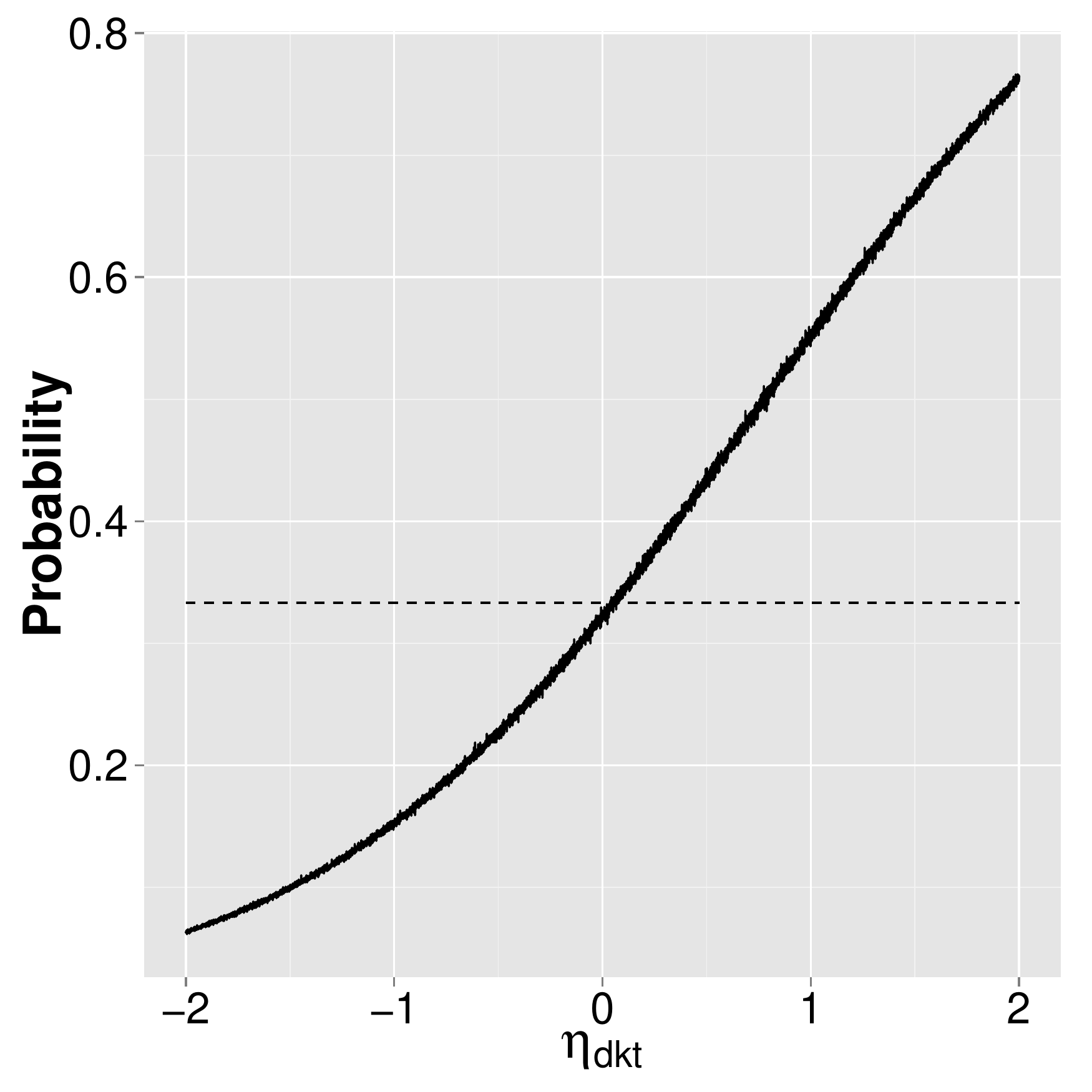}
  \caption{$E[ P(z_{n,d,1}=k) | \eta_{d,k,1}]$}
  \label{subfig:Exp_Prob_Eta}
\end{subfigure}
\end{figure}

We specify that $a^2=0.25$.  This choice of variance for the distribution of $\eta_{d,k,t} | \alpha_{k,t}$ allows individual documents a wide range of topic proportions even if one specific topic is more prevalent overall.  The histogram in Figure \ref{subfig:Eta_Hist} presents samples from the marginal distribution for $\eta_{d,k,1}$.  Again, simply considering the uncertainty in $\eta_{d,k,1}$ is insufficient for examining the uncertainty in the document topic proportions.  The histogram in Figure \ref{subfig:Prob_Eta_Hist} shows samples from the marginal distribution for document topic proportions when $K=3$.  If $K$ increases dramatically, the variance parameters for $C_{k,0}$, $\delta^2$, and $a^2$ need to be re-examined to choose reasonable levels of uncertainty for the $K-1$ probability simplex associated with each document.  

Conditioning on the value of $\eta_{d,k,1}$ but retaining uncertainty in the remaining $K-1$ parameters, $\eta_{d,-k,1}$, allows us to examine the relationship between $\eta_{d,k,1}$ and the expected proportion of the $d^{th}$ document alloted to topic $k$.  The solid line in Figure \ref{subfig:Exp_Prob_Eta} represents the expected value of the prior distribution for document topic proportions conditional on the value of $\eta_{d,k,1}$ when $K=3$.  The dashed black line in the same figure represents the topic proportion of $\frac{1}{K}$.      

Whenever specifying the number of topics ($K$) and vocabulary terms $(V)$ in a topic model, we advise analyzing the uncertainty on the probability simplices being modeled before attempting to make inference on topics or document proportions.  If topic overlap induced by the prior for $\beta_{k,v,t}$ is too high, the posterior topic overlap will also be quite high and nothing will have been learned.  If uncertainty in document topic proportions is too low -- specifically if $a^2$ is too small -- and documents are modeled as almost certain identical mixtures of $K$ topics, the induced prior belief is that at each time point the corpus contains $D_t$ nearly identical copies of the same document.  The result is that the inference procedure learns a single repeated topic -- corresponding to the single repeated document -- in the corpus.  Again, nothing has been learned.  As noted in \citet{wallach2009rethinking}, priors have an important effect on the ability of topic models to learn latent structure in documents.  

\section{Markov Chain Monte Carlo Algorithm}
\label{sec:Inference}
In this section, we develop a Gibbs sampling algorithm for posterior inference.  The objective is to sample from the joint posterior distribution of three sets of parameters: 1) the full collection of state-space parameters associated with each topic proportion, $\alpha_{\cdot,1:T} = \{ \alpha_{1,1:T}, \ldots, \alpha_{K, 1:T} \}$; 2) the full collection of document-specific topic proportions, $\eta_{\cdot,\cdot,1:T}$; and 3) the full collection of parameters associated with the dynamic probability distributions over vocabulary terms, $\beta_{\cdot, \cdot, 1:T}$.  The target posterior distribution is thus $p(\alpha_{\cdot, 1:T},\eta_{\cdot, \cdot, 1:T}, \beta_{\cdot, \cdot, 1:T} | W_{\cdot,1:T} )$.  Note that we are not inherently interested in the topic assignment of each word in each document.  Since conditioning on $Z_{\cdot,1:T}$ will be necessary for deriving the full conditionals, we sample the word-specific topic indicators, but we do not store them -- effectively marginalizing them out in our target posterior.  

By iteratively sampling from the full conditionals, as derived in Appendix \ref{App:Appendix_A}, we are able to construct a valid Gibbs sampler for the parameters and latent variables in the model.  In order to sample from these full conditionals, we utilize Polya-Gamma data augmentation \citep{PSW2013}.  \citet{chen_nips_2013} introduced the idea of a Polya-Gamma Gibbs sampler for a static logistic-Normal topic model.  We extend this idea to the dynamic setting.  In order to make inference for each $\beta_{k,v,t}$,  it is necessary to introduce an auxiliary $\zeta_{k,v,t} \sim PG(n_{k,t}^y,0)$.  Additionally, in order to make inference on $\eta_{d,k,t}$, it is necessary to introduce the auxiliary random variable $\omega_{d,k,t} \sim PG(N_{d,t},0)$. Note that our target posterior does not include the auxiliary variables.  To marginalize out the auxiliary $\zeta$ and $\omega$ from the posterior, we simply discard them and only store $\alpha_{\cdot, 1:T},\eta_{\cdot, \cdot, 1:T}, \beta_{\cdot, \cdot, 1:T}$. \\

A single MCMC sample from the target posterior is constructed as follows:
\begin{enumerate}
\item \label{itm:Beta_Step} For each topic $k$ and vocabulary term $v$, sample $\beta_{k,v,1:T} | \beta_{k,-v,1:T}, W_{\cdot,1:T}, Z_{\cdot,1:T}, \zeta_{k,v,1:T}$.  This step can be performed independently across topics.  The order in which the $\beta_{k,v,t}$ are updated is randomly permuted in the index $1,\ldots,V$ at each MCMC iteration.   
\item \label{itm:Zeta_Step} For each topic $k$, vocabulary term $v$, and time $t$, independently sample $\zeta_{k,v,t} | \gamma_{k,v,t}$.  
\item \label{itm:Eta_Step} For each document $d$, topic $k$, and time $t$, sample $\eta_{d,k,t} | Z_{\cdot,t}, \eta_{d,-k,t}, \omega_{d,k,t}, \alpha_{k,t}$.  This step can be performed independently across documents $d$ and time $t$.  The order in which the $\eta_{d,k,t}$ are updated is randomly permuted in the index $1,\ldots,K$ at each MCMC iteration. 
\item \label{itm:Omega_Step} For each document $d$, topic $k$, and time $t$, independently sample $\omega_{d,k,t} | \psi_{d,k,t}$.  The $\psi_{d,k,t}$ is a function of $\eta_{d,\cdot, t}$. Its construction will be detailed in Section \ref{subsec:FC_Eta}.
\item \label{itm:Alpha_Step} For each topic $k$, independently sample $\alpha_{k,1:T} | \eta_{\cdot,k,1:T}$.
\item \label{itm:Z_Step} For each word $n$, document $d$, and time $t$, independently sample $z_{n,d,t} |w_{n,d,t}, \eta_{d,\cdot,t}, \beta_{\cdot,\cdot,t}$. 
\end{enumerate}

Several steps in this sampling procedure are easily parallelized.  In particular, Steps \ref{itm:Beta_Step} and \ref{itm:Alpha_Step} can be parallelized across topics. Step \ref{itm:Eta_Step} can be parallelized across documents and time.  Step \ref{itm:Zeta_Step} can be parallelized across vocabulary terms, topics, and time.  Step \ref{itm:Omega_Step} can be parallelized across documents, topics, and time.  Step \ref{itm:Z_Step} can be parallelized across words in a document, documents, and time.  Our implementation of this algorithm performs all sampling steps in C++ using the R--C++ interface, Rcpp.  It also parallelizes these steps where possible using the mclapply function from the R-package parallel.  GPU and distributed computing architectures can be used for faster computation and inference in large corpora.  

\section{Polya-Gamma Approximation}
\label{sec:PG_Approx}
One of the primary computational bottlenecks is sampling from the Polya-Gamma distribution.  Each MCMC sample requires sampling $K\left( VT + \sum_{t = 1}^T D_t \right)$ Polya-Gamma random variables: one for each vocabulary term in each topic at each time point and one for every topic in each document at each time point.  For a corpus of 30,000 documents spanning 25 years that contains $K=20$ topics and 10,000 vocabulary terms, each MCMC sample requires sampling from this distribution $ 5.6  \times 10^{6} $ times.  It is clear that extremely fast sampling from this distribution is necessary for working with large corpora.  

We take advantage of the additive nature of the Polya-Gamma random variable.  Section 4.4 of \citet{PSW2013} notes that sampling $\omega \sim PG(b,c)$ when $b \in \mathbb{N}$ is equivalent to the construction $\omega = \sum_{i = 1}^b \tilde{\omega}_i$, where $\tilde{\omega}_i \sim PG(1,c)$.  For the Polya-Gamma random variates in text analysis, the $b$ parameter is very large.  In the case of $\omega_{d,k,t}$, $b$ corresponds to the number of words in document $d$ at time $t$.  For $\zeta_{k,v,t}$, $b$ corresponds to the number of words in the corpus assigned to topic $k$ at time $t$.  These large parameter values provide an additional computational burden in sampling the Polya-Gamma draws: to sample each of the $5.6 \times 10^{6}$ draws referenced above requires sampling the numerous underlying $PG(1,\cdot)$ variables to construct each draw.  This process is a significant limit to the computational speed.  

To approximate the sampling of a $PG(b,c)$ draw when $b \in \mathbb{N}$, which it always is in this application, we appeal to the additive construction of a $PG(b,c)$ random variable and the Central Limit Theorem.
\citet{chen_nips_2013} also consider an approximate sampler for Polya-Gamma random variables.  They rely on the additive property of the Polya-Gamma and the Central Limit Theorem to linearly transform a Polya-Gamma draw $PG(m,c)$ to approximate a draw from $PG(b,c)$ when $m<b$.  The advantage of our approximation is that we never need to sample from a Polya-Gamma random variable.  We only need to sample from a single approximating Gaussian distribution.  This removes the problem of additivity altogether for sufficiently large values of $b$.  

The Central Limit Theorem provides that:
\begin{align*}
\sqrt{b} \left( \left( \frac{1}{b} \sum_{i=1}^b \tilde{\omega}_i \right) - E[\tilde{\omega}_i] \right) \stackrel{d}{\Rightarrow} N(0, Var(\tilde{\omega}_i) ). 
\end{align*}
This suggests that 
\begin{align*}
\omega = \sum_{i=1}^b \tilde{\omega}_i \stackrel{d}{\approx} N \left( b E[\tilde{\omega}_i], b Var(\tilde{\omega}_i) \right)
\end{align*}
for large values of $b$. The mean and variance of the approximating Normal distribution are the appropriately scaled mean and variance of $\tilde{\omega}_i \sim PG(1,c)$. 

For a Polya-Gamma random variable $\omega \sim PG(b,c)$, its mean is $E[\omega] = \frac{b}{2c} \tanh\left( \frac{c}{2} \right)$.  Its variance is $Var(\omega) = \frac{b}{4c^3} \sech^2 \left( \frac{c}{2} \right) \left( \sinh(c) - c \right)$. While \citet{PSW2013} present the mean of the Polya-Gamma, they do not present the variance.  Appendix \ref{App:Appendix_B} gives a full derivation of the mean and variance of a $PG(b,c)$ random variable utilizing derivatives of the characteristic function.  A derivation that utilizes the Weierstrass Factorization Theorem is presented by William A. Huber on the web forum Cross Validated \citep{Huber2014}.  Because we take a different approach to calculating the variance, we present our calculations in Appendix \ref{App:Appendix_B}.      

Rather than sampling many times from the $PG(1,c)$, we are able to generate an approximate draw from a $PG(b,c)$ distribution with a single draw from a Gaussian.  Table \ref{table:Comp_Cost} demonstrates that our Gaussian approximation achieves a significant reduction in time required to sample from the Polya-Gamma distribution.    

\begin{table}[h!]
\caption{Comparison of time required to draw 1000 Polya-Gamma samples $PG(b_i, c_i)$ where the parameters $b_i,c_i$ are unique for each sample. $b_i \sim Pois(150)$ and $c_i \sim N(0,1)$.}  
\label{table:Comp_Cost}
\centering
\begin{tabular}{rlrrr}
  \hline
 & Method & replications & elapsed & relative \\ 
  \hline
2 & Gaussian & 100 & 0.03 & 1.00 \\ 
  3 & \citet{chen_nips_2013} & 100 & 0.31 & 11.52 \\ 
  1 & BayesLogit & 100 & 2.75 & 102.04 \\ 
   \hline
\end{tabular}
\end{table}

\begin{figure}[h!]
\centering
\caption{Overlayed histograms of samples from approximate Polya-Gamma samplers with samples from a $PG(100,-1)$ distribution.  Left: approximation of \citet{chen_nips_2013}.  Right: Gaussian approximation.}
\begin{subfigure}{.4\textwidth}
  \centering
  \includegraphics[width=1\textwidth]{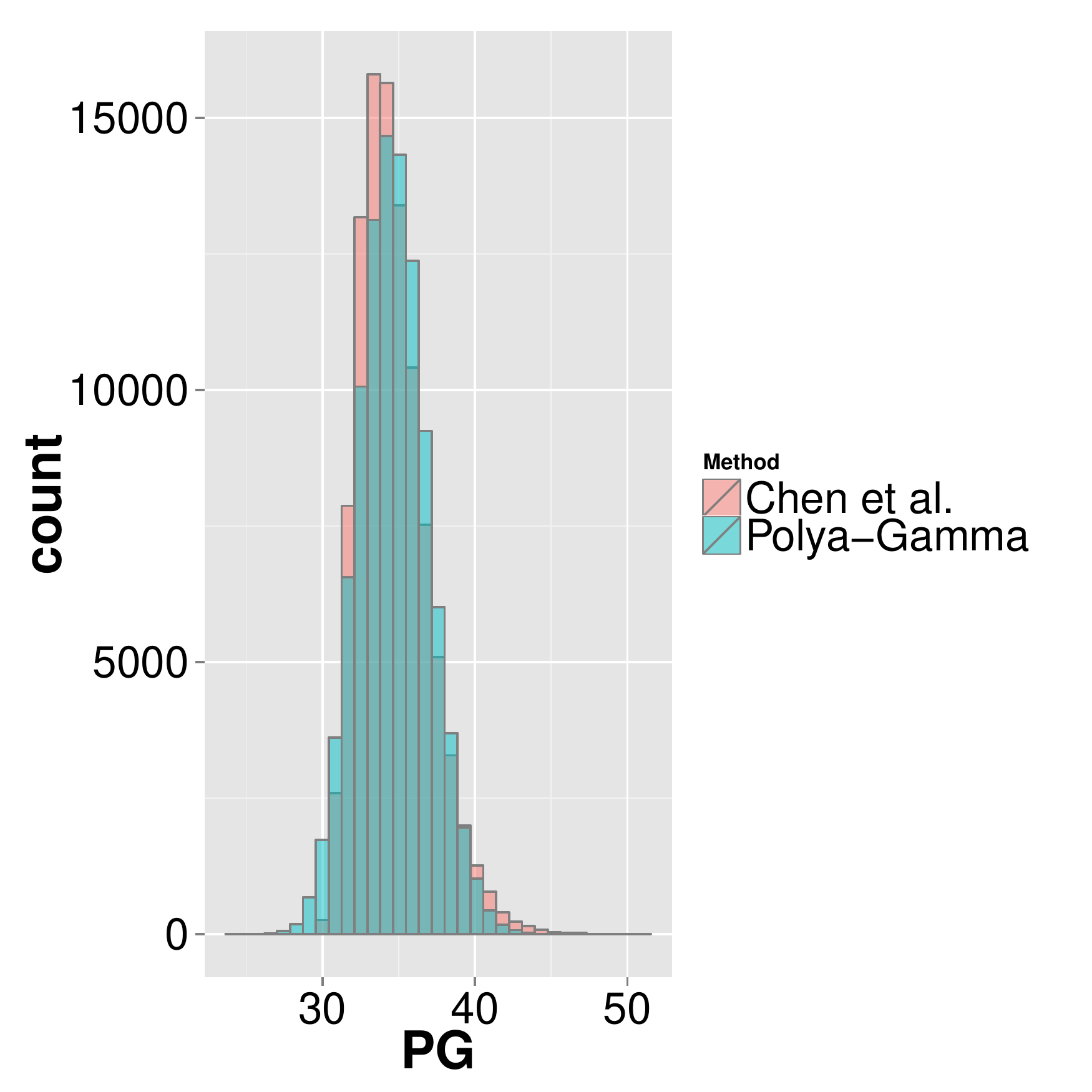}
  \caption{\citet{chen_nips_2013} $(m=5)$ }
  \label{subfig:NIPS_PG_m1}
\end{subfigure}%
\begin{subfigure}{.4\textwidth}
  \centering
  \includegraphics[width=1\textwidth]{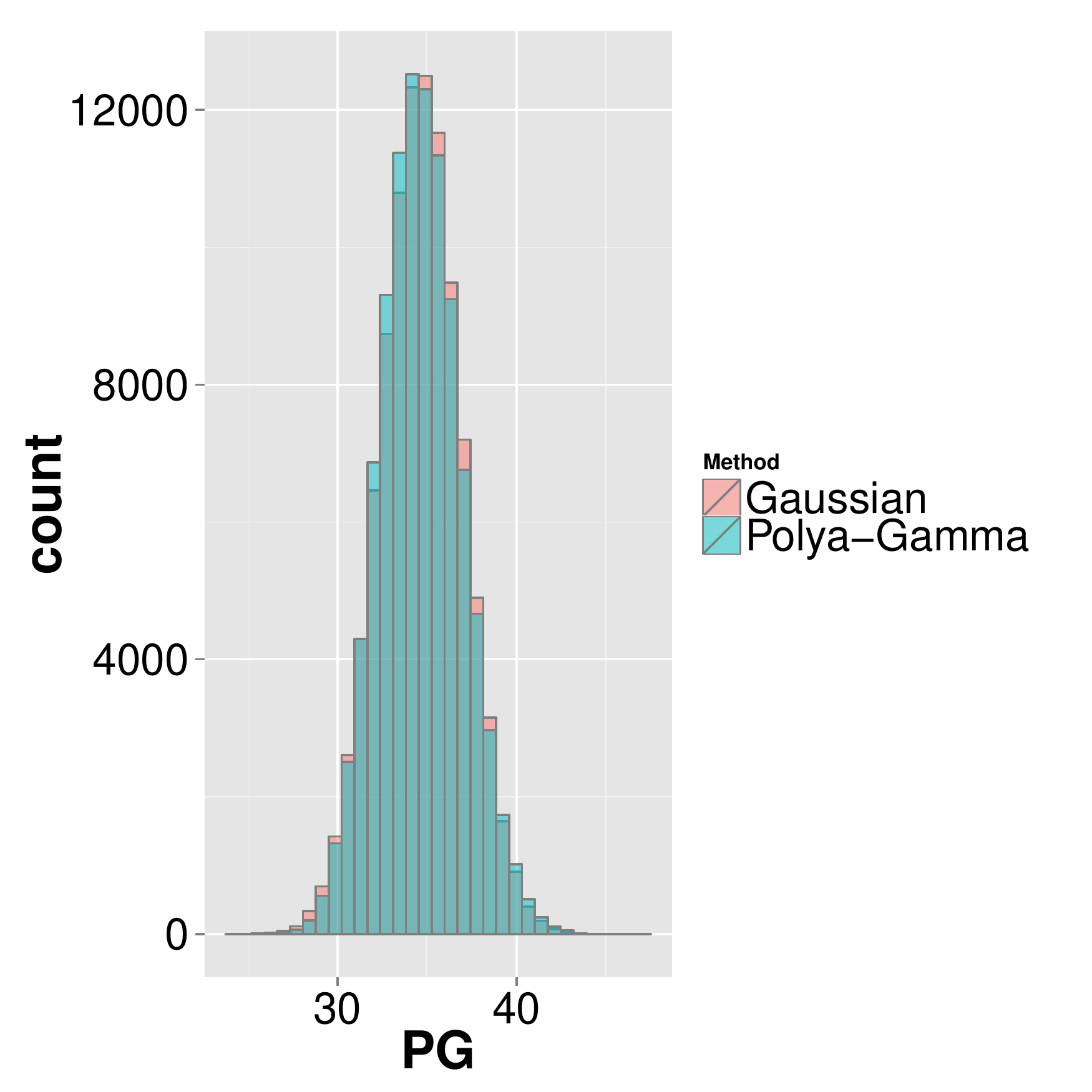}
  \caption{Gaussian}
  \label{subfig:Gaussian_PG_m1}
\end{subfigure}
\end{figure}

Figure \ref{subfig:NIPS_PG_m1} compares a histogram of samples from the \citet{chen_nips_2013} Polya-Gamma approximation to a histogram of samples from the true distribution which are generated from the BayesLogit package in R \citep{BayesLogit}.  Figure \ref{subfig:Gaussian_PG_m1} compares a histogram of samples from the Gaussian distribution to a histogram of samples from BayesLogit.  The Gaussian approximation works very well when $b \in \mathbb{N}$ is larger than $20$.

\section{Simulation Study}
\label{sec:SimStudy}

To validate and examine the reproducibility of our MCMC algorithm, we conducted a simulation study.  We constructed a synthetic data set with $K = 3$ topics and a vocabulary with $V = 1000$ terms.  The objective of this simulation study is to benchmark the computational method in recovering a known truth as compared to existing variational strategies for inference in the DTM \citep{Gerrish2011}.  

The synthetic data set was constructed by sampling a random number of documents at $T = 5$ different time points.  The number of documents at each time point was sampled from a Poisson distribution with mean of $1000$.  Each document was endowed with a random number of words, which was sampled from a Poisson distribution with mean $150$.

The proportion of each document allocated to the three respective topics was generated by sampling from the DTM data generating model for document proportions with $m_{k,0} =0$, $C_{k,0} = 0.025$, $\delta^2 = 0.001$, and $a^2 = 0.5$.  Setting $\delta^2 = 0.001$ makes it likely that there is no overall trend to the topics.  Setting $a^2 = 0.5$ ensures heterogeneity in the corpus.

\begin{figure}[h!]
\centering
\caption{Posterior means for probabilities of $v^{th}$ term for each topic}
\label{fig:Topic_Post_Mean}
\begin{subfigure}{.3\textwidth}
  \centering
  \includegraphics[width=1\textwidth]{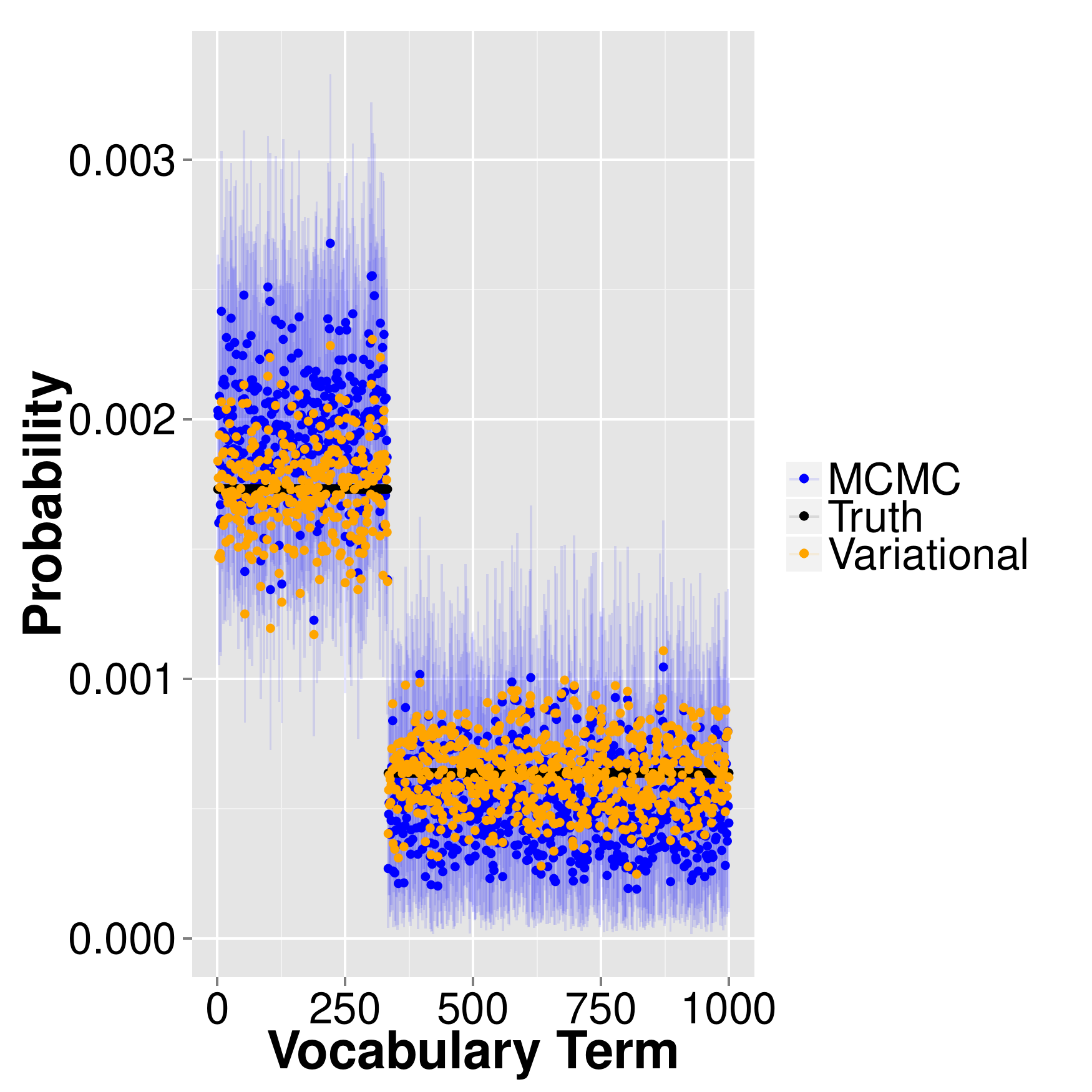}
  \caption{Topic 1}
  \label{subfig:Topic_1_Post}
\end{subfigure}%
\begin{subfigure}{.3\textwidth}
  \centering
  \includegraphics[width=1\textwidth]{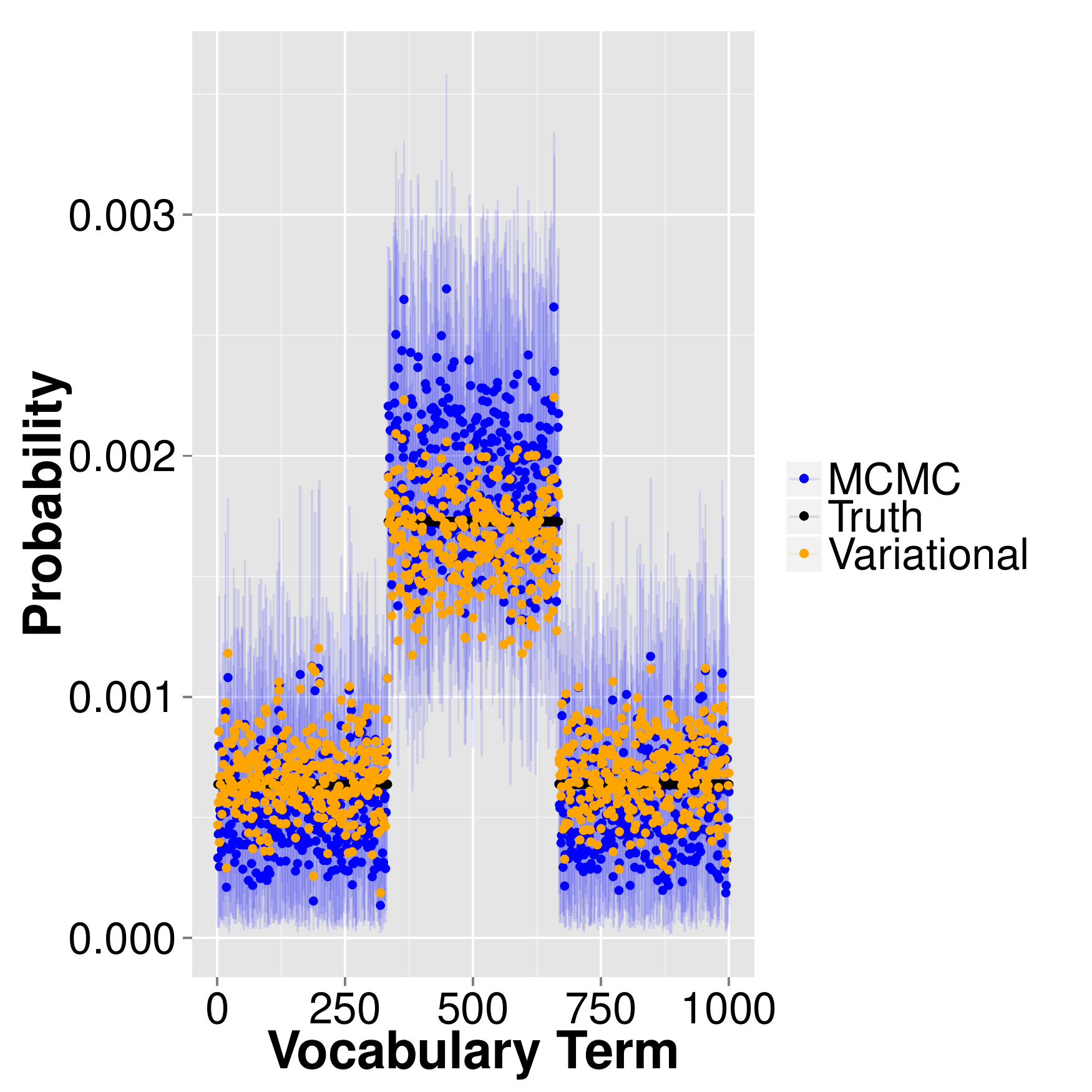}
  \caption{Topic 2}
  \label{Topic_2_Post}
\end{subfigure}
\begin{subfigure}{.3\textwidth}
  \centering
  \includegraphics[width=1\textwidth]{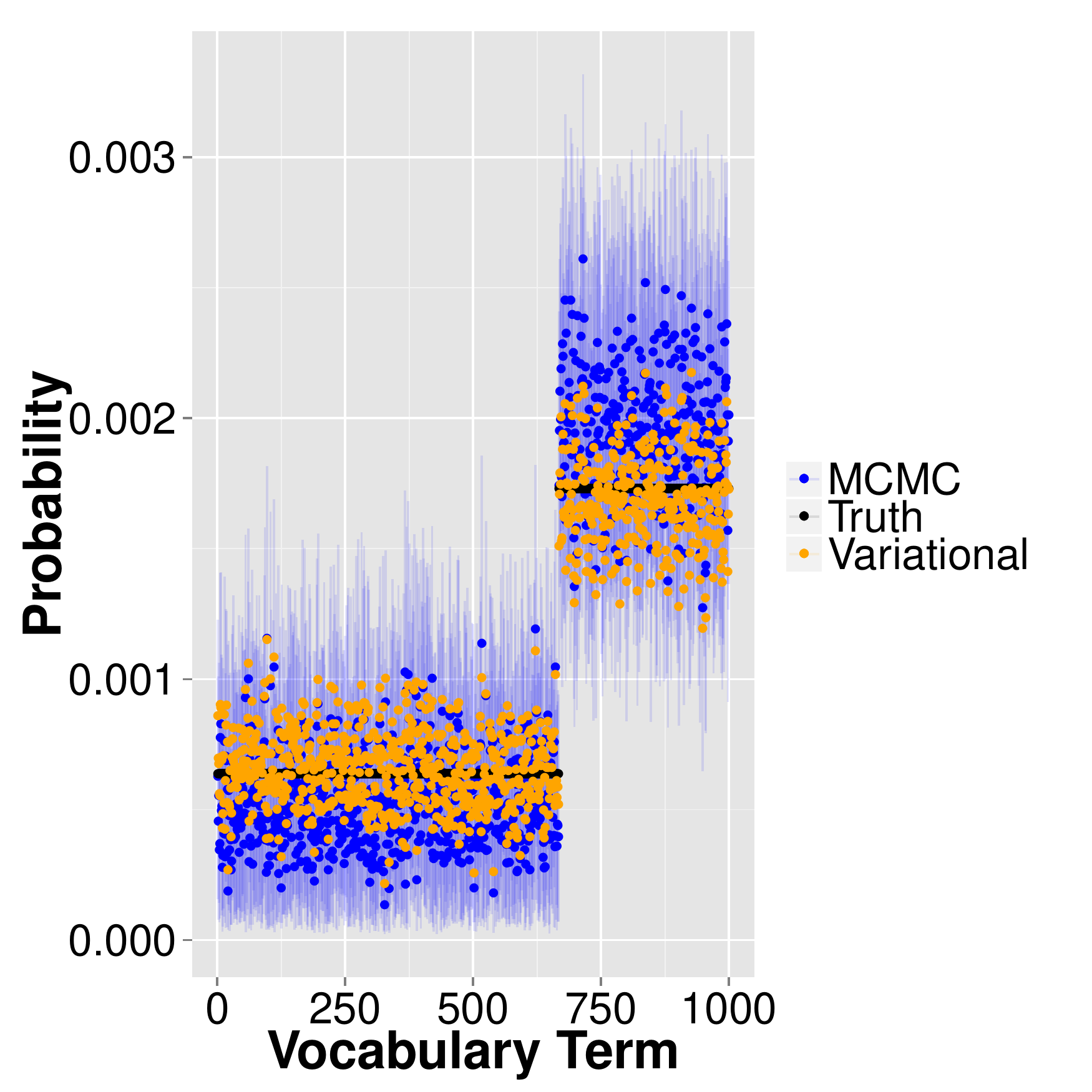}
  \caption{Topic 3}
  \label{subfig:Topic_3_Post}
\end{subfigure}
\end{figure}

The three distributions over the vocabulary terms were constructed so that three disjoint subsets of vocabulary terms would occur with high probability in three separate topics.  The first topic places high probability on vocabulary terms 1 through 333. The second topic places high probability on terms 334-667.  The third topic places high probability on terms 668-1000.   The black line in Figure \ref{subfig:Topic_1_Post} presents the truth for topic 1 at $t=1$.  The true topics were allowed to evolve after $t=1$ with an innovation variance of $\sigma^2 = .01$ .  

The blue dots in Figure \ref{fig:Topic_Post_Mean} represent the posterior mean of the topic probability for each vocabulary term, as estimated by the MCMC algorithm.  The light blue verticals associated with each blue dot represent the 95\% credible interval for the probability.  The orange dots represent the variational estimate from the DTM release of \citet{Gerrish2011}.  Figure \ref{fig:Topic_Post_Mean} demonstrates that the posterior means for the probability of the $v^{th}$ term in each topic correspond reasonably well to the true probability for both MCMC and variational methods.

\begin{figure}[h!]
\centering
\caption{Left: Total Variation distance between estimated vocabulary distributions (i.e. topics) and truth.  Middle: Total Variation distances between estimated document-specific topic proportions and true topic proportions for all documents at time $t=5$. Right: Marginal probability of topics over time.}
\label{fig:DTM_Compare}
\begin{subfigure}{.3\textwidth}
  \centering
  \includegraphics[width=1\textwidth]{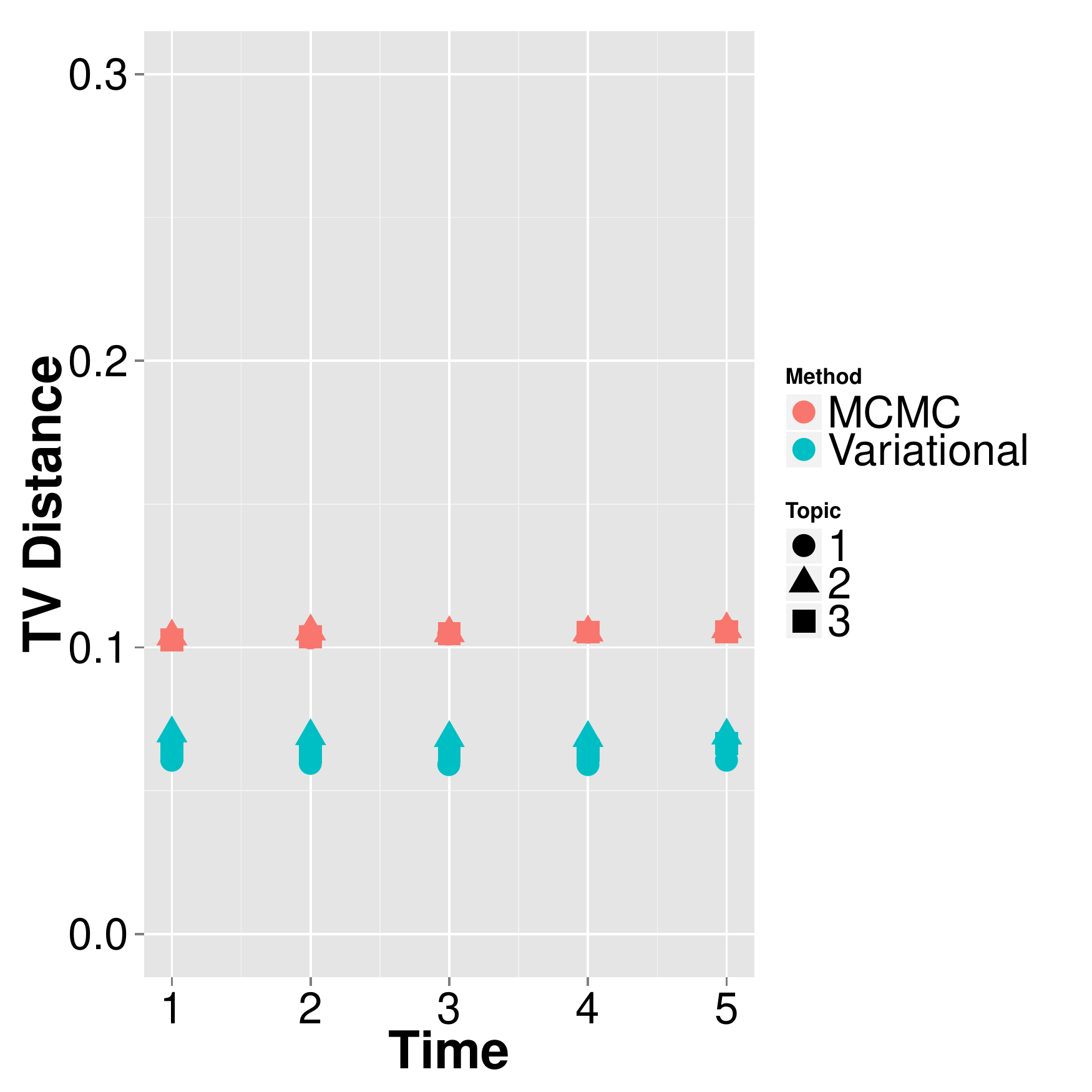}
  \caption{Topic TV}
  \label{subfig:Topic_TV}
\end{subfigure}%
\begin{subfigure}{.3\textwidth}
  \centering
  \includegraphics[width=1\textwidth]{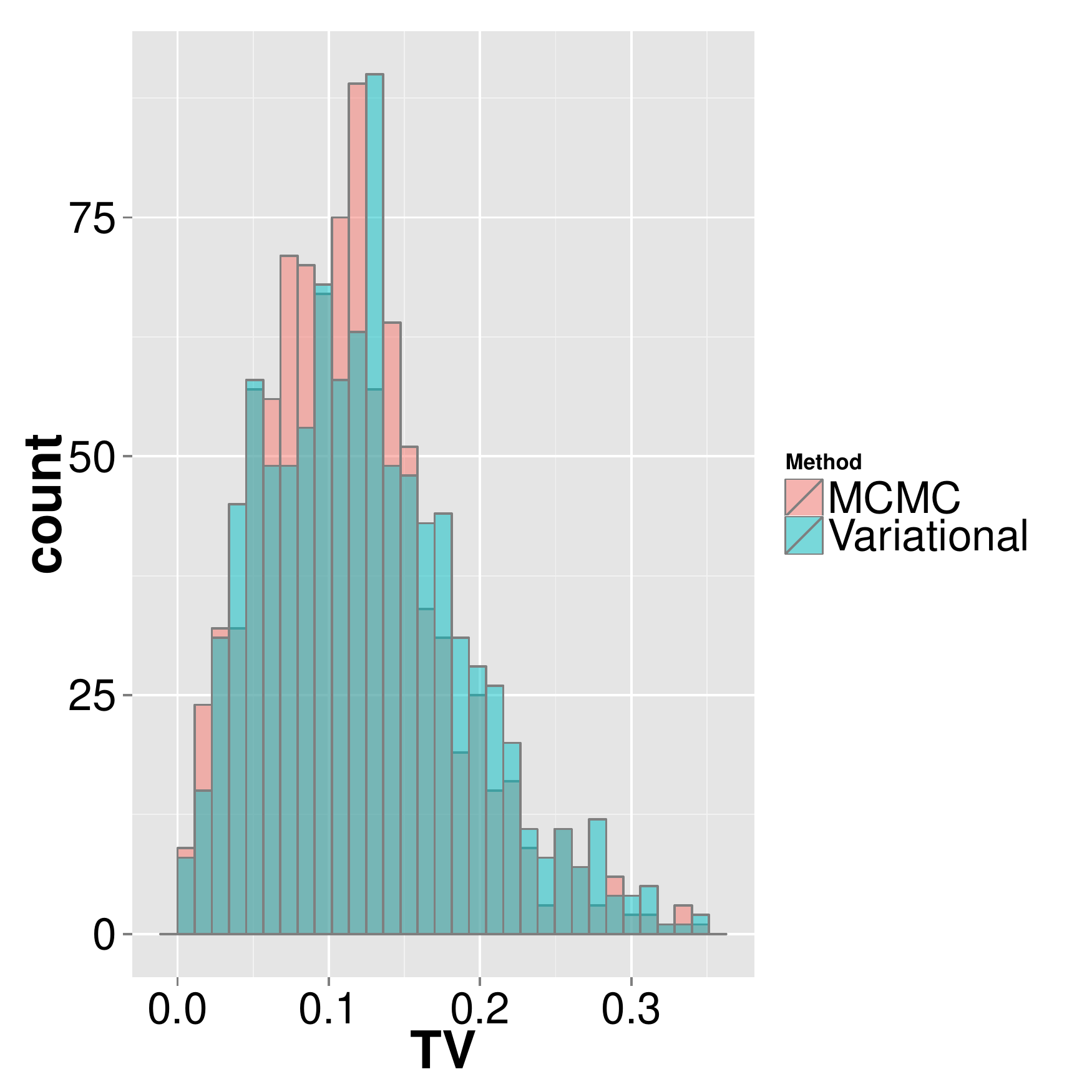} 
  \caption{Document $\%$ TV}
  \label{subfig:Doc_TV}
\end{subfigure}
\begin{subfigure}{.3\textwidth}
  \centering
  \includegraphics[width=1\textwidth]{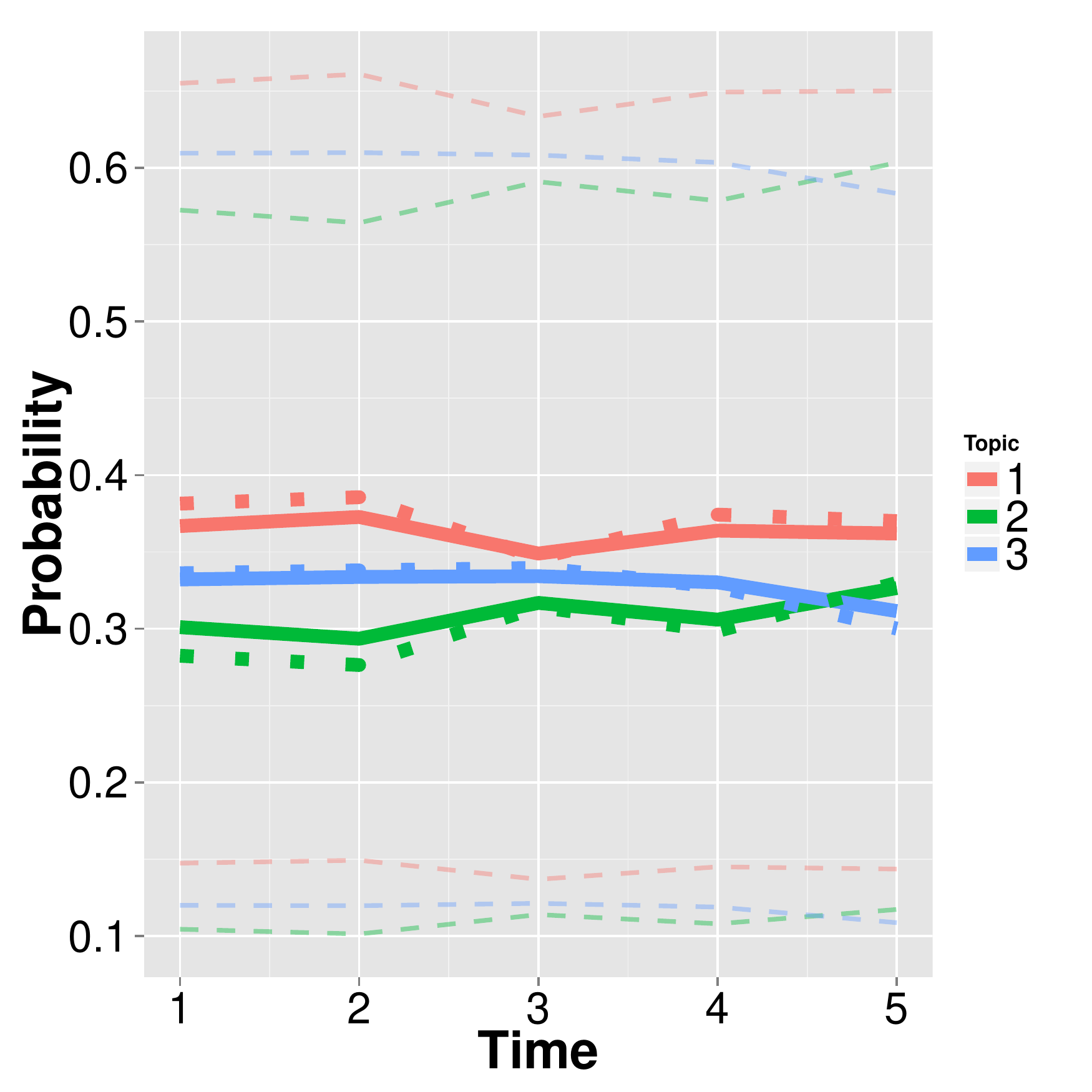}
  \caption{Topic $\%$}
  \label{subfig:Doc_Pct}
\end{subfigure}
\end{figure}

Figure \ref{subfig:Topic_TV} demonstrates that the variational Kalman filter slightly outperforms MCMC in recovering topics.  This confirms what appears visually evident in Figure \ref{fig:Topic_Post_Mean}.  However, the variational approximation, which endows each document with its own Dirichlet distributrion, is slightly outperformed by the DLTM when estimating document topic proportions.  The estimates in the DLTM benefit from sharing information across documents.  This is especially true when the number of topics, $K$, is misspecified, which will be discussed more in Section \ref{subsec:misspecification}.  Sharing information across documents is essential in identifying which topics are extraneous to the corpus as a whole.  Information sharing across documents is also important when the marginal probability of a topic increases rapidly.   Section \ref{subsec:Complex_Trends} demonstrates that when the marginal probabilities of topics exhibit linear or quadratic trends, the DLTM outperforms the variational implementation of DTM in estimating document topic proportions.      

Figure \ref{subfig:Doc_Pct} demonstrates the marginal probability of each topic over the time span of the synthetic corpus.  The solid lines represent the posterior mean from the MCMC algorithm.  The light dashed lines are the $95\%$ credible interval associated with each topic.  The true trend used to simulate the data is presented with the thick dashed line.  While the uncertainty interval for these trends is wide, it should be viewed as the uncertainty in the topic proportions of a new document introduced to the corpus at each time point.  It is appropriate to have large uncertainty in the thematic composition of a new document.  Figure \ref{subfig:Doc_Pct} is one of the primary advantages of our MCMC algorithm.  The variational approximation is unable to infer global topic trends because each document is allowed its own Dirichlet distribution.

\subsection{Reproducibility}
\label{subsec:reproducibility}
The second objective of our simulation study is to examine the reproducibility of our MCMC algorithm.  To do this, we ran five different simulations where each simulation was initialized very differently.  To initialize the parameters in each run of the simulation, we sampled from the data generating model with high variance to provide an overly disperse initialization.  Each latent topic variable $z_{n,d,t}$ was initialized randomly by sampling from the full conditional for $z_{n,d,t}$ given initializations of the parameters.  

We ran the five different MCMC simulations for 600,000 iterations and thinned the chain by only recording every $100^{th}$ sample -- resulting in 6,000 MCMC samples.  We discarded the first 2,000 as a burn-in period.  On an 8 core workstation, this took approximately one day.

Before comparing posterior summaries for topics across the different chains, we relabeled the topics in chains 2-5 so that they were all consistent with the topics from the first chain.  To match the topics in chains 2-5 to the first topic in chain 1, we computed the total variation distance between each topic in each chain to the topics in chain 1.  We then relabeled the topics so as to minimize the total variation distance with the constraint that no label can be re-used.  Visual inspection of the probability distributions confirmed this method to be effective for ensuring consistency in labels across the multiple different chains.  

\begin{figure}[h!]
\centering
\caption{Left: Maximum TV distance between posterior means of topics from chains 2-5 and chain 1 for each topic. Right: Boxplot of maximum TV distances between posterior means of document topic proportions from chains 2-5 and chain 1 for all documents}
\begin{subfigure}{.4\textwidth}
  \centering
  \includegraphics[width=1\textwidth]{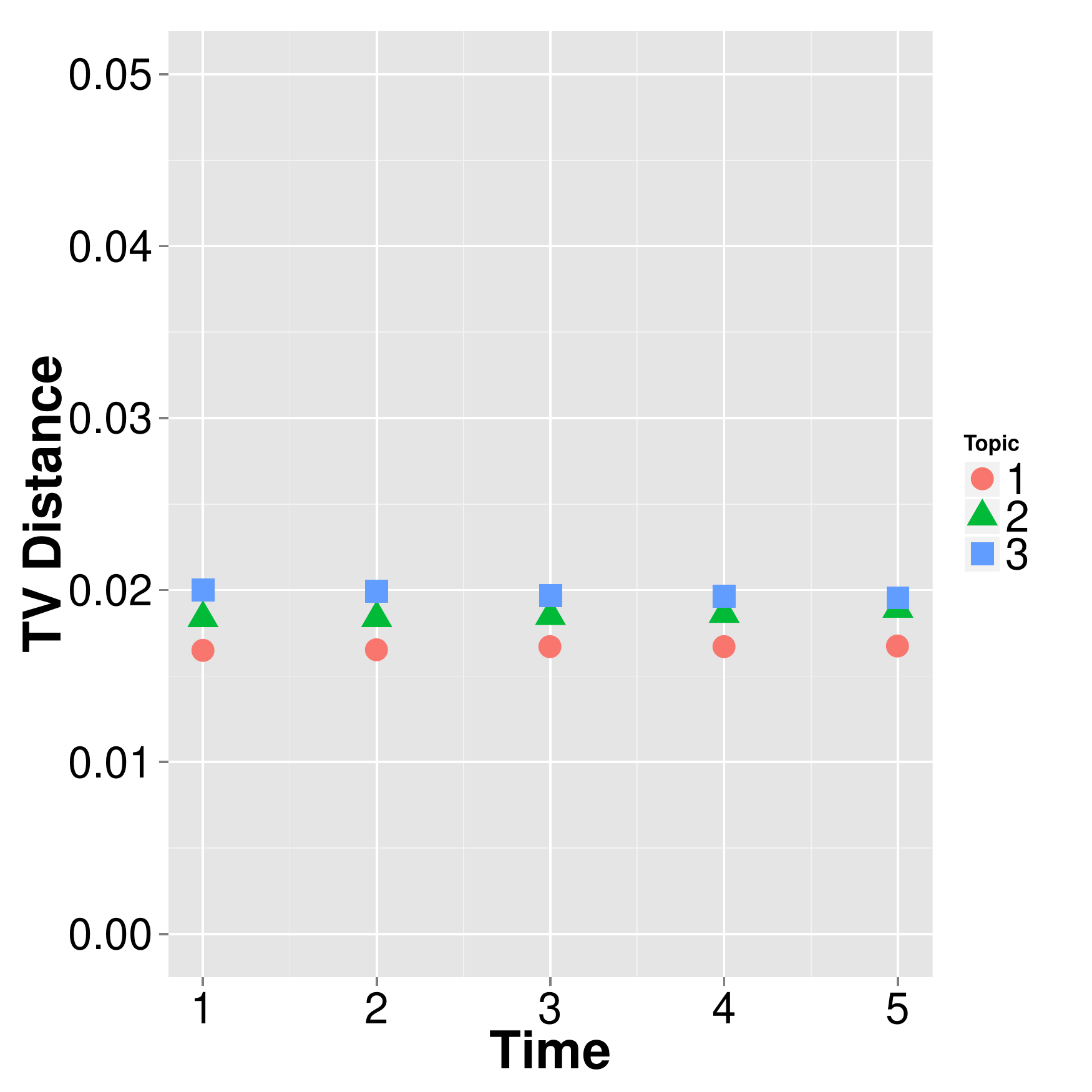}
  \caption{Topic TV}
  \label{subfig:Max_Topic_TV}
\end{subfigure}%
\begin{subfigure}{.4\textwidth}
  \centering
  \includegraphics[width=1\textwidth]{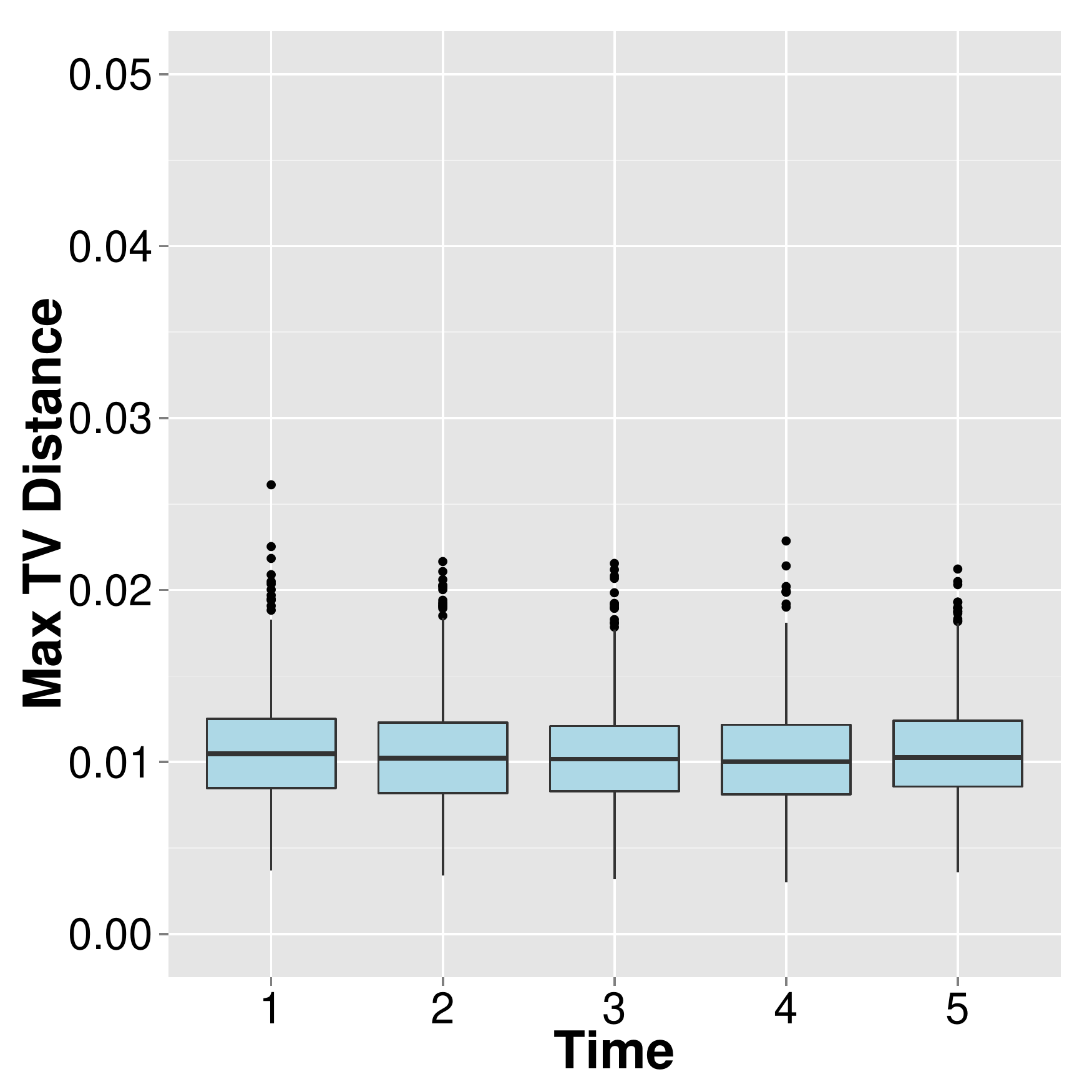} 
  \caption{Document $\%$ TV}
  \label{subfig:Max_Doc_TV}
\end{subfigure}
\end{figure}

Figure \ref{subfig:Max_Topic_TV} was constructed by computing the total variation (TV) distance between the posterior means of chains 2-5 and chain 1 and then taking the maximum.  As an example, for topic 3 the maximum TV distance between the posterior means for chains 2-5 and the posterior mean for chain 1 is 0.02.  Similarly, for topic 1, the maximum TV distance between the posterior means for chains 2-5 and the posterior mean for chain 1 is approximately 0.015.  The magnitude of these maximum TV distances is an encouraging sign that the MCMC algorithm has converged.  

Next we look at the maximum TV distances across document topic proportions.  For each document, we compute the TV distance between the posterior means for topic proportions from chains 2-5 and the posterior mean for topic proportion from chain 1.  We then take the maximum TV distance for each document.  The boxplot in \ref{subfig:Max_Doc_TV} presents the distribution of the maximum TV distances of chains 2-5 to the first chain for all documents in the corpus.

To more completely assess the convergence of our MCMC algorithm, we borrow ideas from \citet{gelman1992} and consider the total variation distance across the different chains and within a single chain.  If the magnitude of the TV distances presented in Figure \ref{subfig:Max_Topic_TV} is small compared to the across chain and within chain TV distances, we are more confident in the MCMC convergence.

\begin{figure}[h!]
\centering
\caption{ Left: Across Chain TV Distance to initialization 1 for topic 1. Right: Within Chain TV for each topic in initialization 1.}
\begin{subfigure}{.4\textwidth}
  \centering
  \includegraphics[width=1\textwidth]{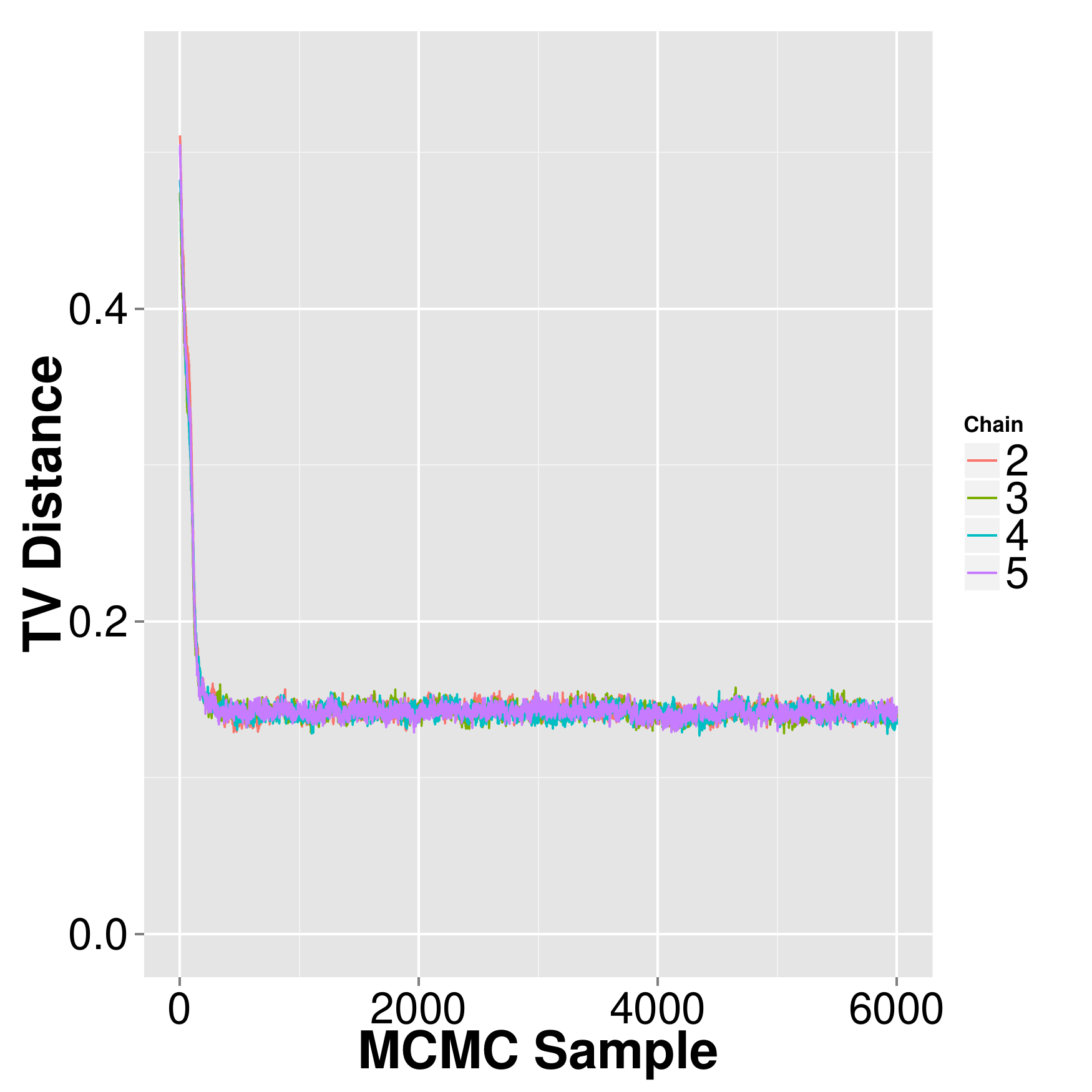}
  \caption{Across Chain TV}
  \label{subfig:Gelman_Rubin_Across}
\end{subfigure}%
\begin{subfigure}{.4\textwidth}
  \centering
  \includegraphics[width=1\textwidth]{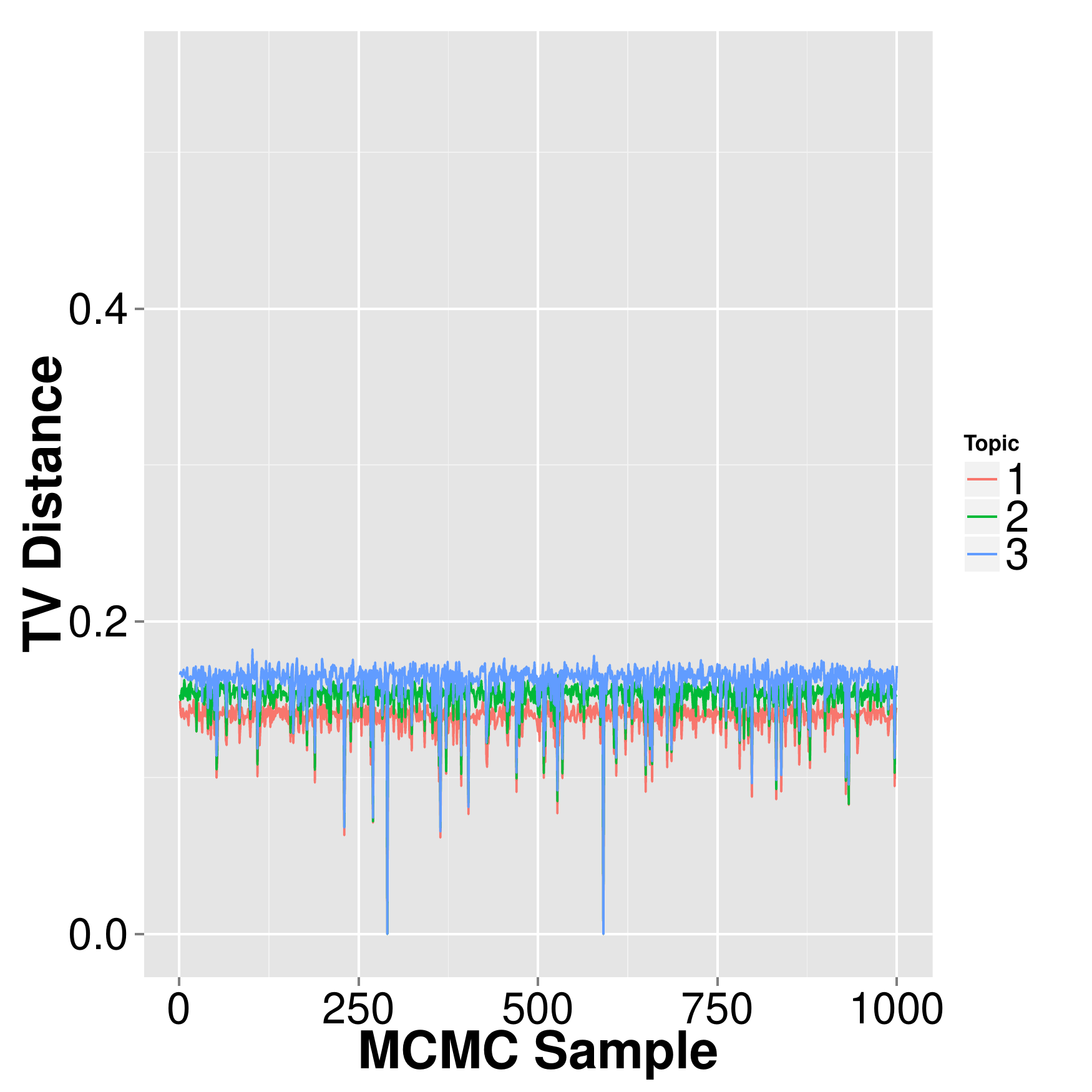} 
  \caption{Within Chain 1 TV}
  \label{subfig:Gelman_Rubin_Within}
\end{subfigure}
\end{figure}

Figure \ref{subfig:Gelman_Rubin_Across} shows the total variation distance for topic 1 computed between MCMC samples from chains 2-5 and chain 1.  After the $500^{th}$ recorded sample, the TV distance across chains settles near $0.15$.  Figure \ref{subfig:Gelman_Rubin_Within} shows the total variation between 1000 randomly selected pairs of post-burn-in posterior samples of topics within the first chain.

The magnitude of the TV distance across posterior means is relatively small compared to the TV distances of MCMC samples across chains and within the same chain.  This supports the conclusion that the MCMC algorithm has in fact converged.  

\subsection{Misspecification of $K$}
\label{subsec:misspecification}
One of the critical questions for topic modeling is how to specify the number of topics in a corpus.  Rarely is a true number of topics known to the researcher.  For this reason, behavior of the model and computational algorithm when the number of topics is misspecified is of practical importance.  In this section, we let $K=6$ and repeat the analysis of our synthetic data.   

\begin{figure}[H]
\centering
\caption{Posterior mean for $v^{th}$ term in Topics 1-6.}
\label{fig:Beta_Post_K6_1}
\begin{subfigure}{.3\textwidth}
  \centering
  \includegraphics[width=1\textwidth]{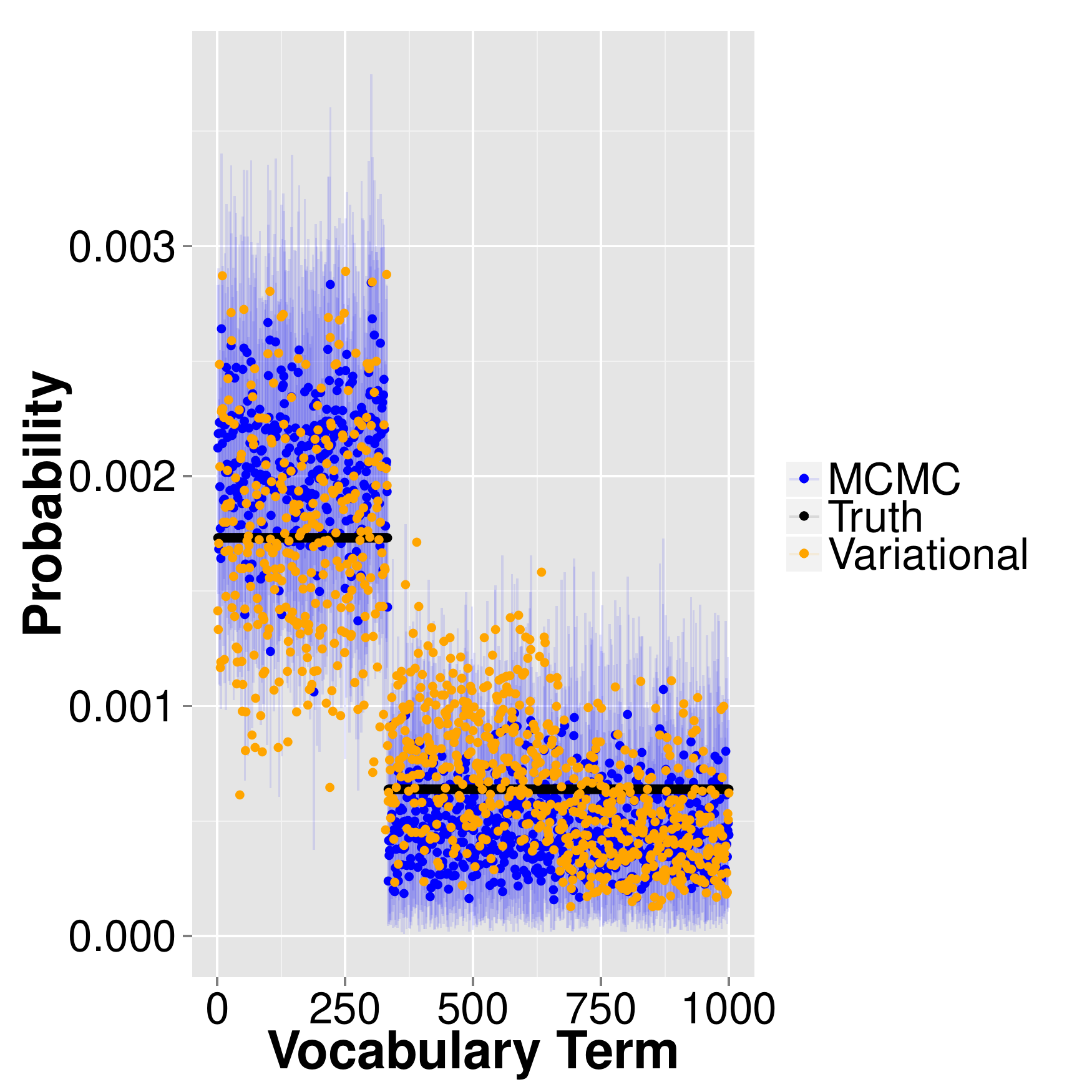}
  \caption{Topic 1}
\end{subfigure}%
\begin{subfigure}{.3\textwidth}
  \centering
  \includegraphics[width=1\textwidth]{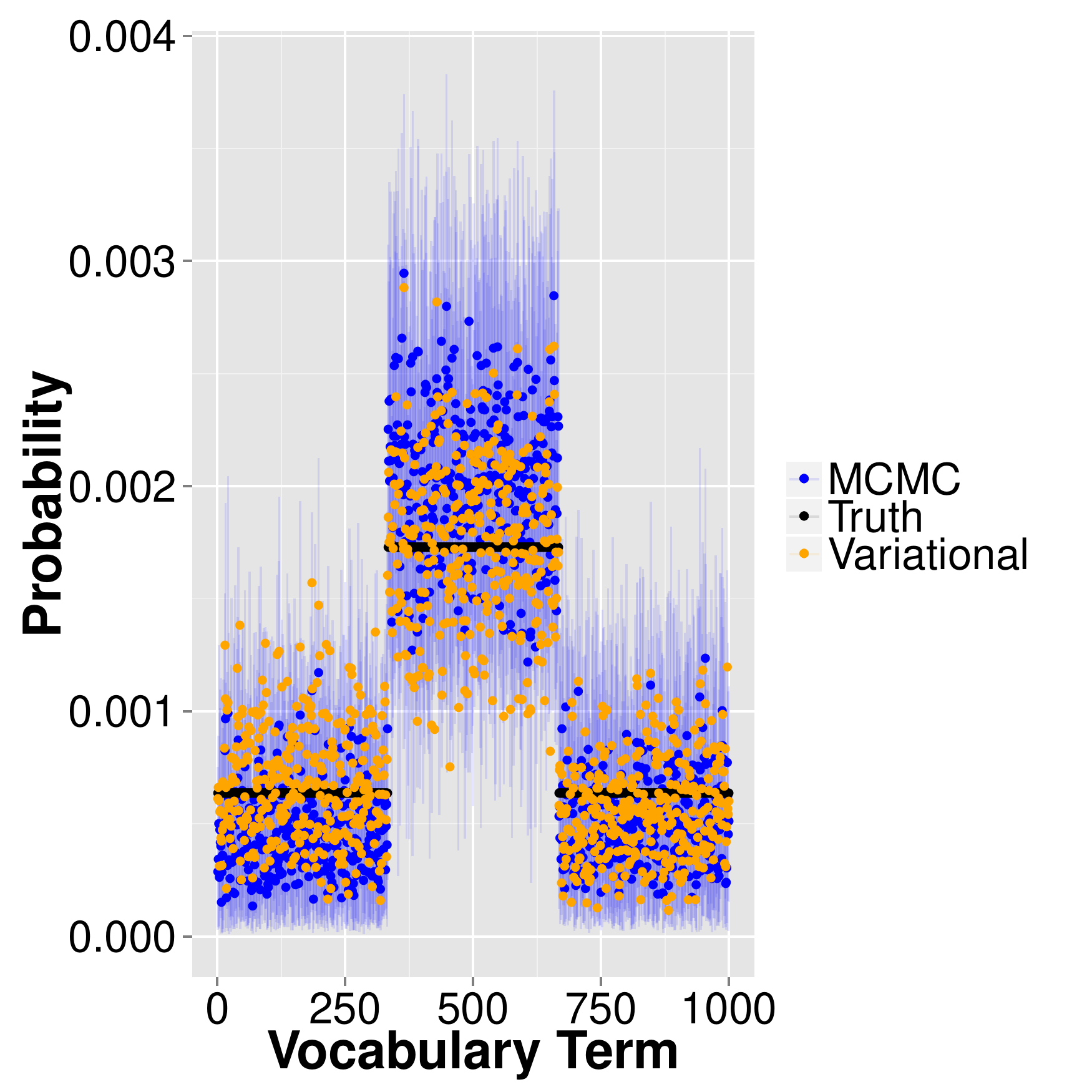}
  \caption{Topic 2}
\end{subfigure}
\begin{subfigure}{.3\textwidth}
  \centering
  \includegraphics[width=1\textwidth]{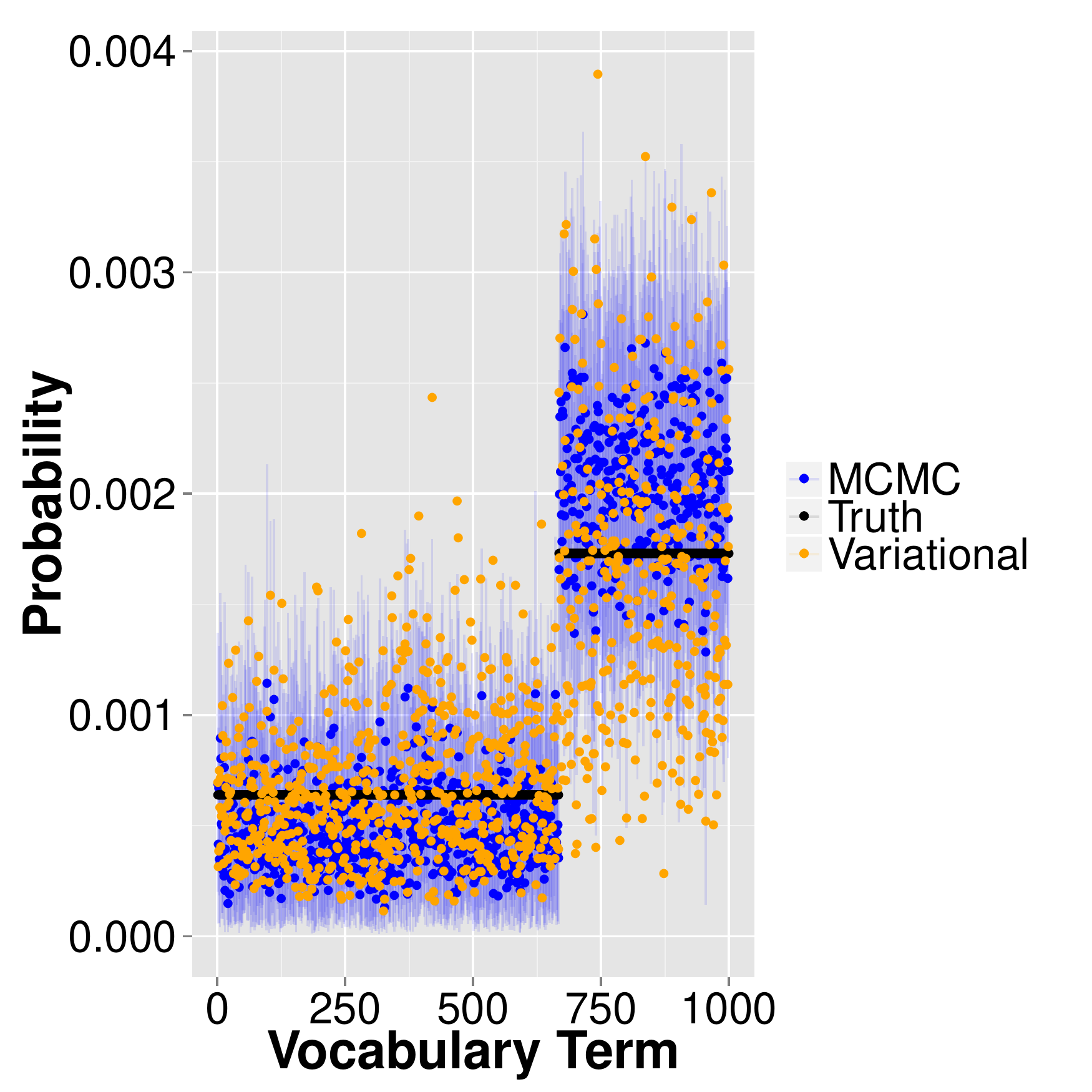}
  \caption{Topic 3}
\end{subfigure}
\begin{subfigure}{.3\textwidth}
  \centering
  \includegraphics[width=1\textwidth]{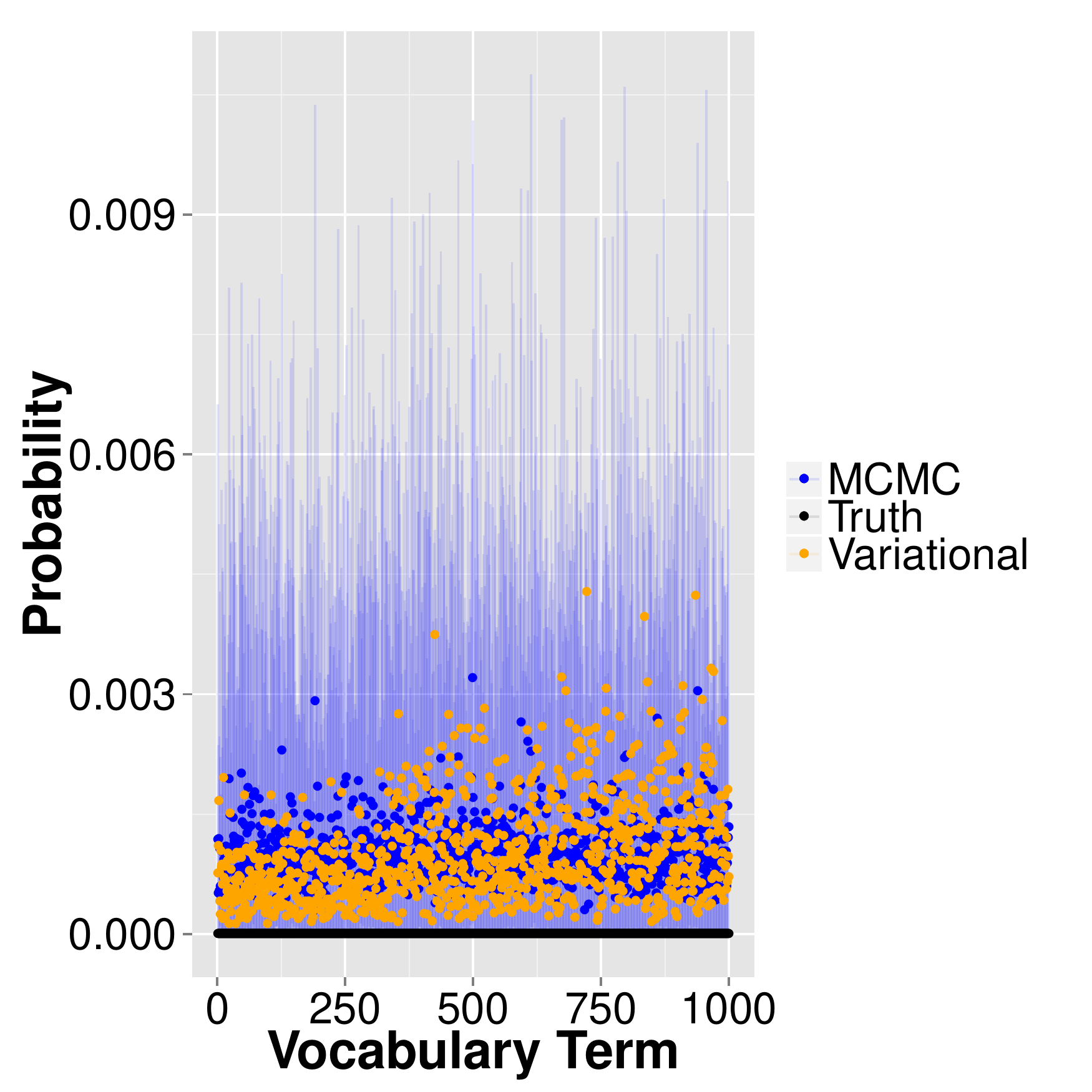}
  \caption{Topic 4}
\end{subfigure}%
\begin{subfigure}{.3\textwidth}
  \centering
  \includegraphics[width=1\textwidth]{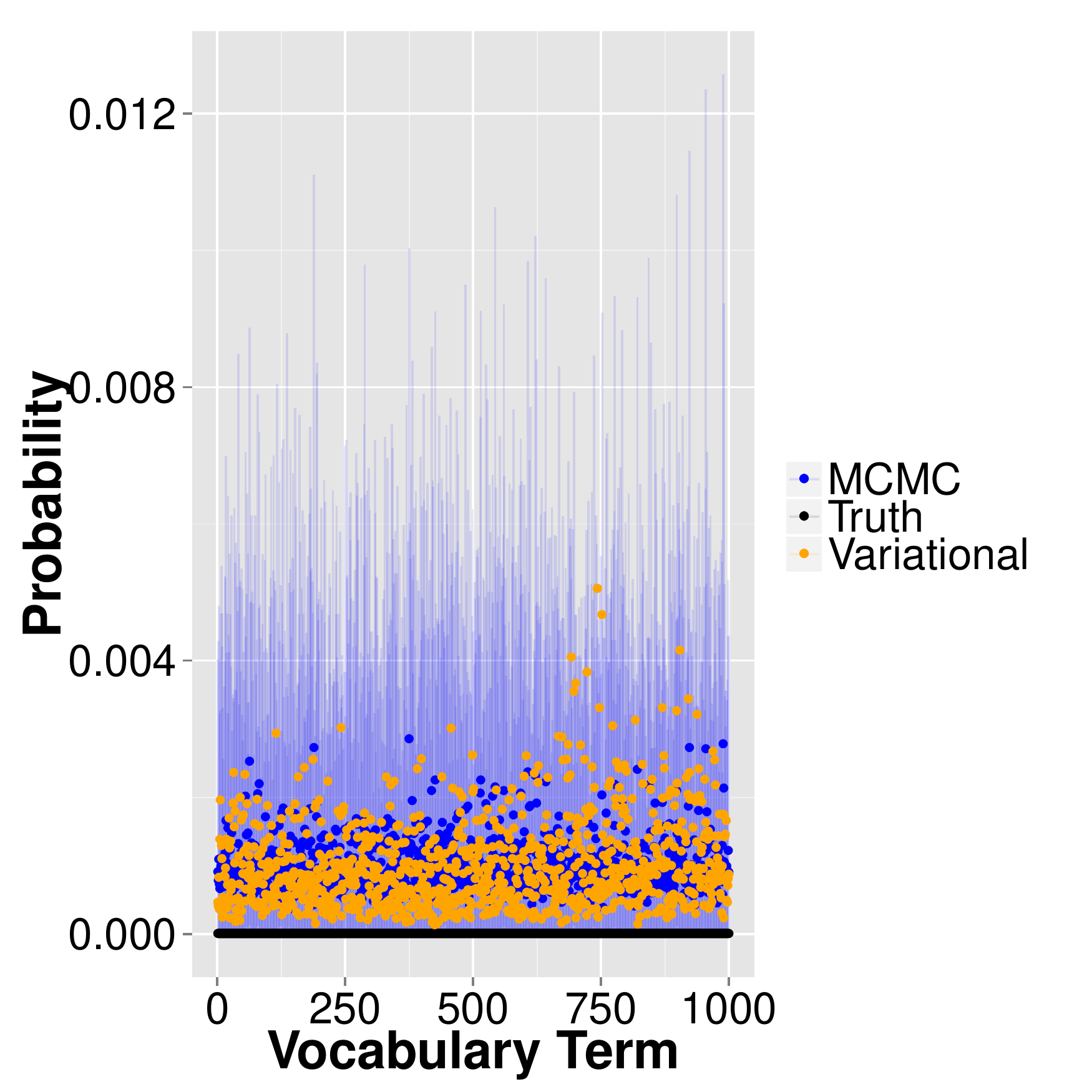}
  \caption{Topic 5}
\end{subfigure}
\begin{subfigure}{.3\textwidth}
  \centering
  \includegraphics[width=1\textwidth]{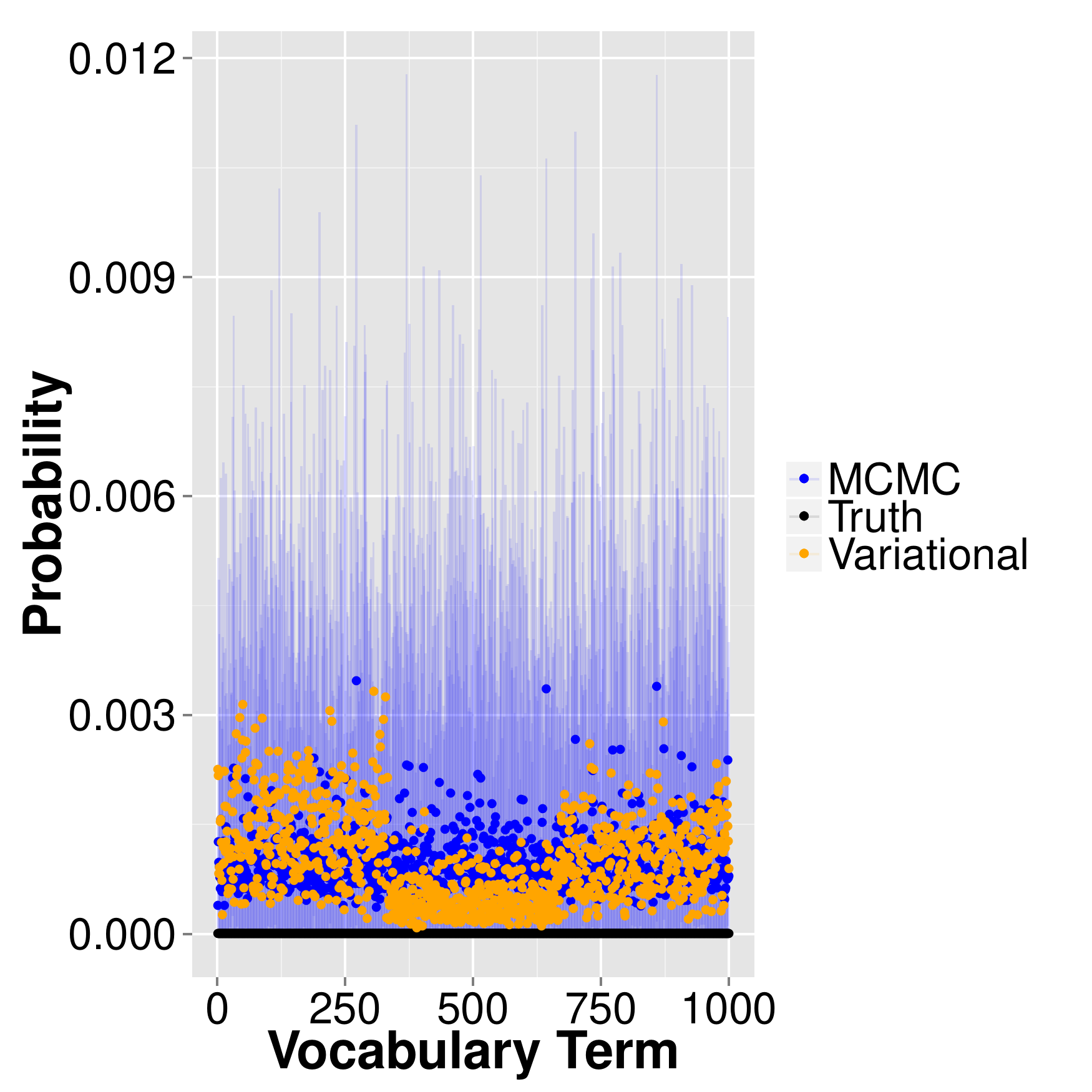}
  \caption{Topic 6}
\end{subfigure}
\end{figure}

Figure \ref{fig:Beta_Post_K6_1} presents the posterior means for Topics 1-3 and 4-6, respectivly.  Topics 1,2 and 3 correspond to the three topics which generated the synthetic corpus.  Topics 4,5, and 6 are extraneous.

\begin{figure}[h!]
\centering
\caption{Left: Total Variation distance between estimated vocabulary distributions (i.e. topics) and truth.  Middle: Total Variation distances between estimated document-specific topic proportions and true topic proportions for all documents at time $t=5$. Right: Marginal probability of topics over time.}
\label{fig:Topic_Post_Mean_LL}
\begin{subfigure}{.3\textwidth}
  \centering
  \includegraphics[width=1\textwidth]{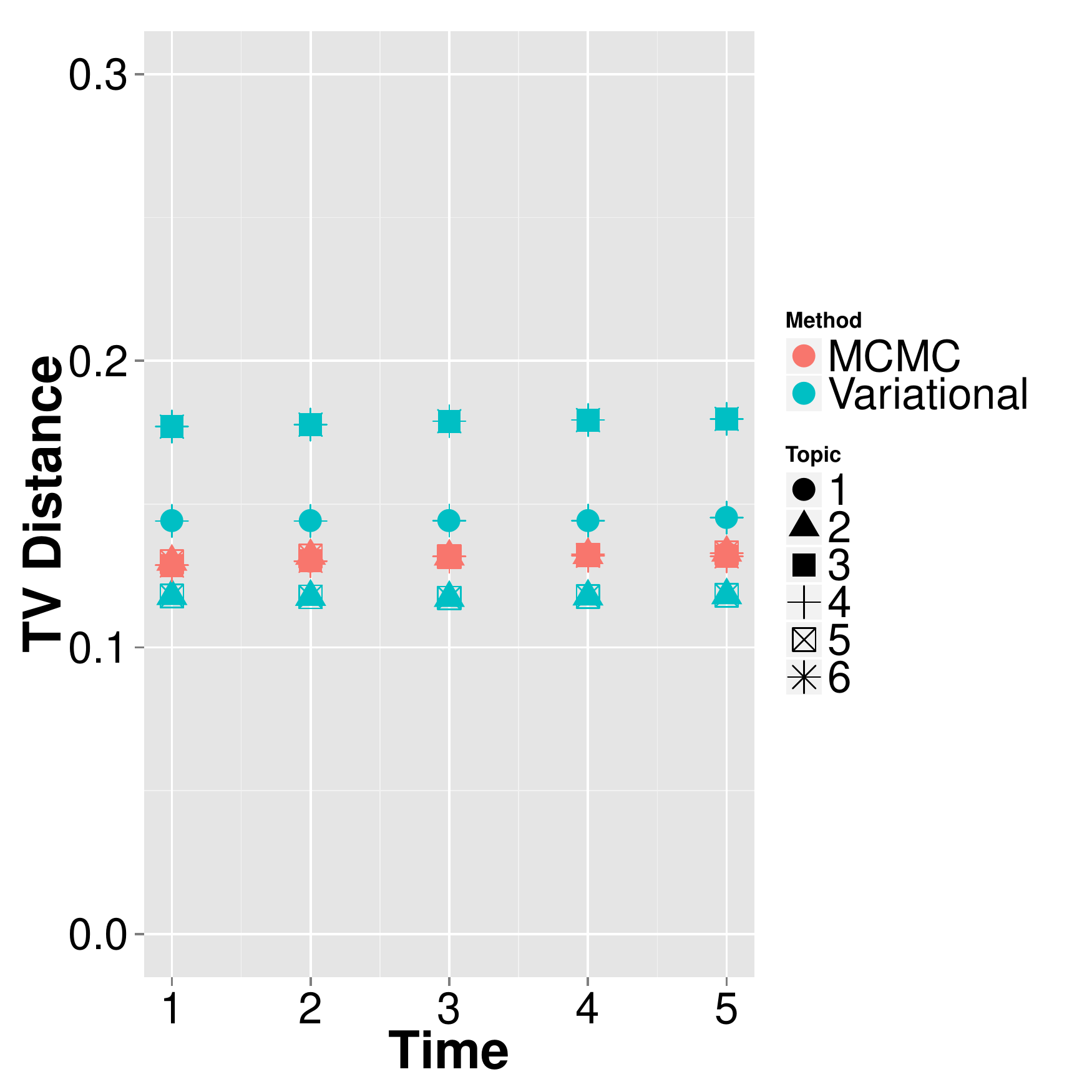}
  \caption{TV distances}
  \label{subfig:Eta_K6}
\end{subfigure}
\begin{subfigure}{.3\textwidth}
  \centering
  \includegraphics[width=1\textwidth]{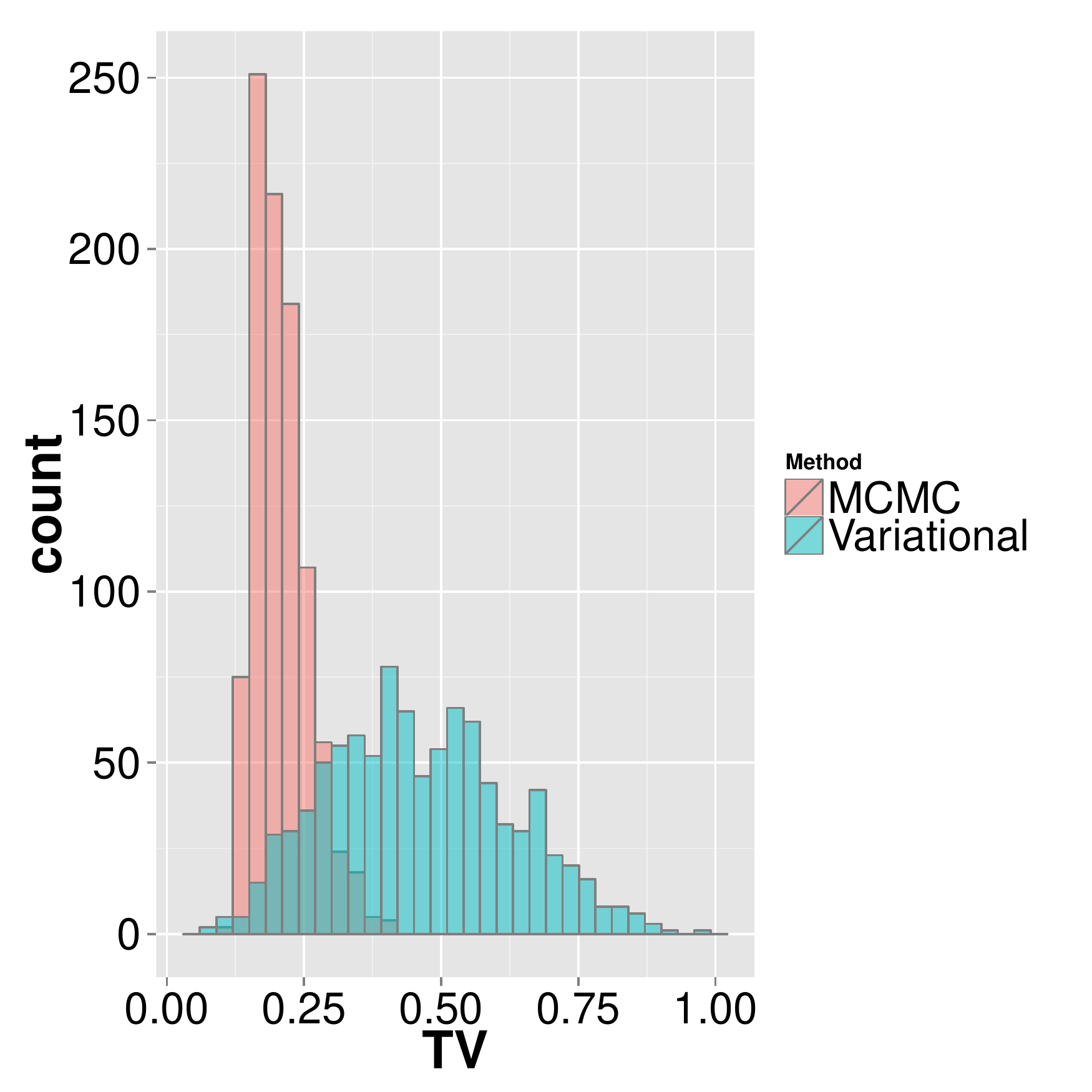}
  \caption{TV distances}
  \label{subfig:Eta_K6}
\end{subfigure}
\begin{subfigure}{.3\textwidth}
  \centering
  \includegraphics[width=1\textwidth]{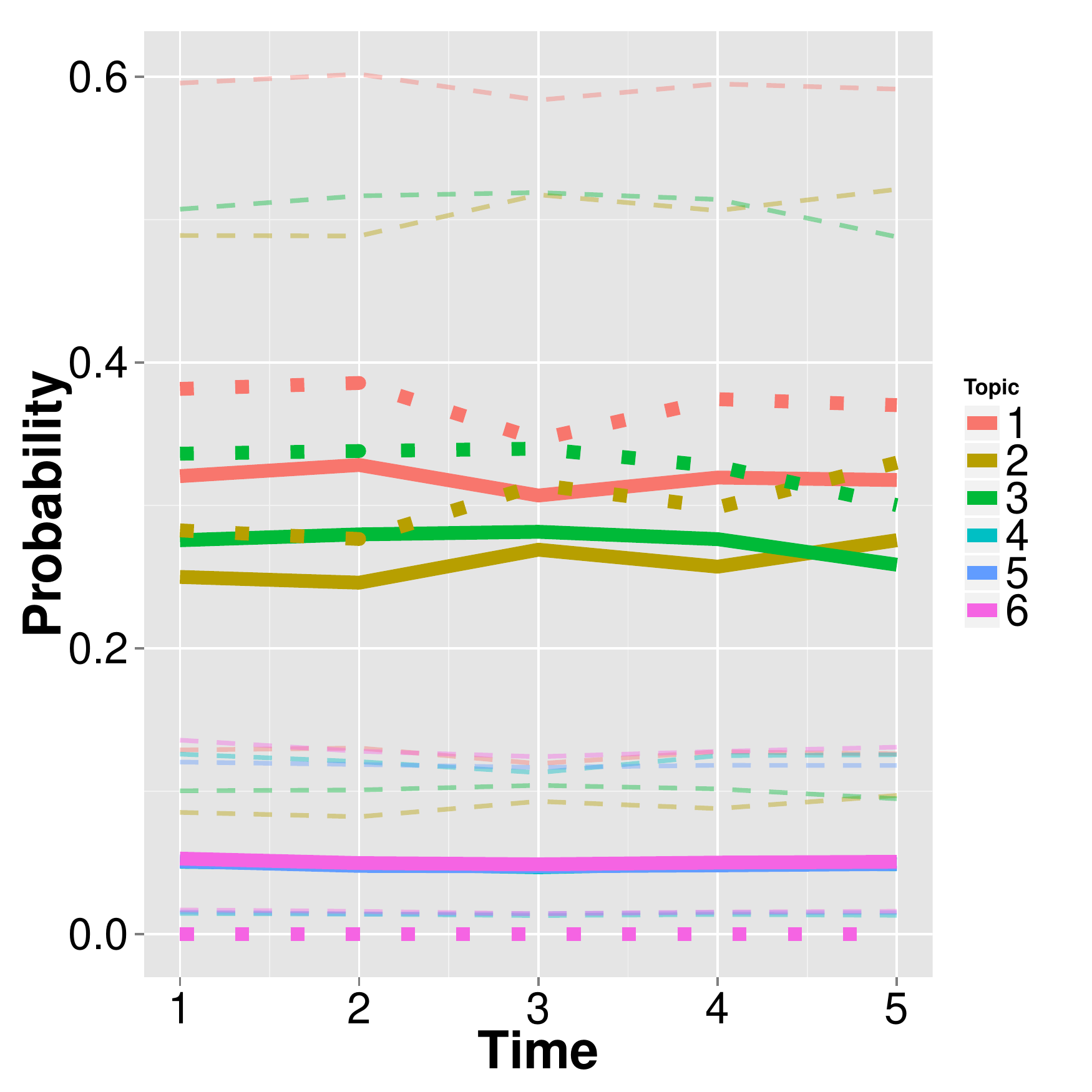}
  \caption{Topic Proportions}
  \label{subfig:Topic_Proportions_K6}
\end{subfigure}%
\end{figure}

The extremely wide uncertainty intervals for each vocabulary term in topics 4-6 suggest that there is no meaningful thematic differentiation amongst these topics.  Taken together, these figures demonstrate that even if the researcher misspecifies the number of topics to be larger than the true number, the DLTM will still identify the three topics actually present in the data.  The uncertainty intervals are valuable for identifying that a topic lacks a distinct theme.  The variational DTM also recovers the three original topics.  The advantage of the DLTM is in estimating the document-specific topic proportions.  By sharing information across documents, the DLTM infers which topics are unnecessary to model the corpus as a whole.  This is evidenced by the global proportions of topics 4,5, and 6 in Figure \ref{subfig:Topic_Proportions_K6}.  In the previous simulation study, the uncertainty intervals around the topic proportions was quite wide.  Figure \ref{subfig:Topic_Proportions_K6} demonstates that the posterior proportions for topics 4,5, and 6 concentrates around low probabilities.  The lack of thematic differentiation and the low marginal probabilities indicate that topics 4, 5, and 6 are extraneous.  The advantage of identifying these globally unnecessary topics is demonstrated by Figure \ref{subfig:Eta_K6}.  The DLTM dramatically outperforms the variational approximation of the DTM when estimating document topic proportions.  Borrowing information across documents is essential when the number of topics is misspecified.

\subsection{Linear, Quadratic, and Harmonic Trends}
\label{subsec:Complex_Trends}
In this section, we consider simulation examples with linear, quadratic, and harmonic trends.  We maintain the same topics as in previous examples but allow their marginal probabilities in the corpus to exhibit more complex dynamic behavior.  In the linear case, $F_{k,t} = \begin{bmatrix} 1 & 0 \\ \vdots & \vdots \\ 1 & 0 \end{bmatrix}$.    The system matrix is $G_{k,t} = \begin{bmatrix} 1 & 1 \\ 0 & 1 \end{bmatrix}$.  The harmonic case can be simulated by maintaining the $F_{k,t}$ matrix as in the linear case and replacing the system matrix with $G_{k,t} = \begin{bmatrix} cos(\omega) & sin(\omega) \\ -sin(\omega) & cos(\omega) \end{bmatrix}$.  We let $\omega = \frac{\pi}{2}$ in our example.  To simulate quadratic trends, $F_{k,t} = \begin{bmatrix} 1 & 0 & 0\\ \vdots & \vdots & \vdots \\ 1 & 0 & 0 \end{bmatrix}$ and $G_{k,t} = \begin{bmatrix} 1 & 1 & 1 \\ 0 & 1 & 1  \\ 0 & 0 & 1 \end{bmatrix}$. 

\begin{figure}[h!]
\centering
\caption{In this figure, the variational estimates are from the DTM.  The MCMC estimates are from the DLTM.  First row: Total Variation distance between estimated vocabulary distributions (i.e. topics) and truth.  Second row: Total Variation distances between estimated document-specific topic proportions and true topic proportions for all documents at time $t=5$. Third row: Marginal probability of topics over time.  The solid lines are the posterior means.  The light dashed lines are the posterior 95\% credible intervals.  The thick dashed lines are the true trends.}
\label{fig:Topic_Trend_Post}
\begin{subfigure}{.3\textwidth}
  \centering
  \includegraphics[width=1\textwidth]{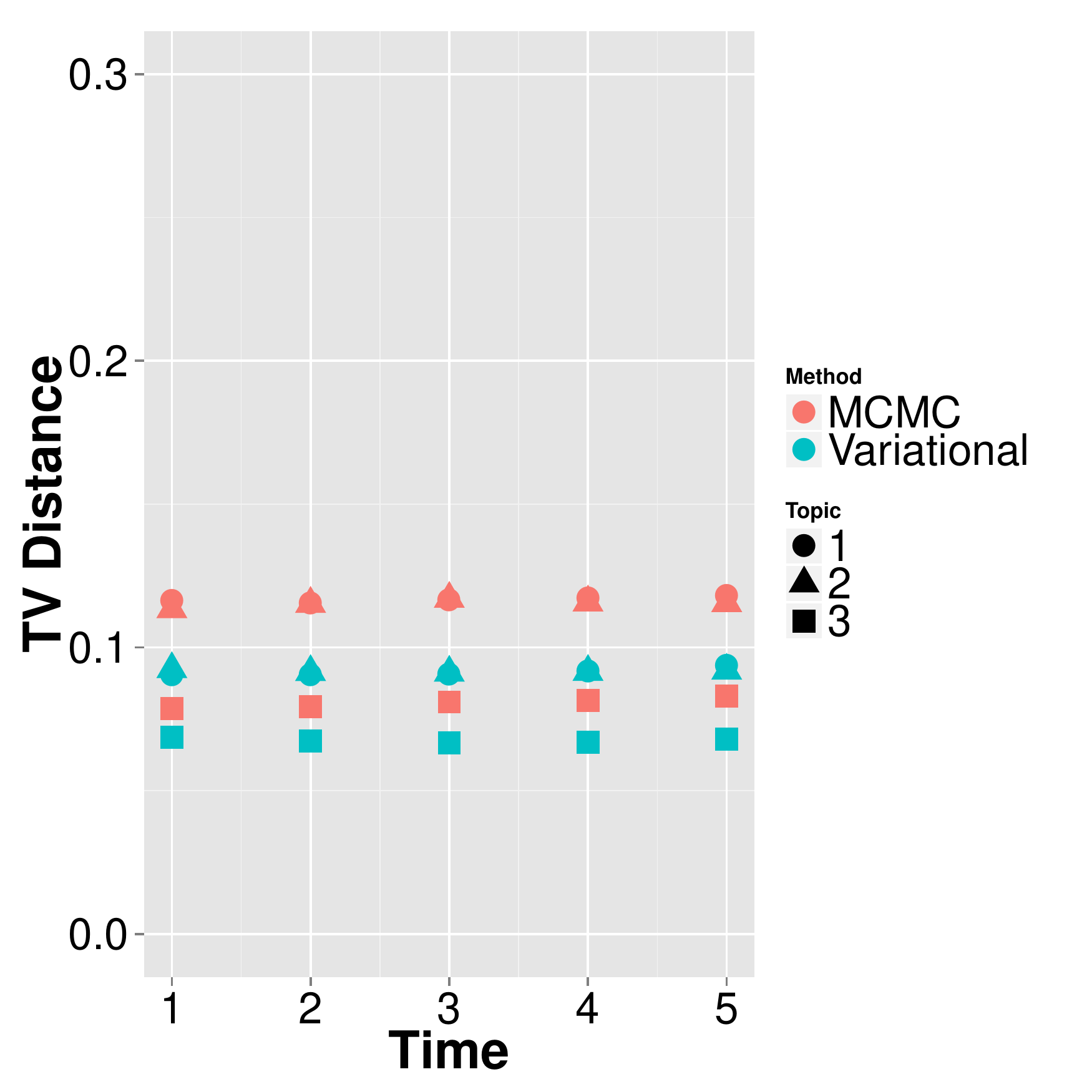}
  \includegraphics[width=1\textwidth]{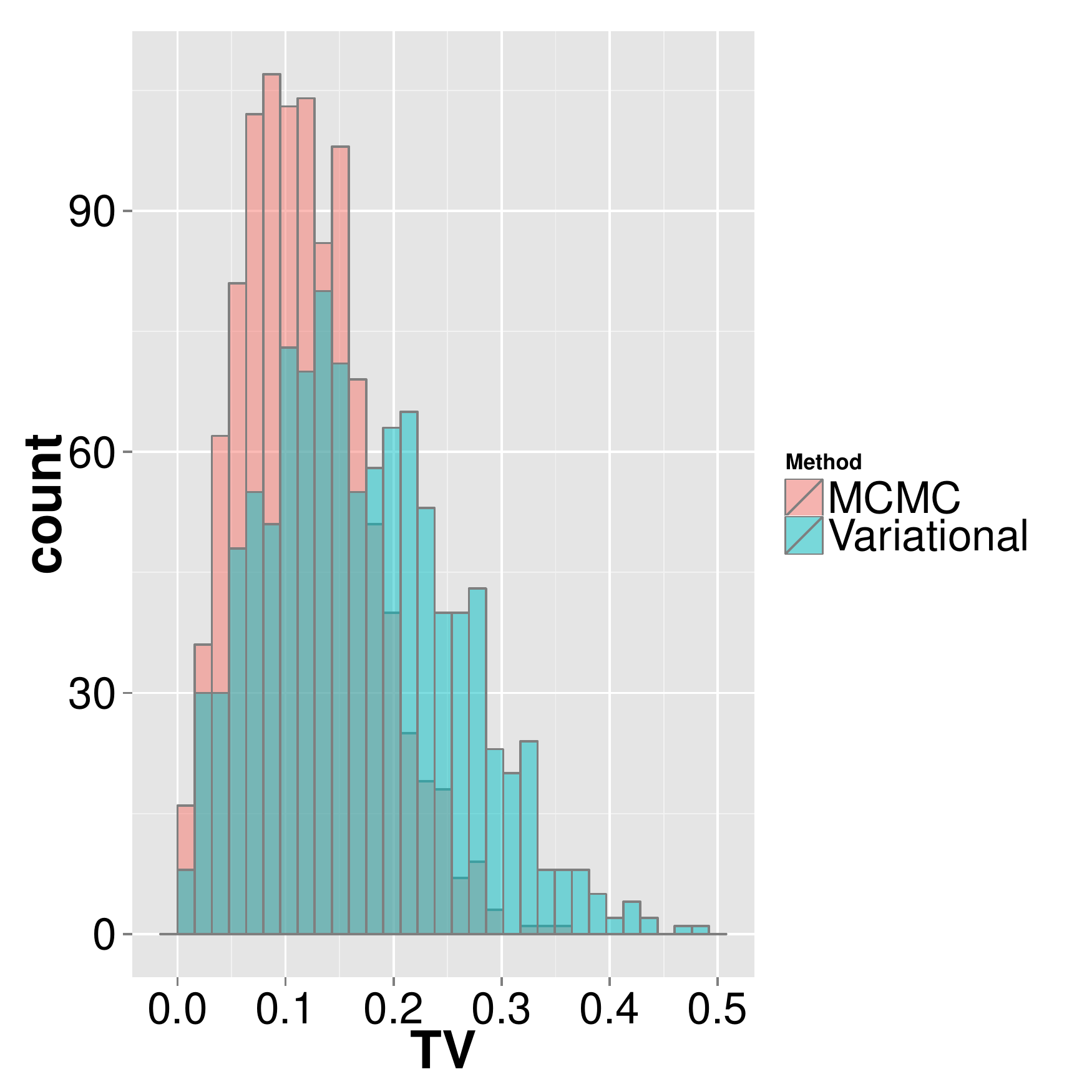}
  \includegraphics[width=1\textwidth]{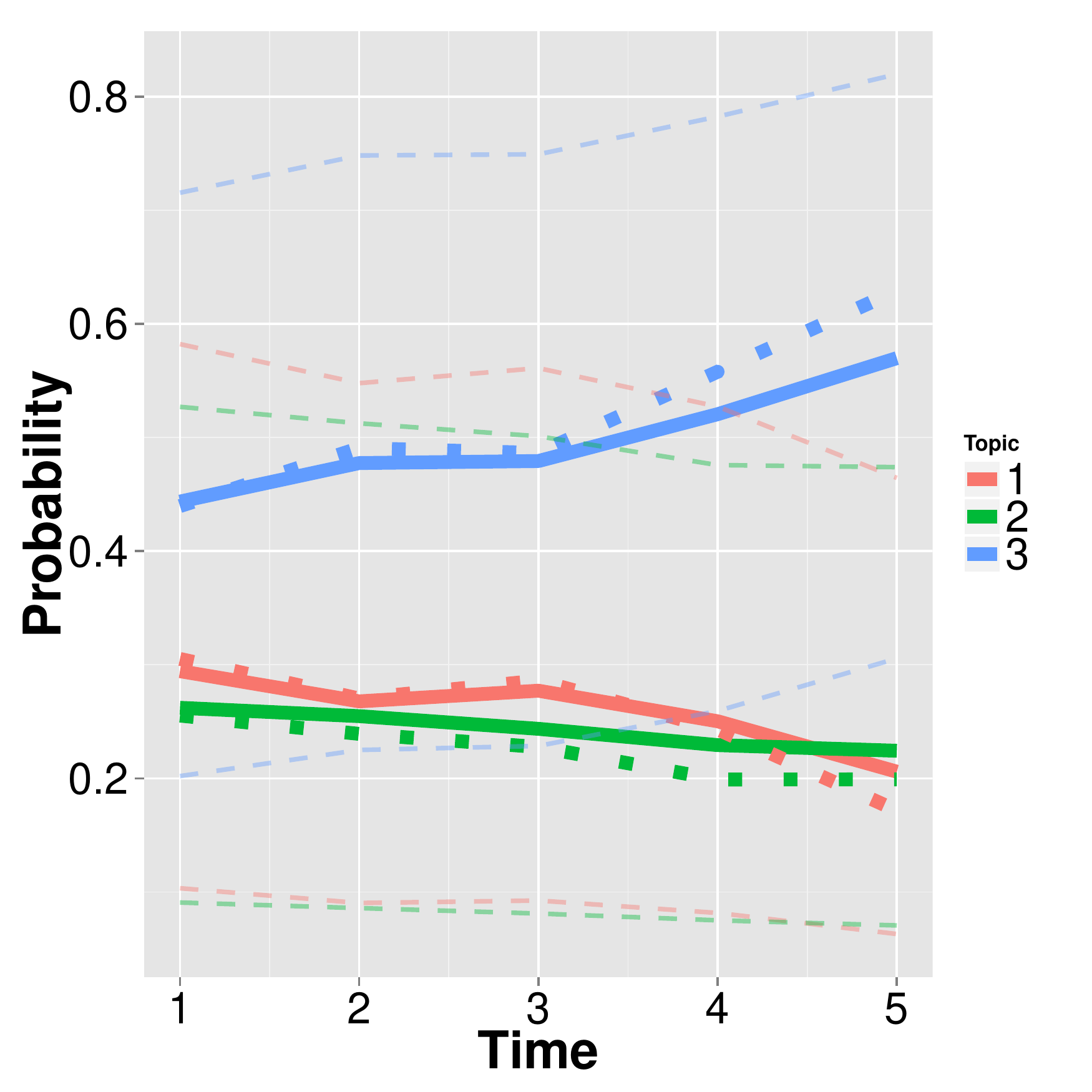}
  \caption{Linear Trend}
  \label{subfig:Linear_Trend}
\end{subfigure}
\begin{subfigure}{.3\textwidth}
  \centering
  \includegraphics[width=1\textwidth]{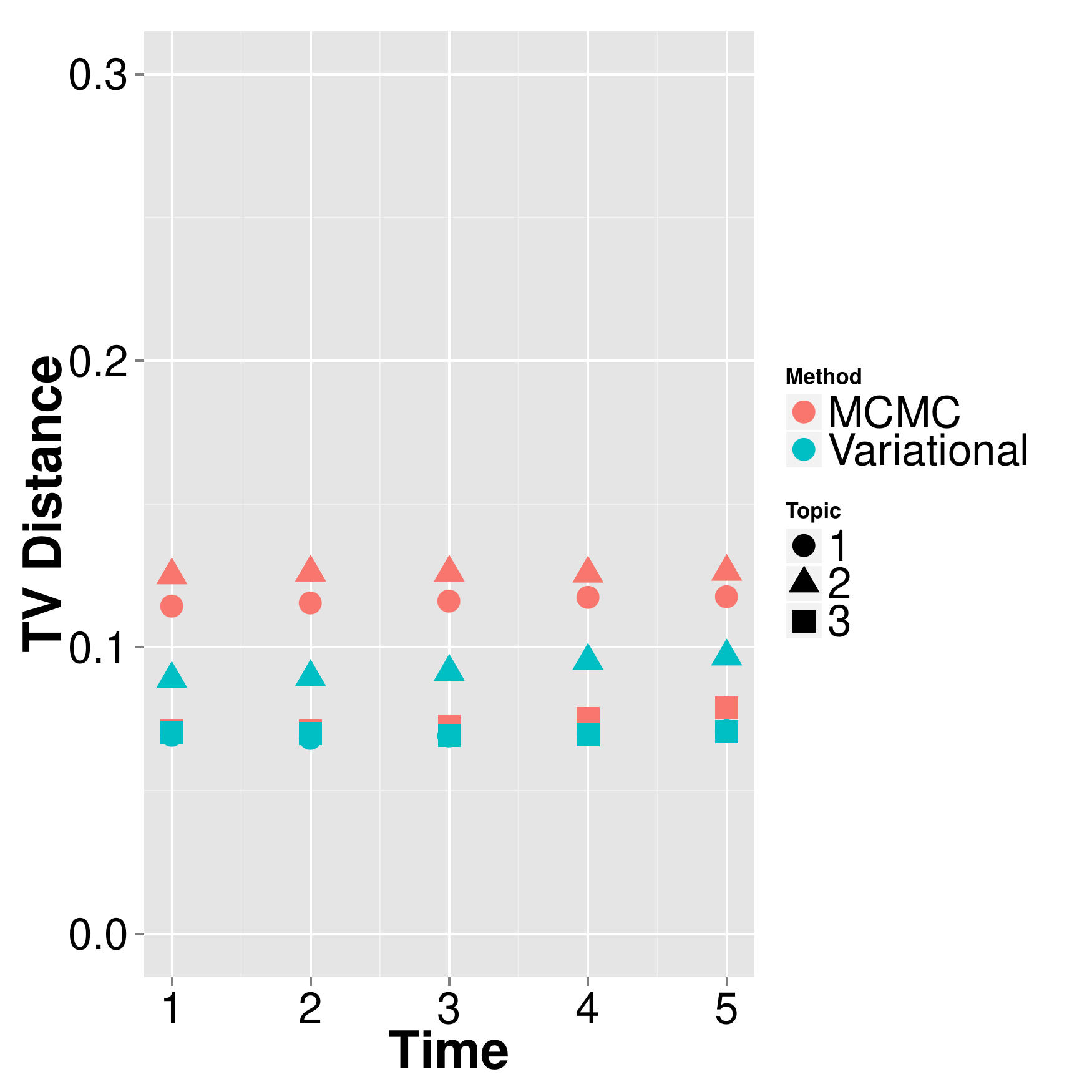}
  \includegraphics[width=1\textwidth]{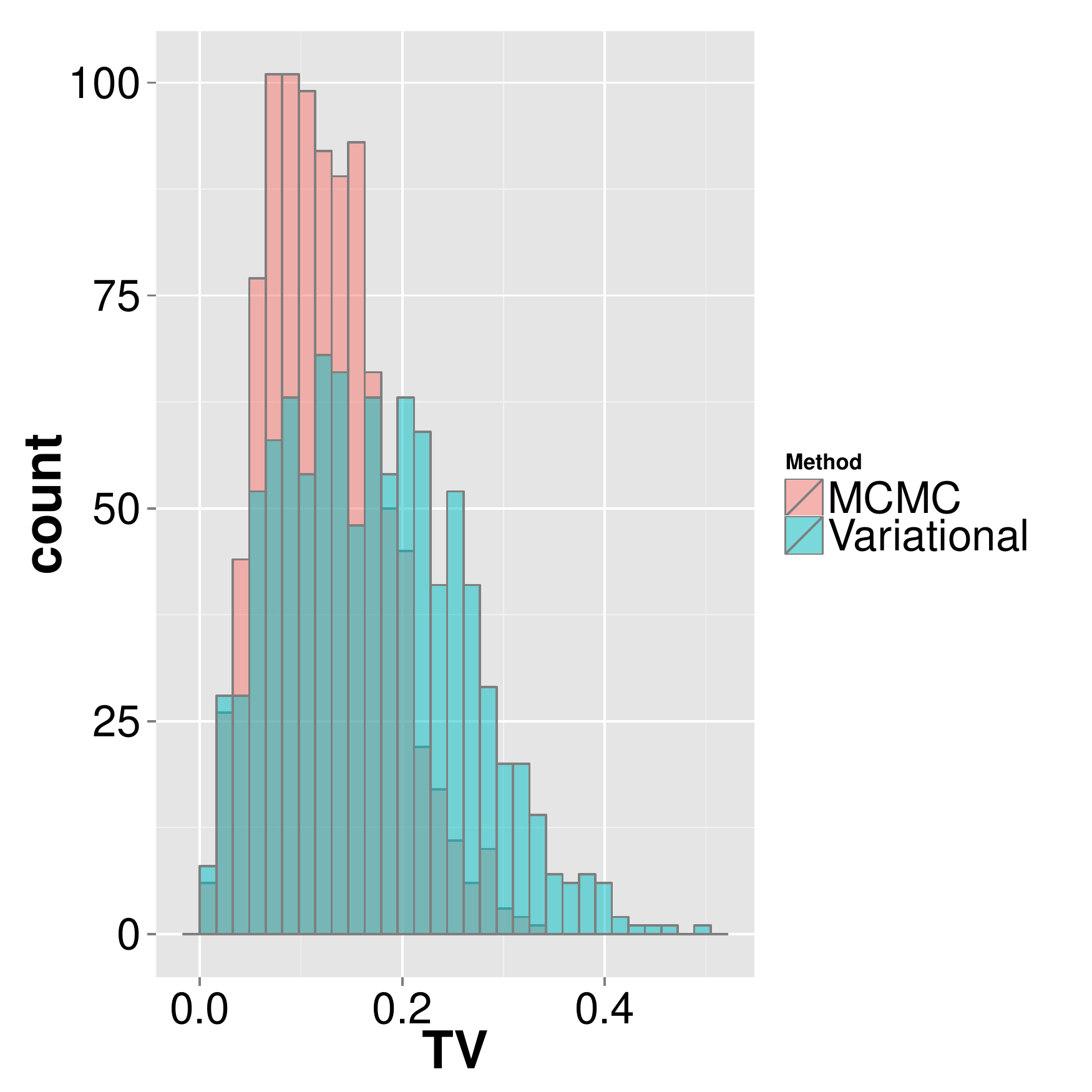}
  \includegraphics[width=1\textwidth]{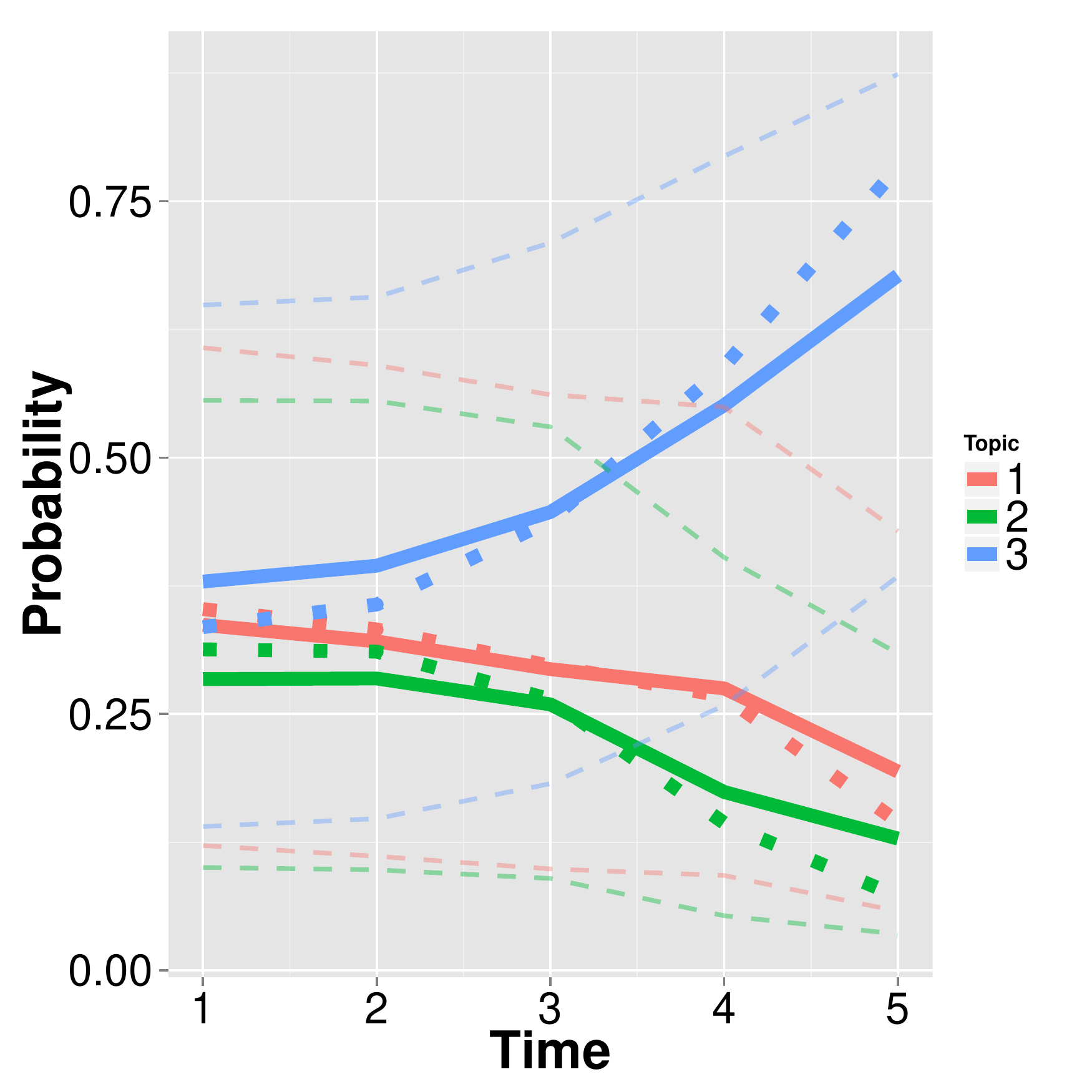}
  \caption{Quadratic Trend}
  \label{subfig:Quadratic_Trend}
\end{subfigure}
\begin{subfigure}{.3\textwidth}
  \centering
  \includegraphics[width=1\textwidth]{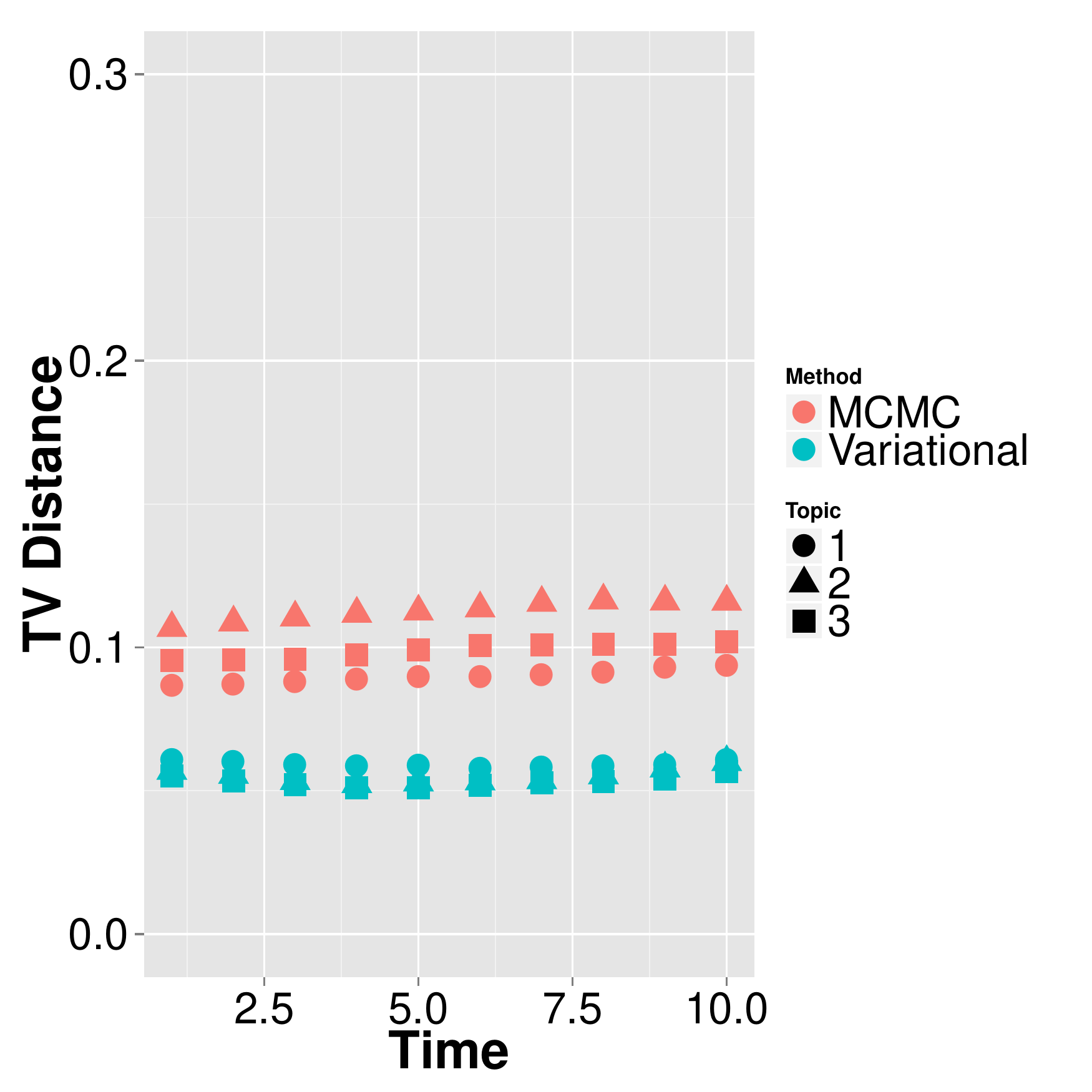}
  \includegraphics[width=1\textwidth]{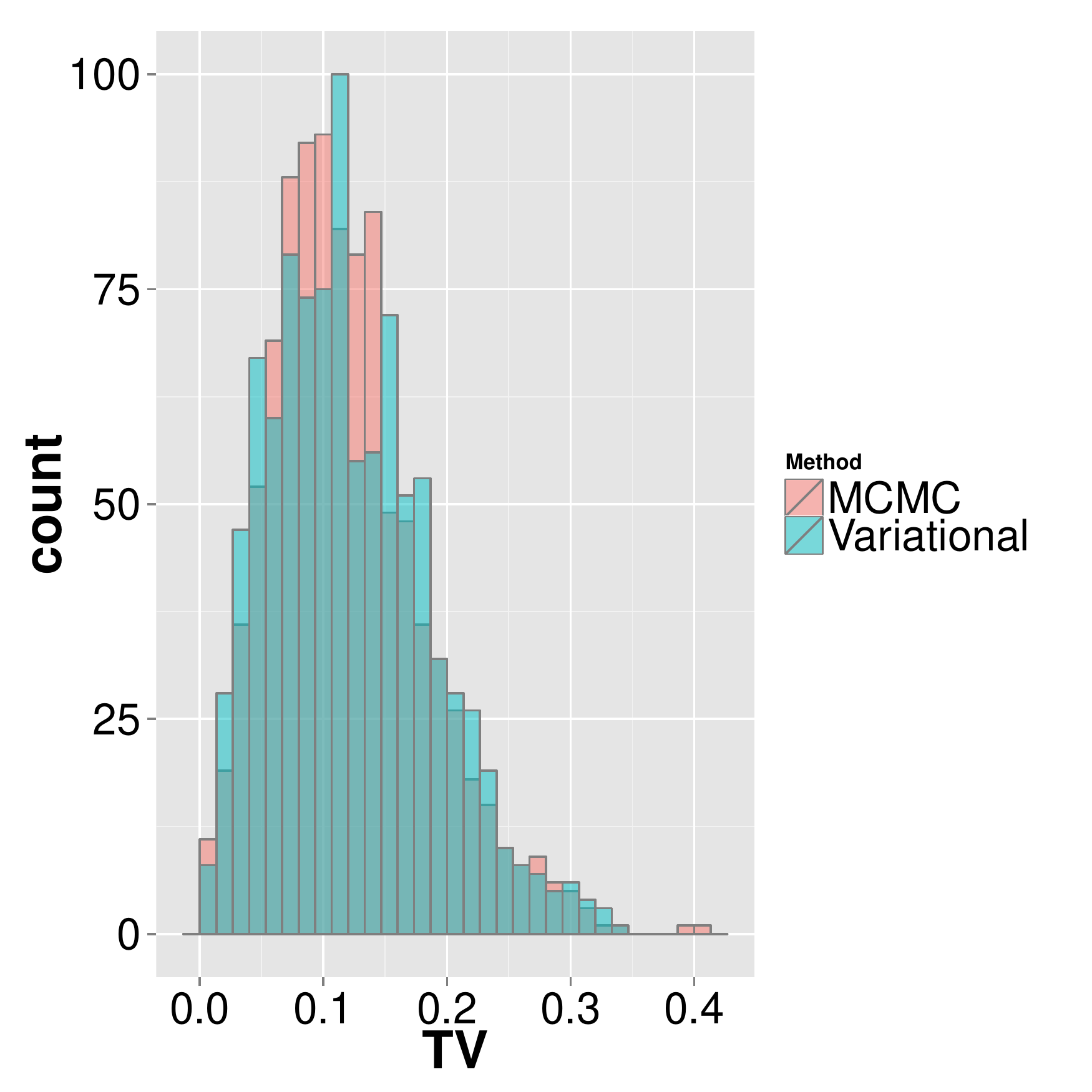}
  \includegraphics[width=1\textwidth]{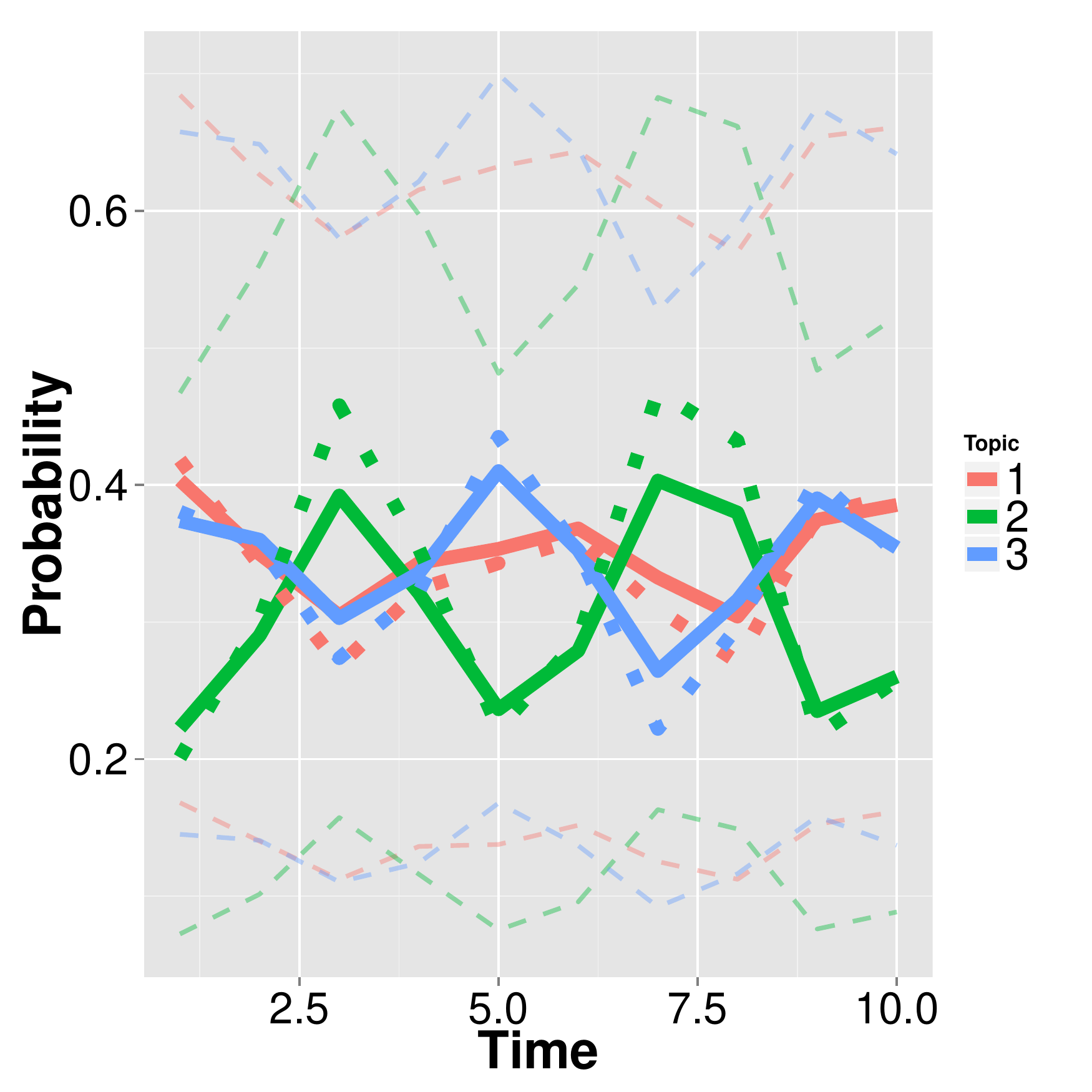}
  \caption{Harmonic Trend}
  \label{subfig:Harmonic_Trend}
\end{subfigure}%
\end{figure}

The topics recovered from these analyses are presented in Appendix \ref{App:DLTM_Complex}.  Variational inference for recovering topics slightly outperforms the MCMC and DLTM.  The outperformance is summarized in the first row of Figure \ref{fig:Topic_Trend_Post}.  In the second row, the DLTM and MCMC outperform the variational DTM in estimating document-specific topic proportions.  This is especially true in the linear and quadratic cases.  In the third row, the the marginal probabilities of topics are compared to the true simulated trend.  The thick solid lines are the posterior means of the trends.  The thick dashed lines are the true marginal probabilities.  The thin dashed lines are the 95\% credible intervals of the posterior distribution for each topic's marginal probability.  The DLTM nicely recovers the simulated trends.

\begin{figure}[h!]
\centering
\caption{First row: One-step-ahead predictions for a DLTM with trend (thick solid line) and the random walk DTM (thin solid line) compared with the truth (thick dashed line).  Light dashed lines are the 95\% credible intervals for the DLTM estimate.  Second row: prediction errors for the DLTM and DTM}
\label{fig:Prediction}
\begin{subfigure}{.25\textwidth}
  \centering
  \includegraphics[width=1\textwidth]{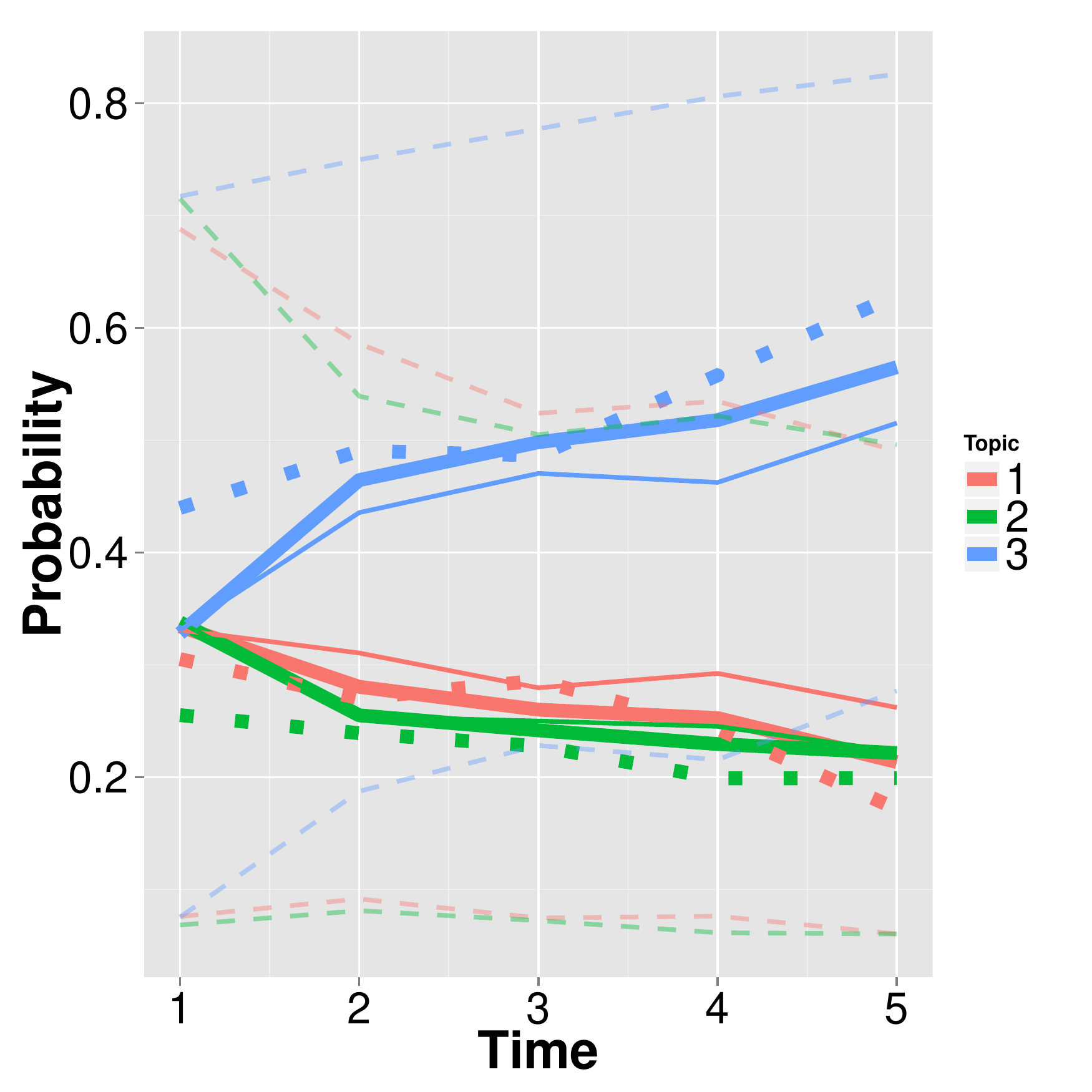}
    \includegraphics[width=1\textwidth]{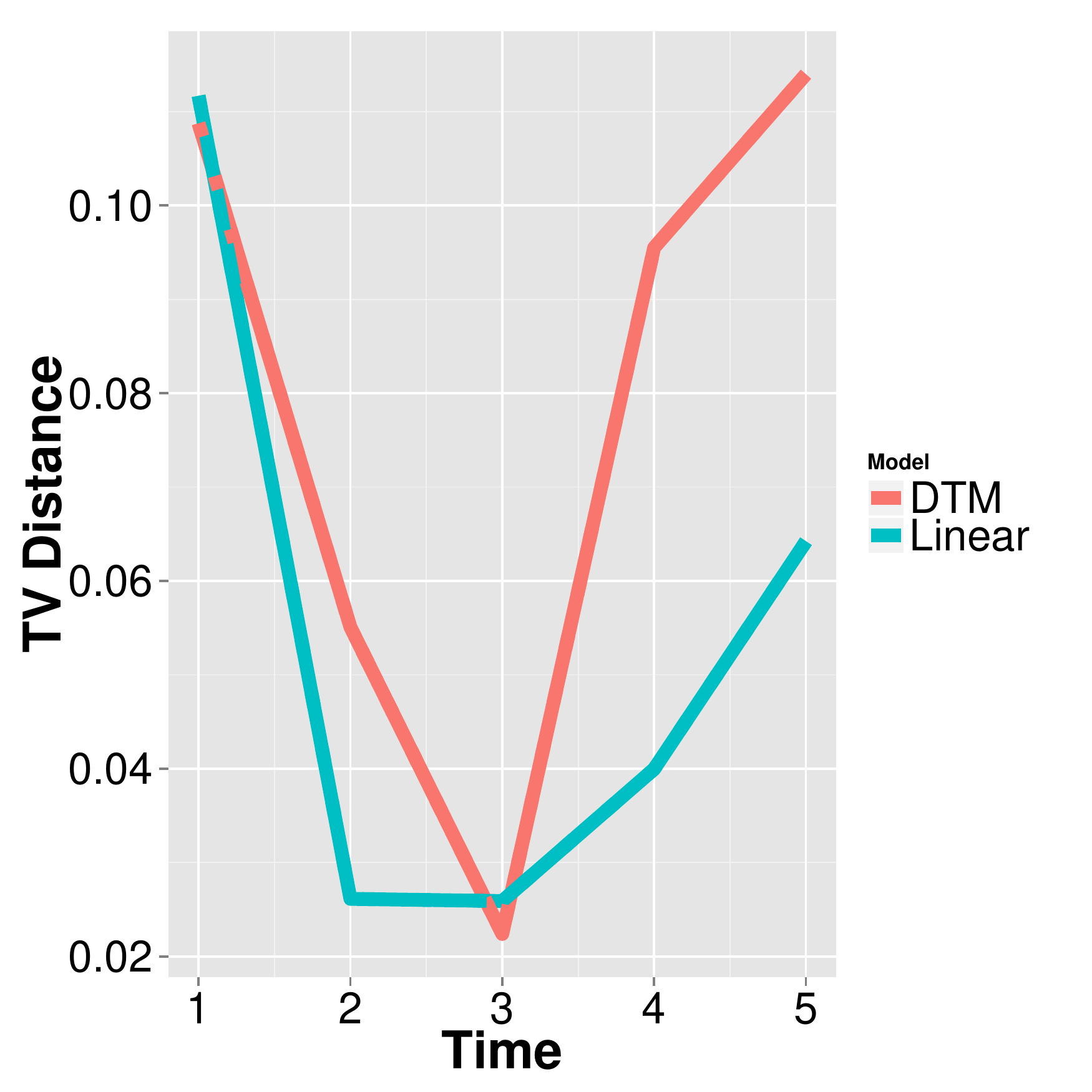}
  \caption{Linear Trend}
  \label{subfig:Linear_Prediction}
\end{subfigure}%
\begin{subfigure}{.25\textwidth}
  \centering
  \includegraphics[width=1\textwidth]{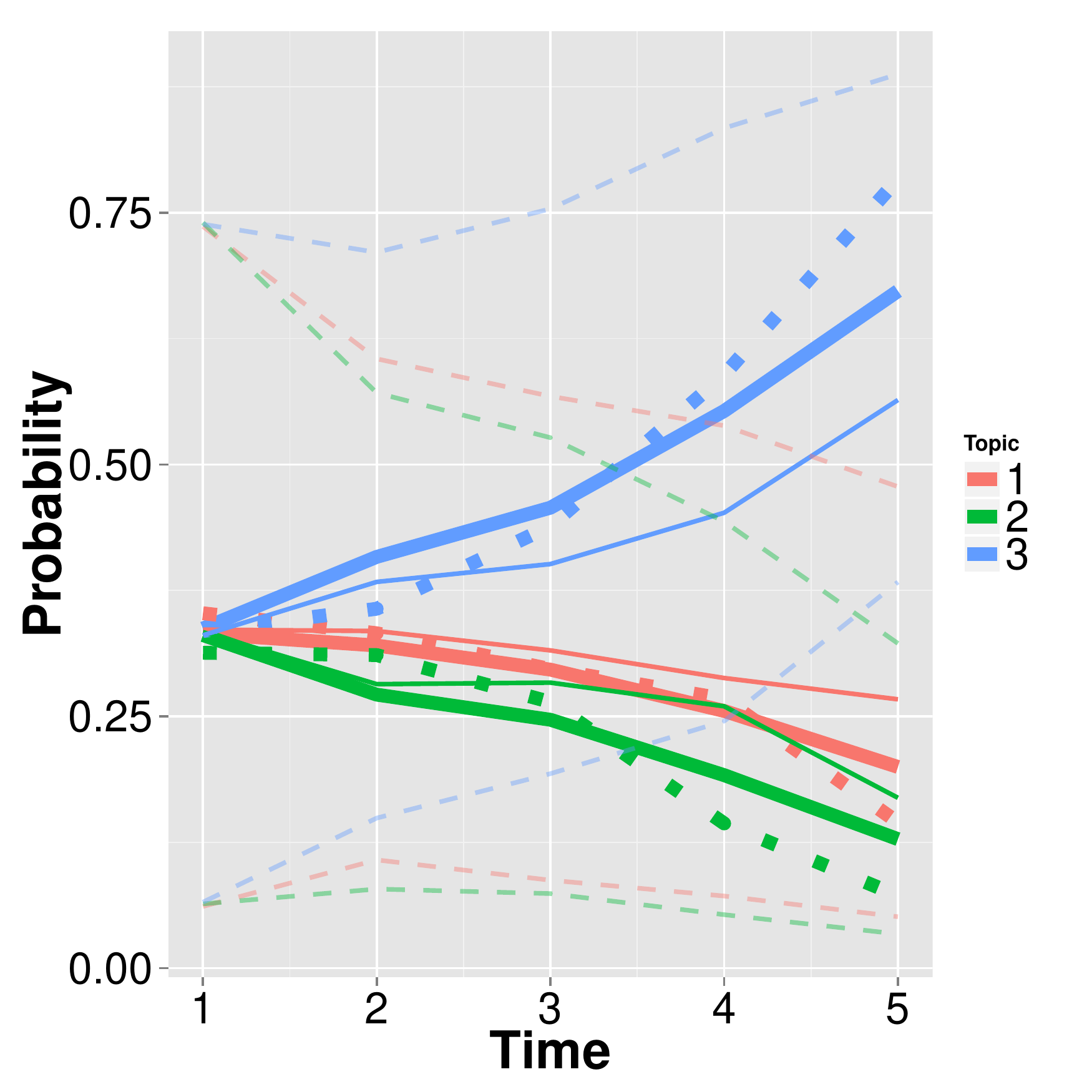}
  \includegraphics[width=1\textwidth]{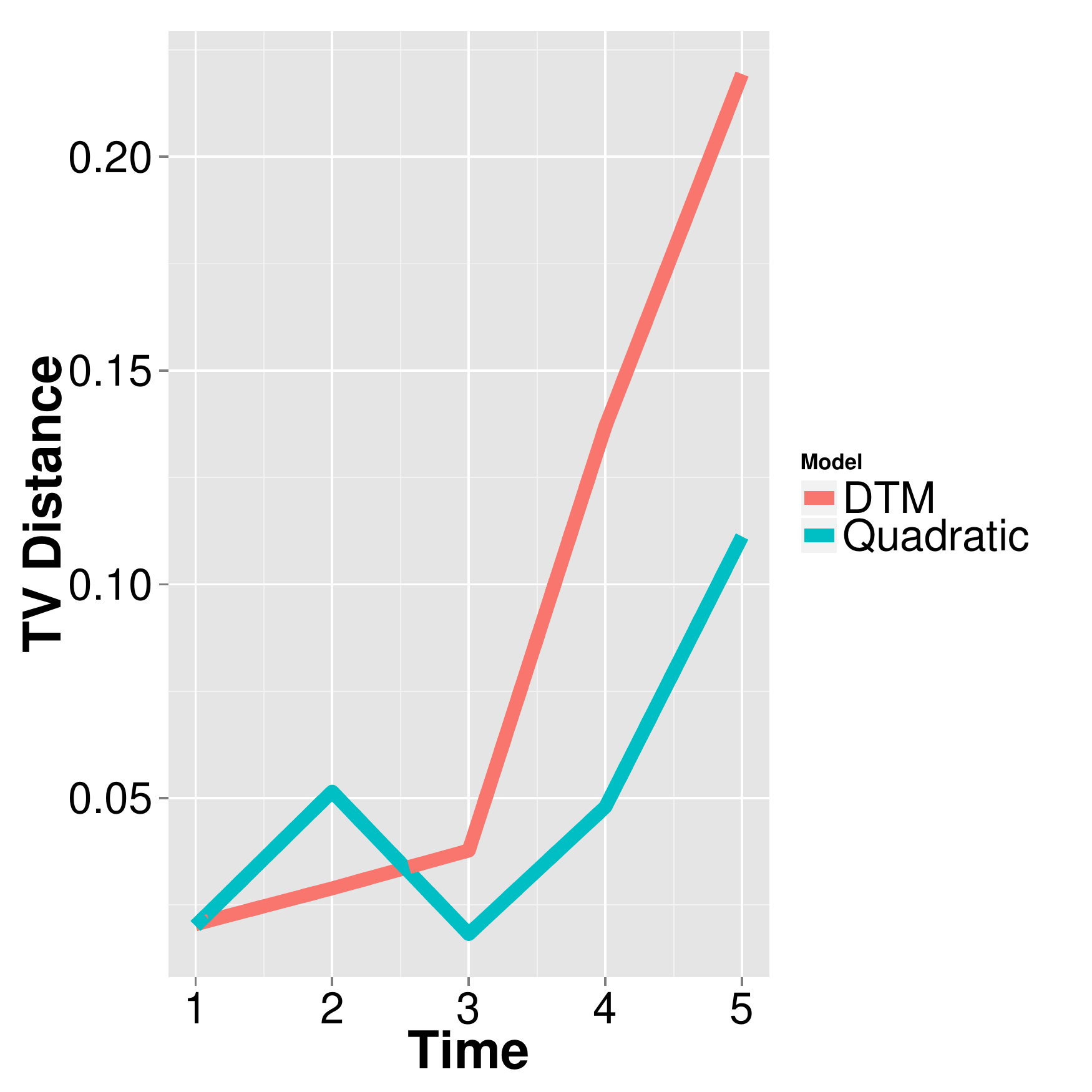}
  \caption{Quadratic Trend }
  \label{subfig:Quadratic_Prediction}
\end{subfigure}
\begin{subfigure}{.25\textwidth}
  \centering
  \includegraphics[width=1\textwidth]{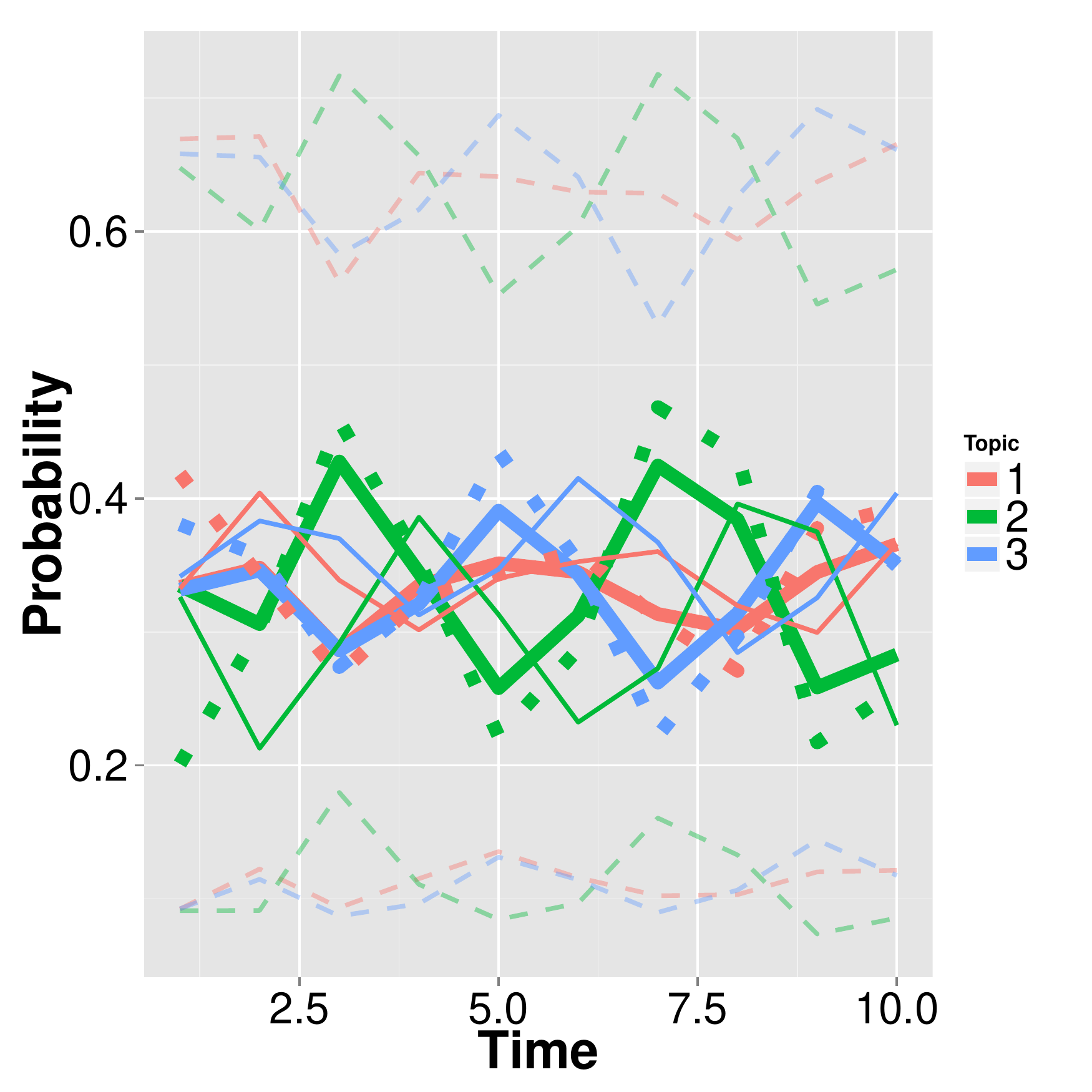}
  \includegraphics[width=1\textwidth]{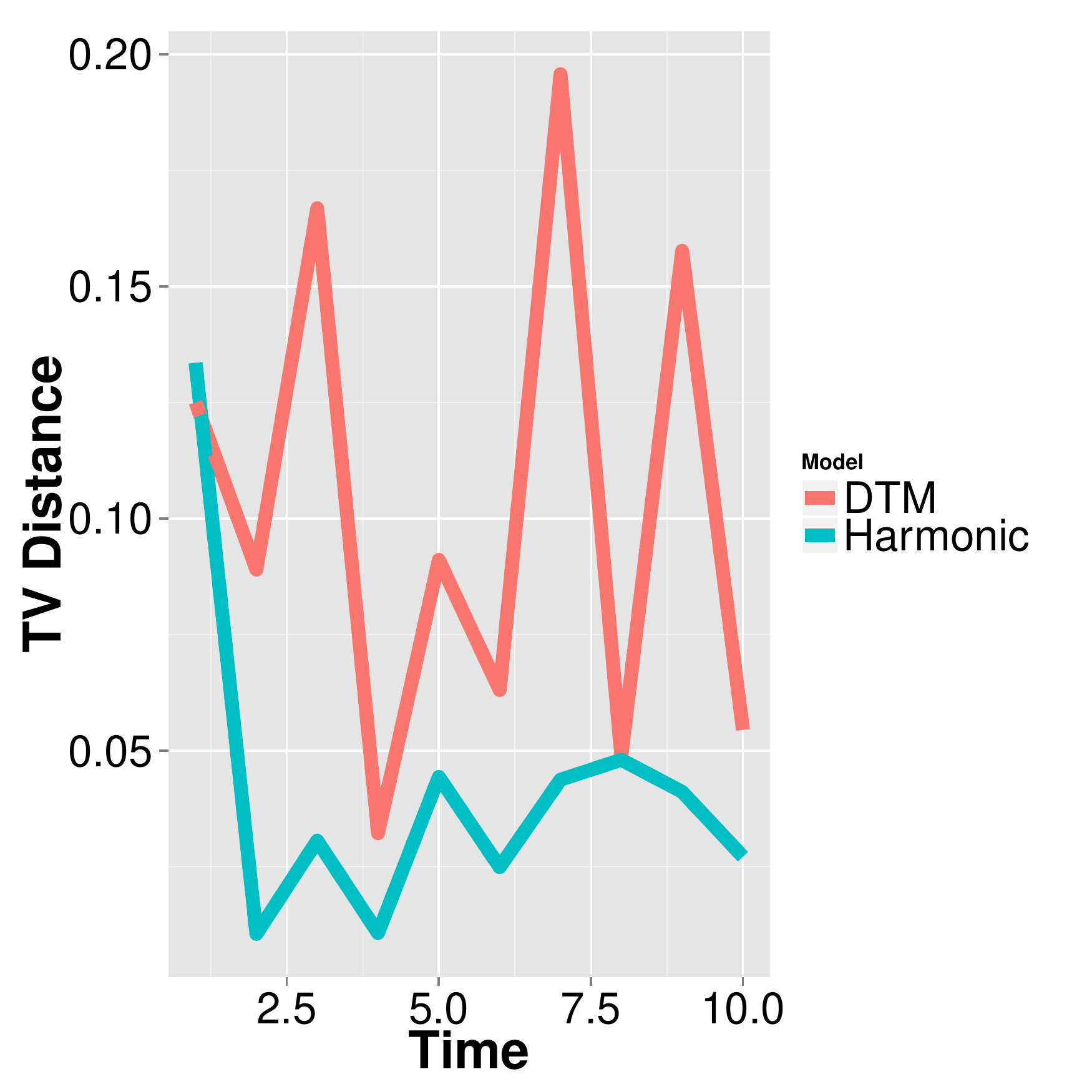}
  \caption{Harmonic Trend }
  \label{subfig:Harmonic_Prediction}
\end{subfigure}
\end{figure}

Explicitly incorporating trend components is important for prediction as well as inference.  Figure \ref{fig:Prediction} presents the one-step-ahead predictions from the DLTM and the MCMC implementation of the DTM.  The DLTM outperforms the DTM in the linear and quadratic cases as trends become more established.  This outperformance is significant by the final time point.  The outperformance of the DLTM in the harmonic case is significant throughout.  The DTM over smoothes the data, and this results in forecasts that oscillate with a phase shift from the truth.  

\section{PubMed Abstracts}
\label{sec:CaseStudy}

One motivation for adding dynamic structure in topic trends is to model the rapid increase and decrease of a topic's prevalence in scientific literature.  Figure \ref{subfig:ML_Trends} demonstrates the exponentially fast emergence and vanishing of topics Autistm and Polio in the biomedical literature.  We model this rapid growth and decay of topic probabilities by endowing the forecast function for $\eta_{k,t}$ with a locally linear time-trend.  Chapter 7 of \citet{westharrison} demonstrates that this local linear trend can be achieved by letting each row of the $F_{k,t}$ matrix be $\begin{bmatrix} 1 & 0 \end{bmatrix}$ and specifying $G_{k,t} = \begin{bmatrix} 1 & 1 \\ 0 & 1 \end{bmatrix}$.

We apply this locally linear trend model to a corpus of abstracts from PubMed.  PubMed is a free web-based search tool allowing users access to over 25 million references and abstracts in the life science and biomedical literature.  Our corpus is a random sample of 8,009 PubMed abstracts spanning 30 years from 1985-2014.  There are several steps to pre-processing the data: stemming, vocabulary exclusion, and document exclusion.  

Stemming is a procedure which reduces a word to its most basic root.  As an example, running and runs are truncated to the root word run.  All words in the corpus were stemmed using the Snowball algorithm \citep{SnowballC}.  

The vocabulary for our analysis was selected by excluding words based on three separate criteria.  First, stop words including the, and, but, and so were removed.  Second, all words which were observed fewer than 25 times across all years and fewer than 10 times within a specific year were removed.  One of the assumptions of the model is that the vocabulary is fixed over time.  We selected our vocabulary to meet that assumption.  Third, we manually removed most adjectives and adverbs.  Words that describe a degree or intensity of an object or action are not particularly useful for discerning semantic topics.  We retained a vocabulary of 307 terms.  

Abstracts that contained fewer than 20 occurrences of vocabulary terms were excluded from the analysis.  We chose the threshold of 20 to allow our Polya-Gamma approximation to be reliable.  For the 8,009 documents included in the analysis, the average document length is 27 words.    

We specify that there are $K=15$ topics in this corpus, which was chosen to accomodate the diversity of the topics in a subsample of documents from a large corpus like PubMed.  The number of topics in the corpus should increase with both the size of the vocabulary and the number of documents in the overall corpus.   

The MCMC simulation was run for $300,000$ samples with thinning of every $100^{th}$ sample.  We discarded $2,000$ of the remaining $3,000$ samples as burn-in.  Parameters and latent variables were initialized as in Section \ref{sec:SimStudy}.  Running this simulation took 25 hours on an 8 core workstation.   

\begin{figure}[h!]
\centering
\caption{The first row of figures represents the time-varying proportions of topics in the corpus.  The solid colored line and the dashed lines represented the posterior mean and 95\% credible interval for specific topics.  The remaining light gray lines are the posterior means for the remaining 14 topics.  The second row of figures represents the distribution over vocabulary terms corresponding to same topic as in the figure directly above. }
\label{fig:PubMed_Topics}
\begin{subfigure}{.2\textwidth}
  \centering
  \includegraphics[width=1\textwidth]{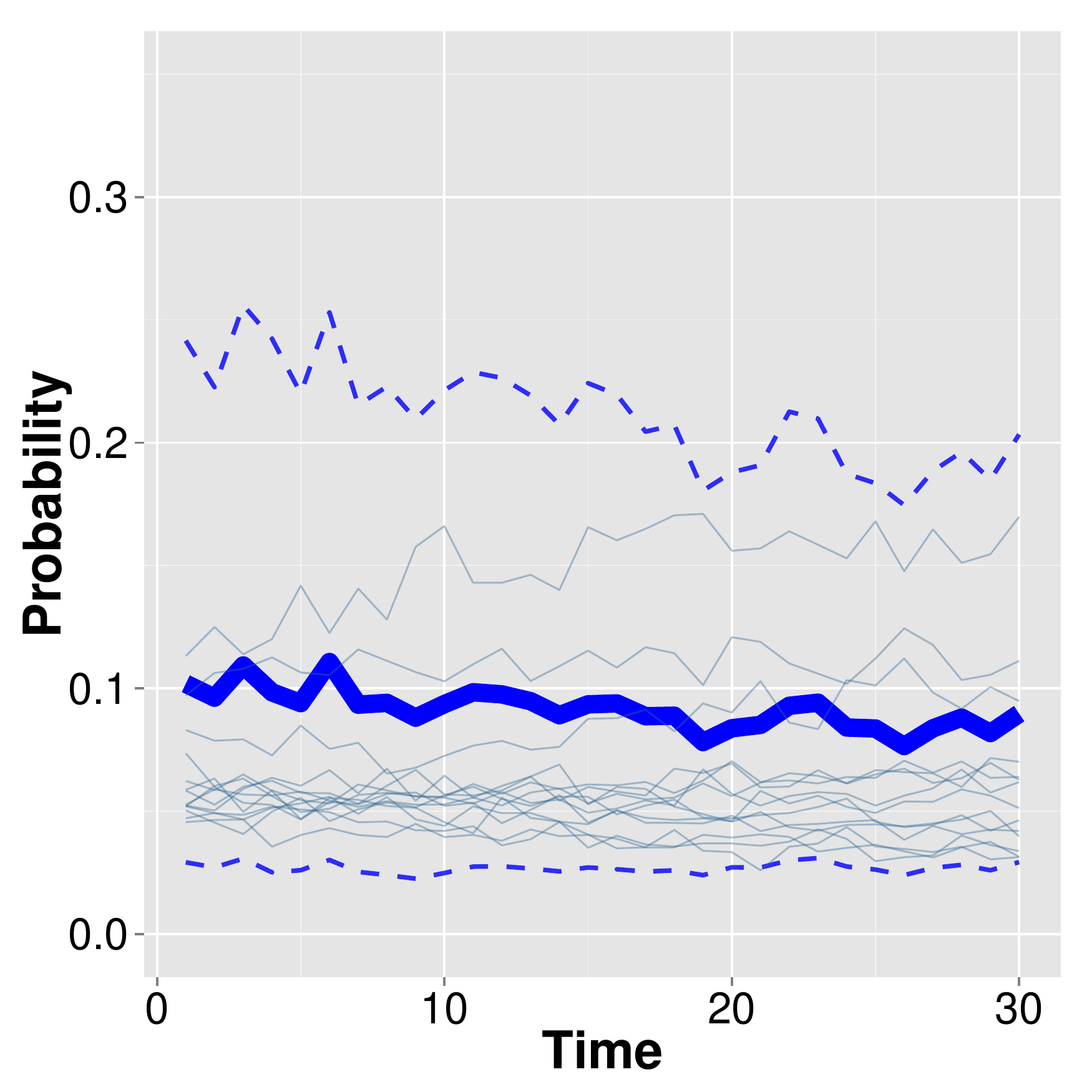}
    \includegraphics[width=1\textwidth]{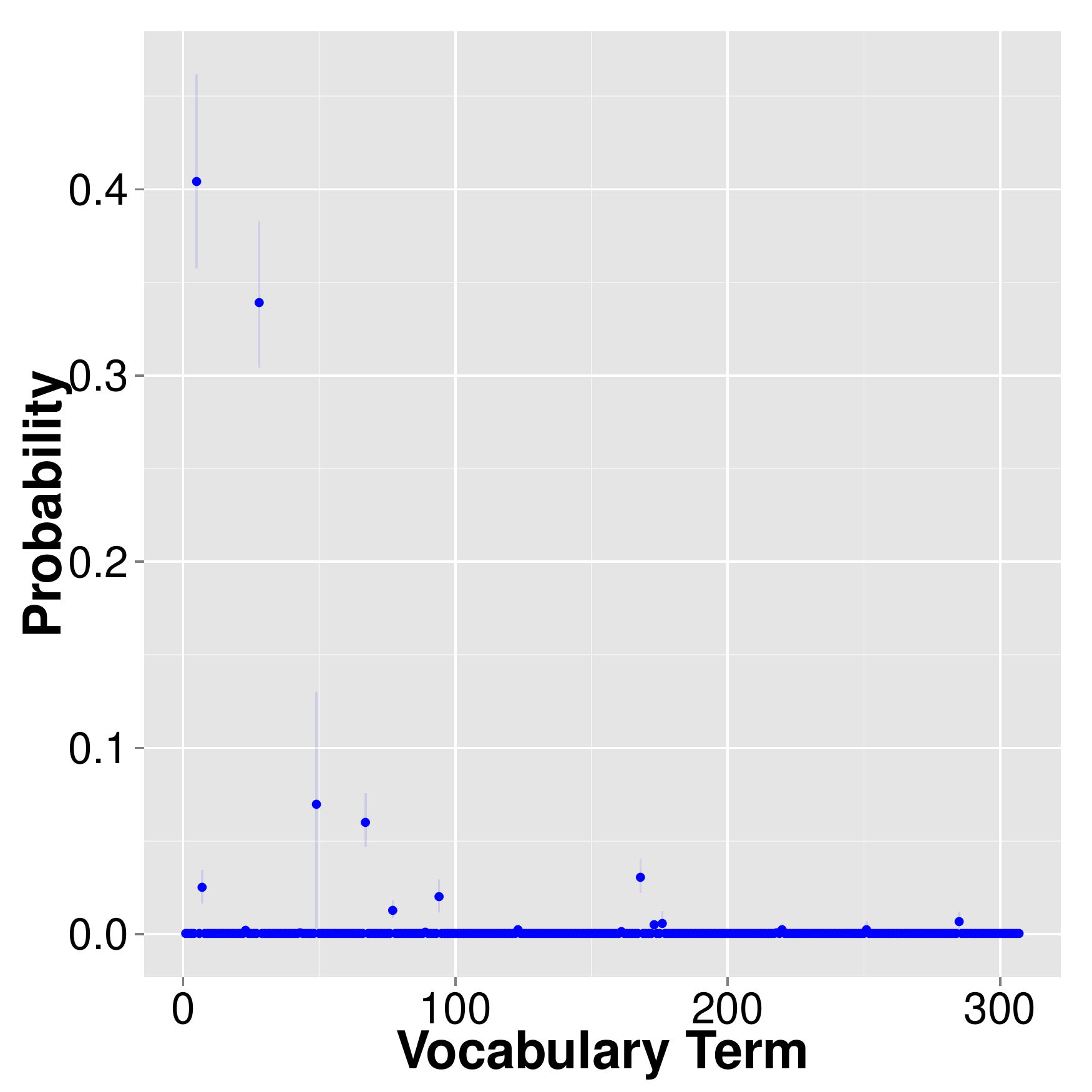}
  \caption{Topic 1}
  \label{subfig:PubMed_Topic_1}
\end{subfigure}
\begin{subfigure}{.2\textwidth}
  \centering
  \includegraphics[width=1\textwidth]{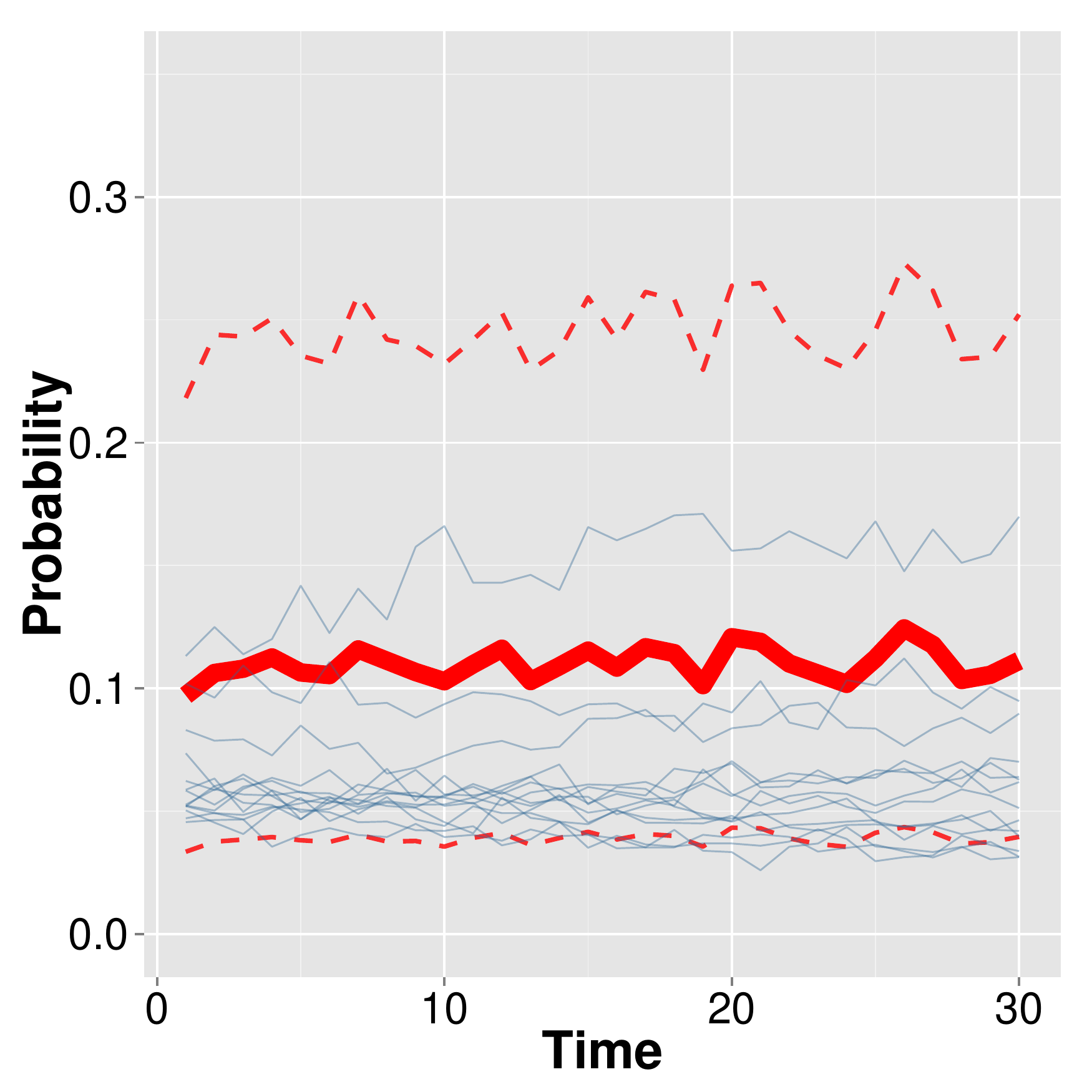}
    \includegraphics[width=1\textwidth]{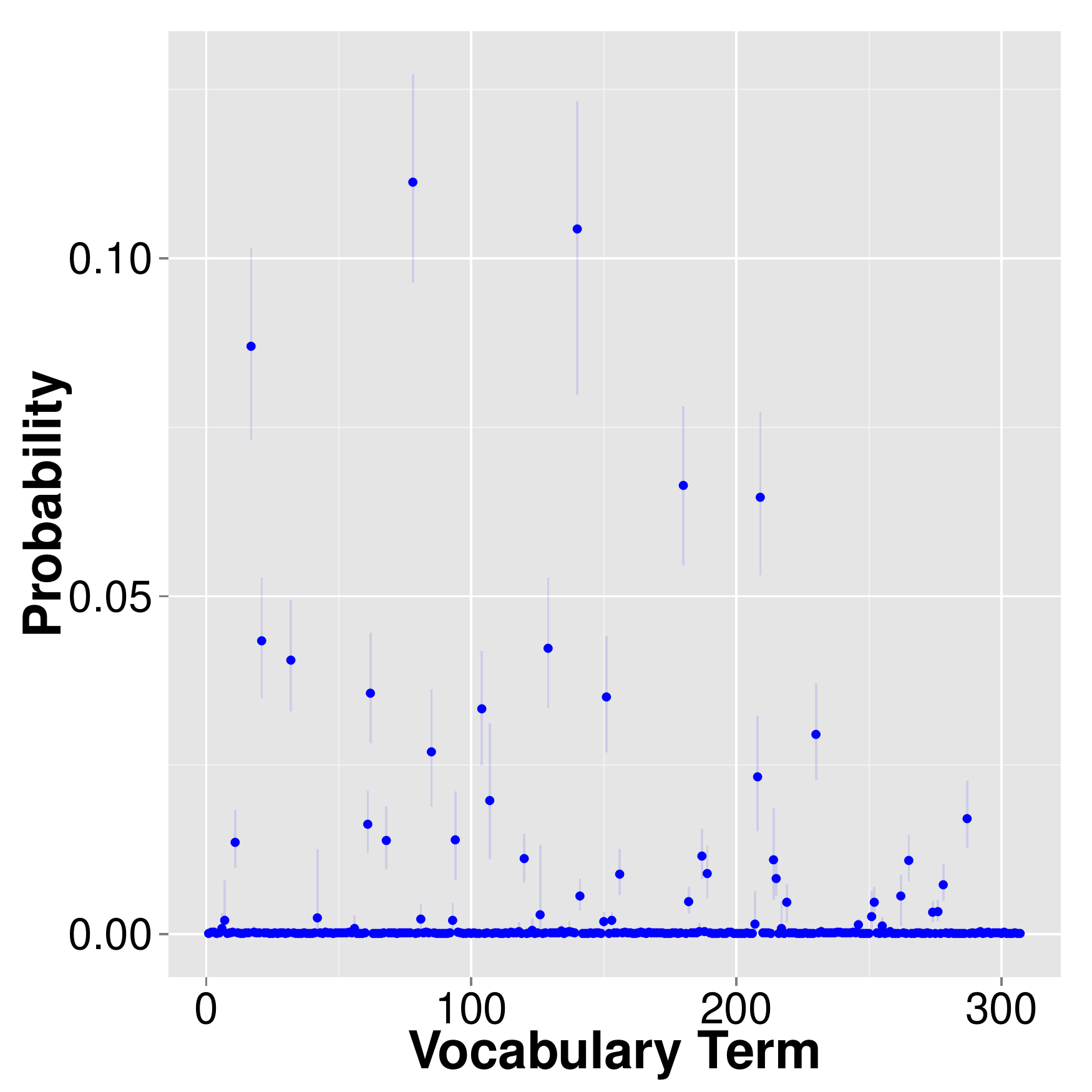}
  \caption{Topic 4}
  \label{subfig:PubMed_Topic_4}
\end{subfigure}
\begin{subfigure}{.2\textwidth}
  \centering
  \includegraphics[width=1\textwidth]{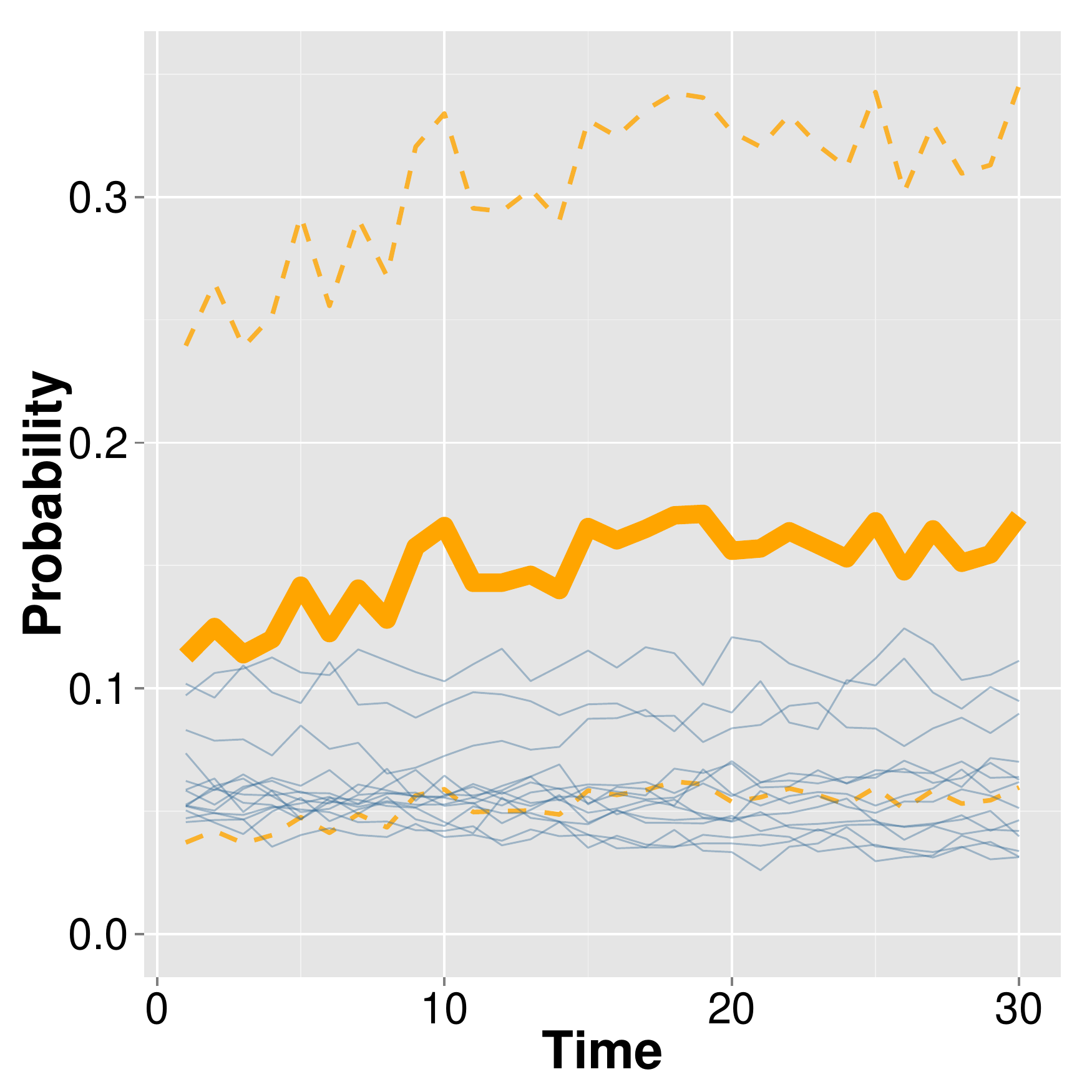}
  \includegraphics[width=1\textwidth]{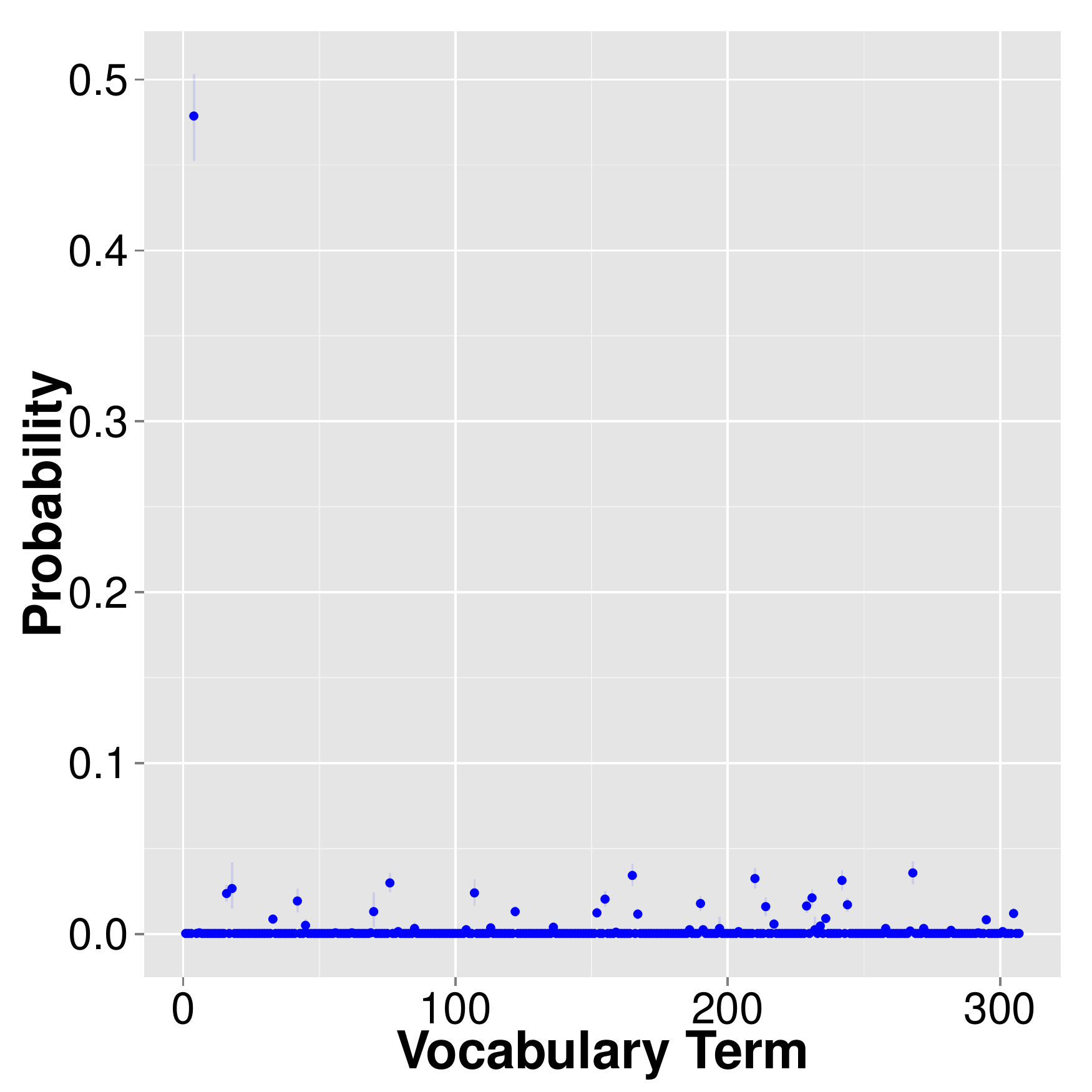}
  \caption{Topic 5}
  \label{subfig:PubMed_Topic_5}
\end{subfigure}
\begin{subfigure}{.2\textwidth}
  \centering
  \includegraphics[width=1\textwidth]{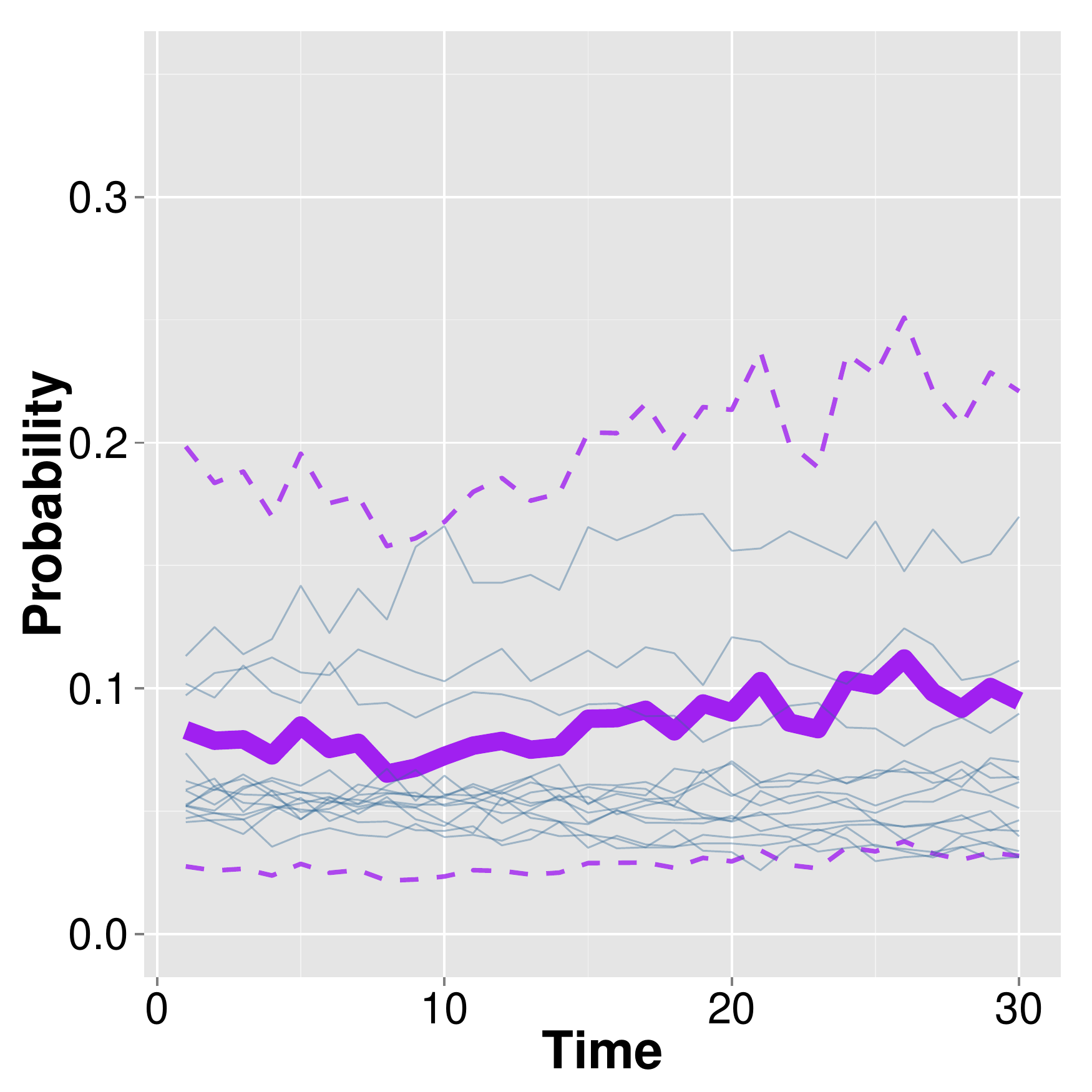}
  \includegraphics[width=1\textwidth]{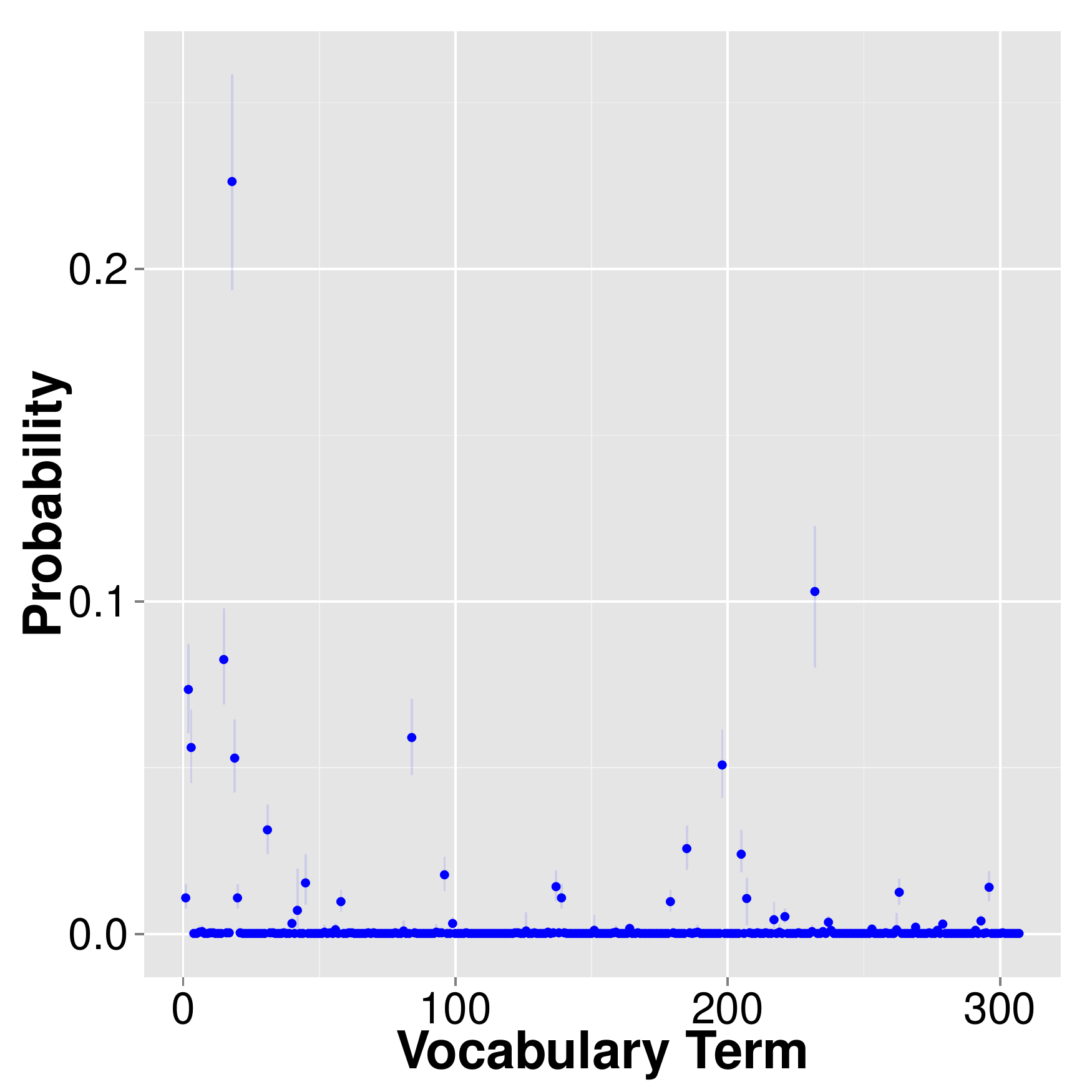}
  \caption{Topic 11}
  \label{subfig:PubMed_Topic_11}
\end{subfigure}
\end{figure}

As presented in the first row of plots in Figure \ref{fig:PubMed_Topics}, there are several topics which receive approximately $10\%$ of the weight in the corpus.  These topics are one, four, five, and eleven.  For enhanced exploration and understanding of topics, we created visualization tools with Google Motion Charts.  All 15 topics can be explored -- both in terms of key words and the dynamics of word probabilities -- at the website \url{https://stat.duke.edu/~cdg28/Topic_Visualization/topic_visualization.html}.  The diffuse prior specified for $\beta_{k,v,t}$ in Section \ref{sec:Prior} enables each topic to load heavily on a few key words.  To visually demonstrate that topics concentrate mass on a few terms, we present topics one, four, five, and eleven in Figure \ref{fig:PubMed_Topics}.

The top word for topic one in 2014, which is presented in Figure \ref{subfig:PubMed_Topic_1},  is hospital.  The second most likely term is tissues.  In topic four, the top term for 2014 is diagnosis which is followed closely by depressives and time.  This is presented in \ref{subfig:PubMed_Topic_4}.  Figures \ref{subfig:PubMed_Topic_5} and \ref{subfig:PubMed_Topic_11} present topics five and eleven in 2014.  In topic five, the top keyword is care, receiving more than 50\% of the topic's weight.  There is no meaningful second term, as all other terms are clustered tightly near zero probability.  In topic eleven, the top term is parameters, which receives approximately 23\% of the weight.  This is followed by healthy and well, which each receive approximately 10\% of the weight in topic eleven.  The high weight of terms that seem unrelated like healthy and parameter suggests that this topic should possibly be split into two distinct topics.  These terms are unnaturally grouped together because $K=15$ is probably too low to accomodate all of the distinct themes in the corpus.  While we find that, when allowing $K$ to be too large, the model and computational method are able to discern which topics are meaningful, fixing $K$ to be too low still presents a challenge for inference.  At the same time, increasing the number of topics increases the computation time.  There is a tradeoff between clarity of inference and computational speed.    

\section{Conclusion}
\label{sec:Conclusion}

In this paper, we make four contributions to the dynamic topic modeling literature.  The first contribution is a mathematically principled framework for modeling complex dynamic behavior in the time-varying marginal probabilities of topics.  The second contribution is an MCMC algorithm which allows us to quantify uncertainty in the exact posterior distribution of both topics themselves and the document-specific topic proportions. The third contribution is a framework for assessing MCMC convergence in topic modeling.  The fourth contribution is the foundation of a model-based method for choosing the number of topics to model a corpus. In addition, we have developed an approximate Gaussian sampler for the Polya-Gamma random variable which is extremely fast.  This contribution to the literature on modeling count data will allow Polya-Gamma data augmentation to be computationally feasible in applications with very large counts.  

We have several directions for future research in dynamic topic modeling.  The first is model checking.  \citet{wallach2009evaluation} perform model checking in LDA by estimating the probability of out-of-sample documents given the estimated parameters in the LDA model.  \citet{mimno2011bayesian} propose using the posterior predictive distribution as a method for checking model fit in a static case.   In the dynamic case, prediction could also be used as a method for evaluating model fit.  By forecasting documents one-step-ahead in time, the researcher could compare the forecasted documents to the observed documents.  In order to compare the quality of the prediction, a distance metric between the predicted and observed documents is necessary.  One possible choice could be the cosine similarity between two documents.  

From a modeling perspective, there is a lot of interesting work to be done.  \citet{Ahmed2012} propose a dynamic Hierarchical Dirichlet Process (HDP) model for inferring the birth and death of topics.  Preserving the dynamic structure of the DLTM and extending it to a model which explicitly allows for topic birth and death is an interesting direction.  

Perhaps the biggest open question for topic modeling is choosing the number of latent topics.  While the hierarchical Dirichlet process of \citet{Teh2006} allows the data to inform the number of latent topics in the corpus, the DTM, LDA, and DLTM all require the modeler to specify $K$.  We are exploring the latent threshold dynamic models of \citet{nakajima2013a} and \citet{nakajima2013b} as a method for specifying $K$.  Figure \ref{subfig:Topic_Proportions_K6} in Section \ref{subsec:misspecification} demonstrates that when $K$ is misspecified to be too large, the unnecessary topics in the corpus will have low (but non-zero) probability.  The latent threshold method will shrink those low probability topics to zero as called for by the data.  By shrinking the probability of unnecessary topics to zero, the problem of selecting $K$ is potentially resolved.  

The choice of $K$ motivates latent thresholding from a statistical perspective, but there is also an applied motivation.  Consider a corpus where we specify $K=100$ topics.  It is extremely unlikely that each document is composed from $100$ different topics.  By eliminating topics that fall below a certain threshold, we will be able to model more heterogenous corpora with increasing numbers of topics without sacrificing a reasonable model for the individual documents.      

We developed our computational algorithm with an eye toward scalability.  Since our Gaussian approximation to the Polya-Gamma random variable does not slow down as the length and number of documents increases, our data-augmentation and MCMC scheme will continue to work with larger and larger datasets.  Parallel computation offers the greatest opportunity for scaling this computational algorithm to handle enormous corpora.  The graphical model representation of Figure \ref{fig:Graph} and the explicit parallelization of MCMC sampling in Section \ref{sec:Inference} demonstrate how different computing environments could be utilized.\\

\bibliographystyle{apalike}
\bibliography{GTBH_2015}

\appendix

\newpage
\section{Derivation of Likelihood Proportionality}
\label{App:Likelihood}

The likelihood of an entire corpus can be computed by taking advantage of the conditional independencies encoded in the graphical representation of the DLTM as presented in Figure \ref{fig:Graph}. For succinct notation, we let $W_{\cdot,t} = \{ W_{1,t}, W_{2,t}, \ldots, W_{D_t, t} \}$ and $W_{\cdot,\cdot} = \{ W_{\cdot,1}, \ldots, W_{\cdot,T} \}$. Then

\begin{align*}
&p(W_{\cdot,1:T} | Z_{\cdot,1:T}, \alpha_{\cdot,1:T}, \beta_{\cdot,\cdot,1:T} ) = \\
&=\prod_{t=1}^T p(W_{\cdot,t} | Z_{\cdot,t}, \alpha_{\cdot,t}, \beta_{\cdot,\cdot,t})\\
&=\prod_{t=1}^T \prod_{d=1}^{D_t} p(W_{d,t} | Z_{d,t}, \alpha_{\cdot,t}, \beta_{\cdot,\cdot,t}) \\
&=\prod_{t=1}^T \prod_{d=1}^{D_t} \prod_{n=1}^{N_{d,t}} p(w_{n,d,t} | z_{n,d,t}, \alpha_{\cdot,t}, \beta_{\cdot,\cdot,t}) \\
&\propto \prod_{t=1}^T \prod_{d=1}^{D_t} \prod_{n=1}^{N_{d,t}} \left( \frac{e^{\beta_{z_{n,d,t},1,t} } }{\sum_{j=1}^{V} e^{\beta_{z_{n,d,t},j,t} } }  \right)^{\mathbbm{1}_{\{w_{n,d,t} = 1 \} } } \ldots\left( \frac{e^{\beta_{z_{n,d,t},V,t} } }{\sum_{j=1}^{V} e^{\beta_{z_{n,d,t},j,t} } }  \right)^{\mathbbm{1}_{\{w_{n,d,t} = V \} } }.
\end{align*}
If we condition on $z_{n,d,t} = k$, the conditional likelihood is proproportional to:

\begin{align*}  
\ell(\beta_{k,t} | z_{n,d,t} = k ) & \propto \left( \frac{e^{\beta_{k,1,t} } }{\sum_{j=1}^{V} e^{\beta_{k,j,t} } } \right)^{\sum_{d=1}^{D_t} \sum_{n=1}^{N_{d,t}} \mathbbm{1}_{ \{ w_{n,d,t}=1\} } \mathbbm{1}_{\{ z_{n,d,t} = k\} } } \times \ldots \\
& \times \left( \frac{e^{\beta_{k,V,t} } }{\sum_{j=1}^{V} e^{\beta_{k,j,t} } } \right)^{\sum_{d=1}^{D_t} \sum_{n=1}^{N_{d,t}} \mathbbm{1}_{ \{ w_{n,d,t}=V\} } \mathbbm{1}_{\{ z_{n,d,t} = k \}} }.    
\end{align*}
Define the notation $y_{k,v,t} = \sum_{d=1}^{D_t} \sum_{n=1}^{N_{d,t}} \mathbbm{1}_{ \{w_{n,d,t} = v \} } \mathbbm{1}_{\{z_{n,d,t} = k \} }$.  With this notation, the conditional likelihood is: 
\begin{align*}  
\ell(\beta_{k,t} | z_{d,n,t} = k) & \propto \left( \frac{e^{\beta_{k,1,t} } }{\sum_{j=1}^{V} e^{\beta_{k,j,t} } } \right)^{y_{k,1,t} } \ldots \left( \frac{e^{\beta_{k,V,t} } }{\sum_{j=1}^{V} e^{\beta_{k,j,t} } } \right)^{y_{k,V,t} }.    
\end{align*}

Now focus on the probability for an arbitrary term $v$.  The objective is to manipulate the expression to one which is easily amenable to conditioning on $\beta_{k,-v,t}$.  The $\beta_{k,-v,t}$ is the set of all $\beta_{k,j,t}$ parameters where $j \neq v$. Throughout, we will take $\beta_{k,-v,t}$ to be a $(V-1)$-dimensional vector of all terms excluding $\beta_{k,v,t}$. This likelihood manipulation and conditioning strategy is taken directly from \citet{holmes2006}.  So
     
\begin{align*}
Pr(w_{d,n,t} = v | z_{d,n,t} = k )
 = \frac{ e^{\beta_{k,v,t} } } {\sum_{j=1}^V e^{\beta_{k,j,t} } }  
 = \frac{ e^{\beta_{k,v,t} } } { e^{\beta_{k,v,t}} + \sum_{j \neq v } e^{\beta_{k,j,t} } }
 = \frac{ \frac{ e^{\beta_{k,v,t} } }{\sum_{j \neq v} e^{\beta_{k,j,t} }  }  }{ \frac{ e^{\beta_{k,v,t} } } { \sum_{j \neq v} e^{\beta_{k,j,t} }  } + 1 }.
\end{align*}

Define
\begin{align*}
\gamma_{k,v,t} & := \beta_{k,v,t} + \log \frac{1}{\sum_{j \neq v} e^{\beta_{k,j,t} } } = \beta_{k,v,t} - \log \sum_{j \neq v} e^{\beta_{k,j,t} }.
\end{align*}
This definition for $\gamma_{k,v,t}$ yields $ e^{\gamma_{k,v,t} }  = \frac{e^{\beta_{k,v,t} } }{\sum_{j \neq v} e^{\beta_{k,j,t} } }$.  This is significant because we are now able to express the original probability of term $v$ as a function of $\gamma_{k,v,t}$ which separates $\beta_{k,v,t}$ and $\beta_{k,-v,t}$:
 
\begin{center}
$Pr(w_{n,d,t} = v | z_{n,d,t} = k) = \frac{e^{ \beta_{k,v,t} } }{\sum_{j=1}^V e^{\beta_{k,j,t} } } = \frac{ e^{\gamma_{k,v,t} } }{1+ e^{\gamma_{k,v,t} } }$.
\end{center}

Observe that the multinomial likelihood, conditional on $\beta_{k,-v,t}$, is proportional to a reparameterized expression involving $\gamma_{k,v,t}$ which resembles the binomial likelihood.   This expression and binomial form  will enable us to utilize Polya-Gamma data augmentation methods to perform inference with the multinomial likelihood.  Thus
    
\begin{align*}  
\ell(\beta_{k,v,t} | \beta_{k,-v,t}, Z_{\cdot,t} )&\propto \left( \frac{e^{\beta_{z_{n,d,t},1,t} } }{\sum_{j=1}^{V} e^{\beta_{z_{n,d,t},j,t} } } \right)^{y_{k,1,t} } \ldots \left( \frac{e^{\beta_{z_{n,d,t},v,t} } }{\sum_{j=1}^{V} e^{\beta_{z_{n,d,t},j,t} } } \right)^{y_{k,v,t} }  \ldots \left( \frac{e^{\beta_{z_{n,d,t},V,t} } }{\sum_{j=1}^{V} e^{\beta_{z_{n,d,t},j,t} } } \right)^{y_{k,V,t} } \\
& \propto \left(\frac{ \frac{e^{\beta_{z_{n,d,t},1,t} } }{ \sum_{j\neq v} e^{\beta_{z_{n,d,t},j,t } } } }{1+e^{\gamma_{k,v,t} } } \right)^{y_{k,1,t}} \ldots \left( \frac{e^{\gamma_{k,v,t} } }{1+ e^{\gamma_{k,v,t} } } \right)^{y_{k,v,t} } \ldots \left( \frac{\frac{e^{\beta_{z_{n,d,t},V,t} } }{ \sum_{j\neq v} e^{\beta_{z_{n,d,t},j,t } } }    }{1 +e^{\gamma_{k,v,t} } } \right)^{y_{k,V,t} }  \\ 
& \propto \left(\frac{1}{1+e^{\gamma_{k,v,t} } } \right)^{y_{k,1,t}} \ldots \left( \frac{e^{\gamma_{k,v,t} } }{1+ e^{\gamma_{k,v,t} } } \right)^{y_{k,v,t} } \ldots \left( \frac{1}{ 1+e^{\gamma_{k,v,t} } } \right)^{y_{k,V,t} }  \\ 
& \propto \left(\frac{e^{\gamma_{k,v,t} } }{1 + e^{\gamma_{k,v,t} } } \right)^{y_{k,v,t} } \left(\frac{1}{1+e^{\gamma_{k,v,t} } } \right)^{n_{k,t}^y - y_{k,v,t}}
\end{align*}
where $y_{k,v,t} = \sum_{d=1}^{D_t} \sum_{n=1}^{N_{d,t}} \mathbbm{1}_{\{w_{n,d,t}=v \} } \mathbbm{1}_{\{z_{n,d,t} = k \}}$ is the number of times vocabulary term $v$ is assigned to topic $k$ across all documents at time $t$; $n_{k,t}^y = \sum_{j = 1}^V y_{k,j,t}$ is the number of total words assigned to topic $k$ at time $t$; and $\gamma_{k,v,t} = \beta_{k,v,t} - \log \sum_{j \neq v} e^{\beta_{k,j,t} } $, the function which separates $\beta_{k,v,t}$ and $\beta_{k,-v,t}$.
This conditional likelihood allows us to proceed with a Gibbs sampling algorithm as outlined in Section \ref{sec:Inference}.

\newpage
\section{DLTM Topics from Linear, Quadratic, and Harmonic simulation examples}
\label{App:DLTM_Complex}
In Figure \ref{fig:Topic_Post_Mean_LL}, we present the topics recovered from the simulation exercises with linear, quadratic, and harmonic trends.  The first row presents the MCMC and variational topics from the linear trend simulation example.  The second and third row present the topics from the quadratic and harmonic examples, respectively.  

\begin{figure}[h!]
\centering
\caption{Posterior mean for probability of $v^{th}$ term when Topic probability is modeled with local linear trend.  First row is from DLTM with linear trend.  Second row is from DLTM with quadratic trend.  Third row is from DLTM with harmonic trend.}
\label{fig:Topic_Post_Mean_LL}
\begin{subfigure}{.3\textwidth}
  \centering
  \includegraphics[width=1\textwidth]{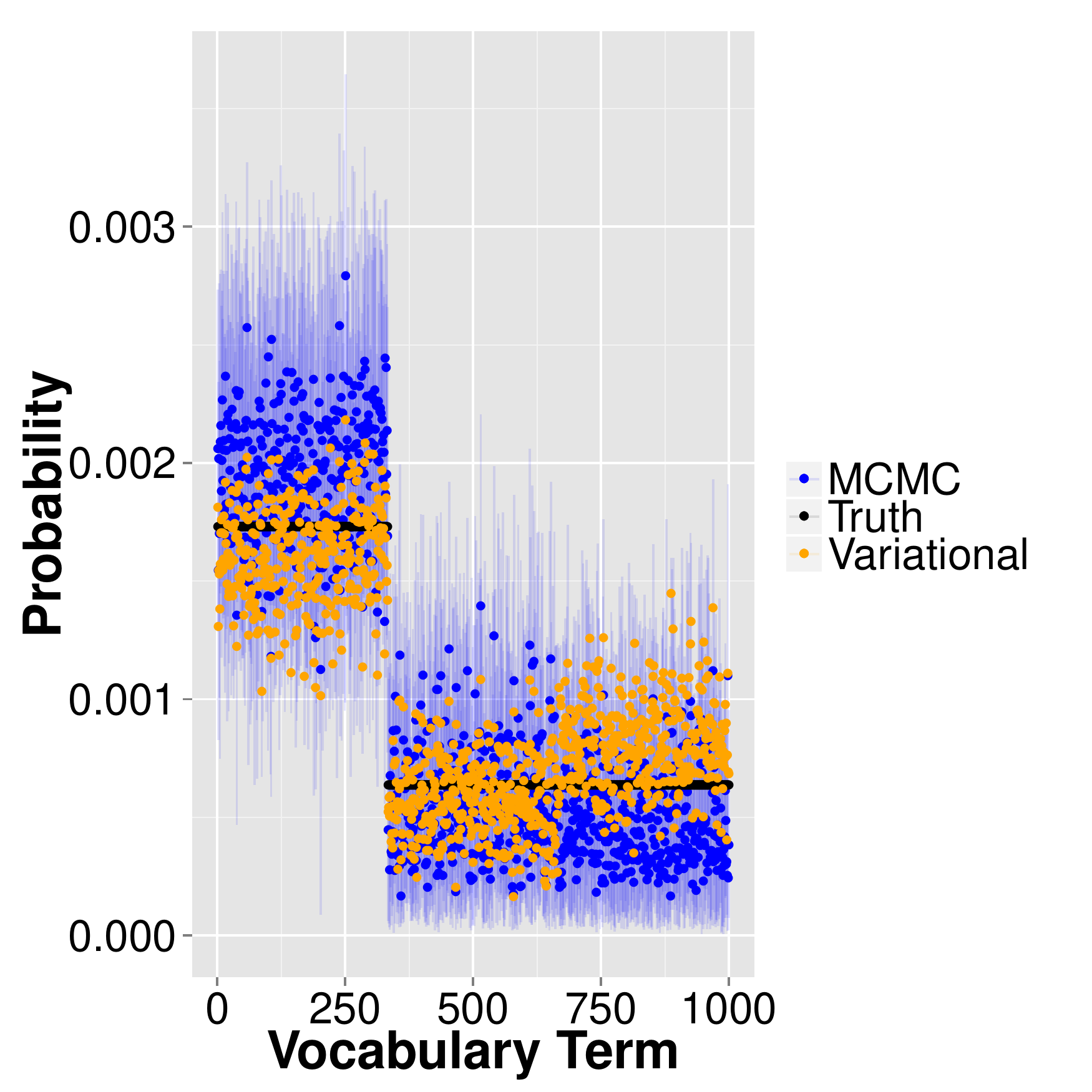}
  \includegraphics[width=1\textwidth]{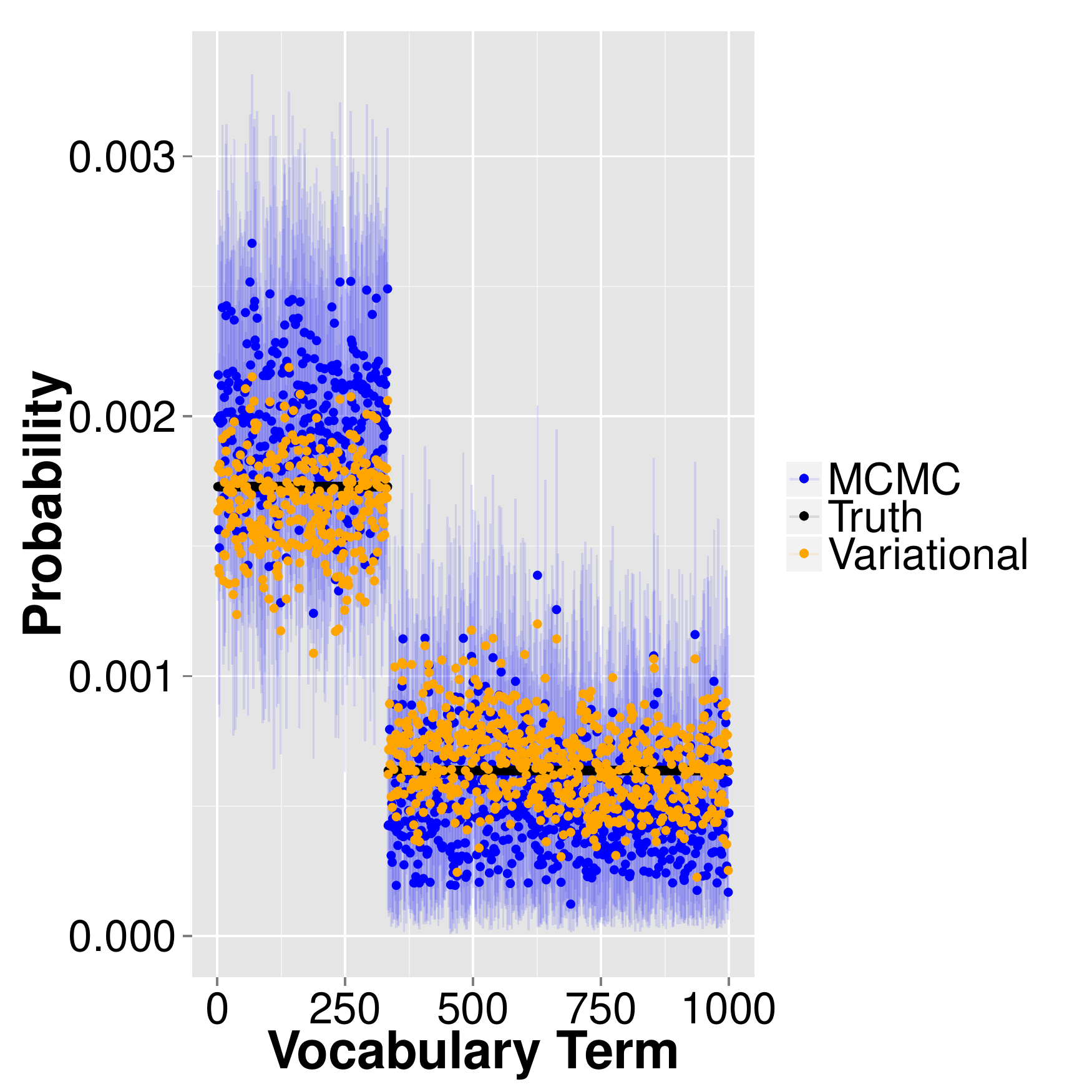}
  \includegraphics[width=1\textwidth]{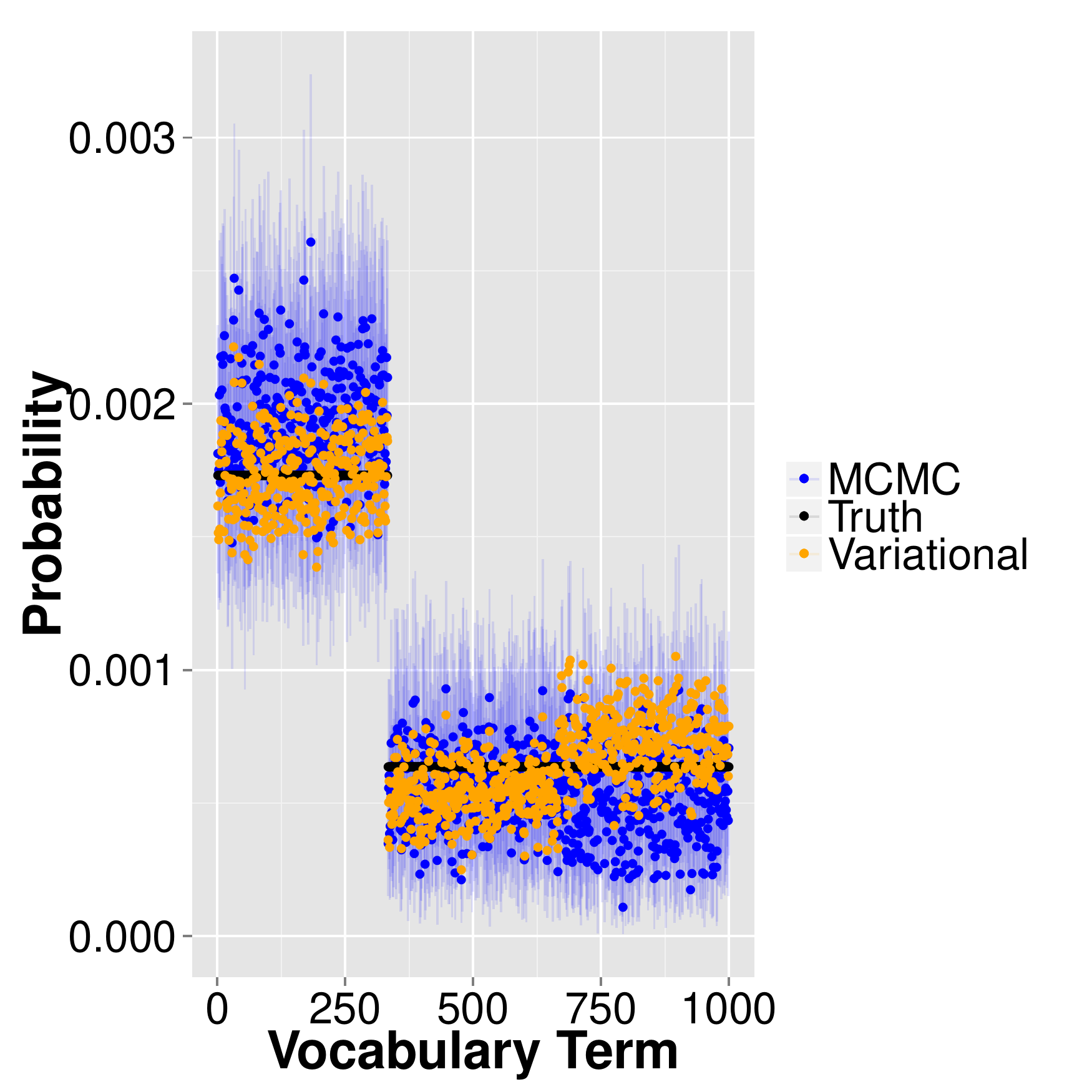}
  \caption{Topic 1}
  \label{subfig:Topic_1_Post_LL}
\end{subfigure}%
\begin{subfigure}{.3\textwidth}
  \centering
  \includegraphics[width=1\textwidth]{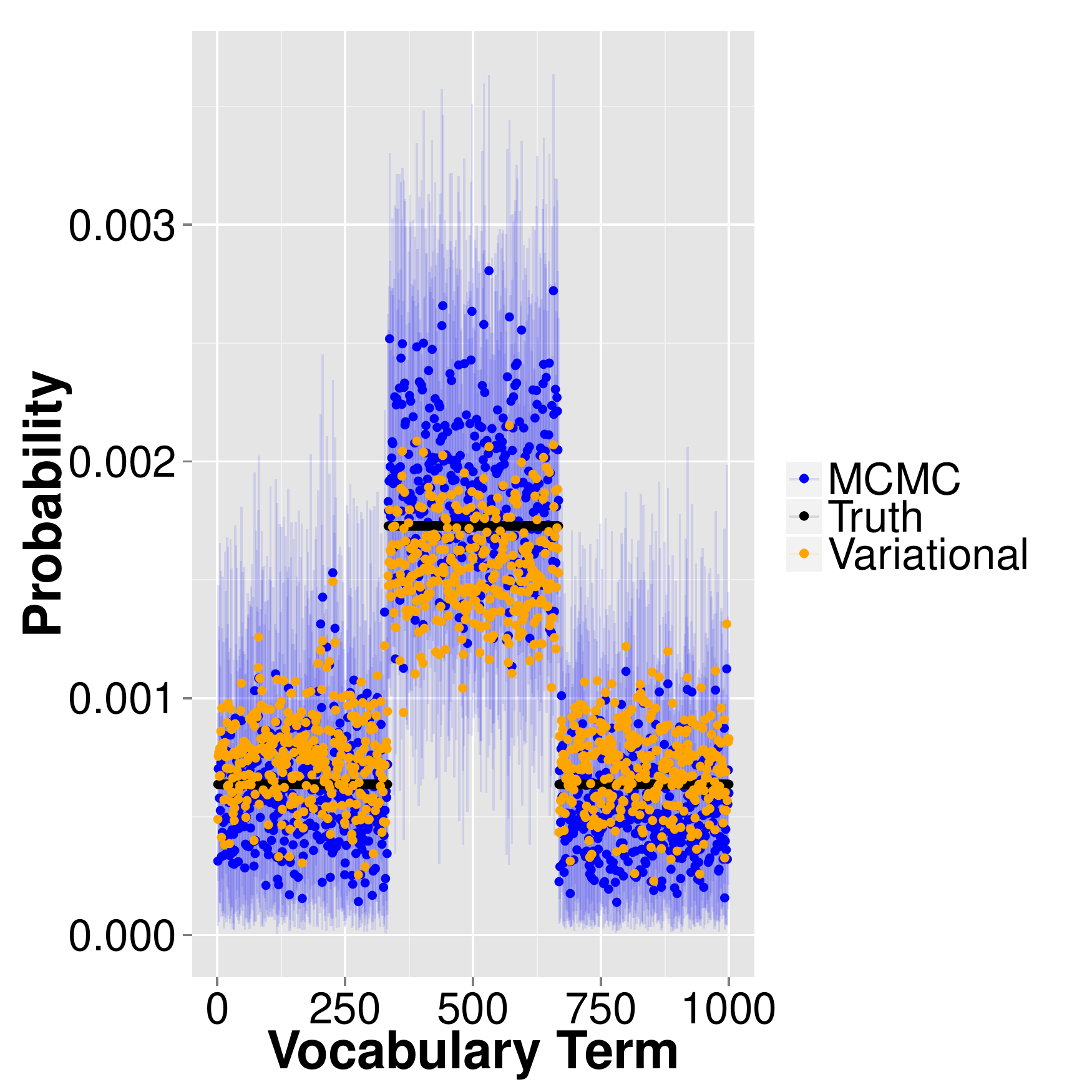}
  \includegraphics[width=1\textwidth]{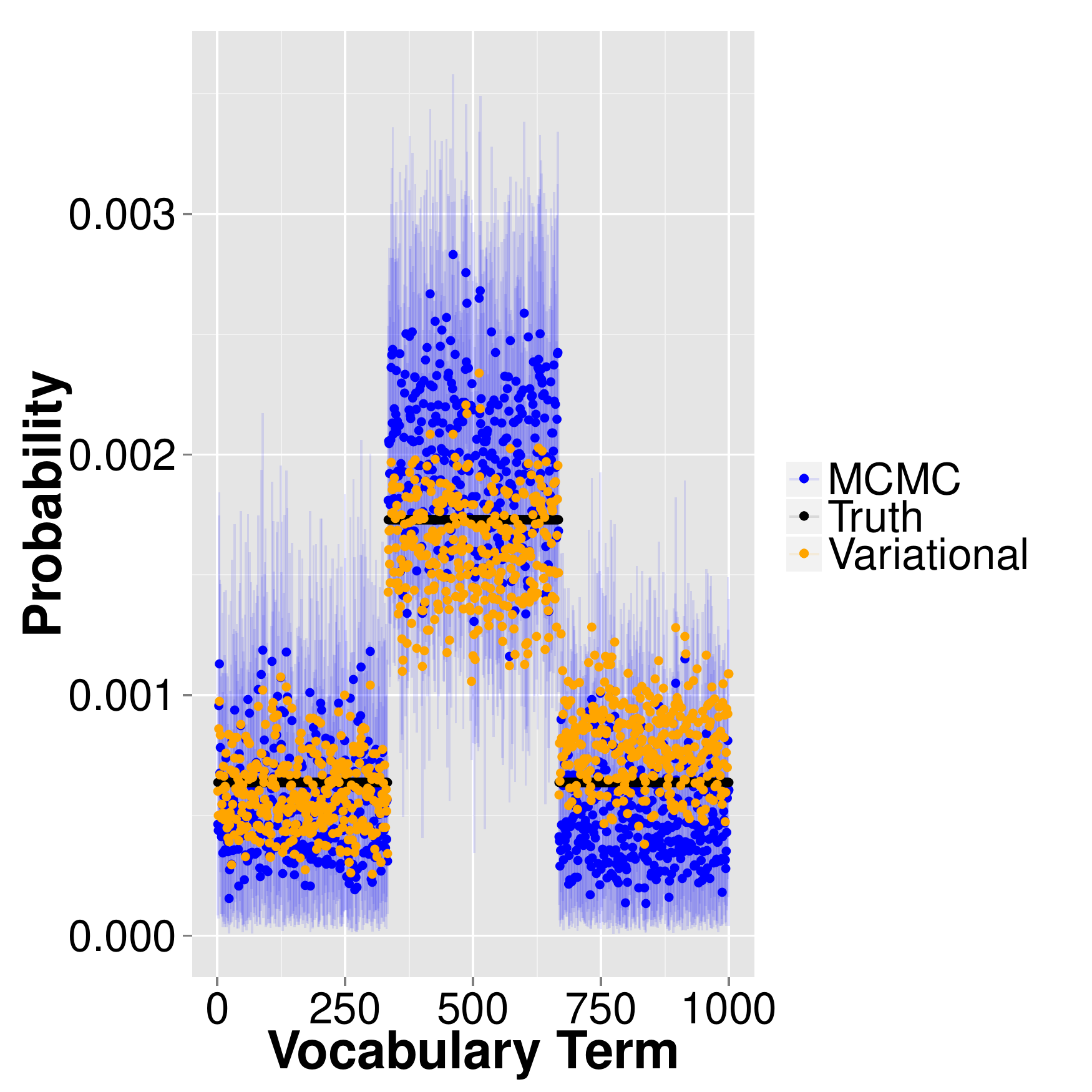}
  \includegraphics[width=1\textwidth]{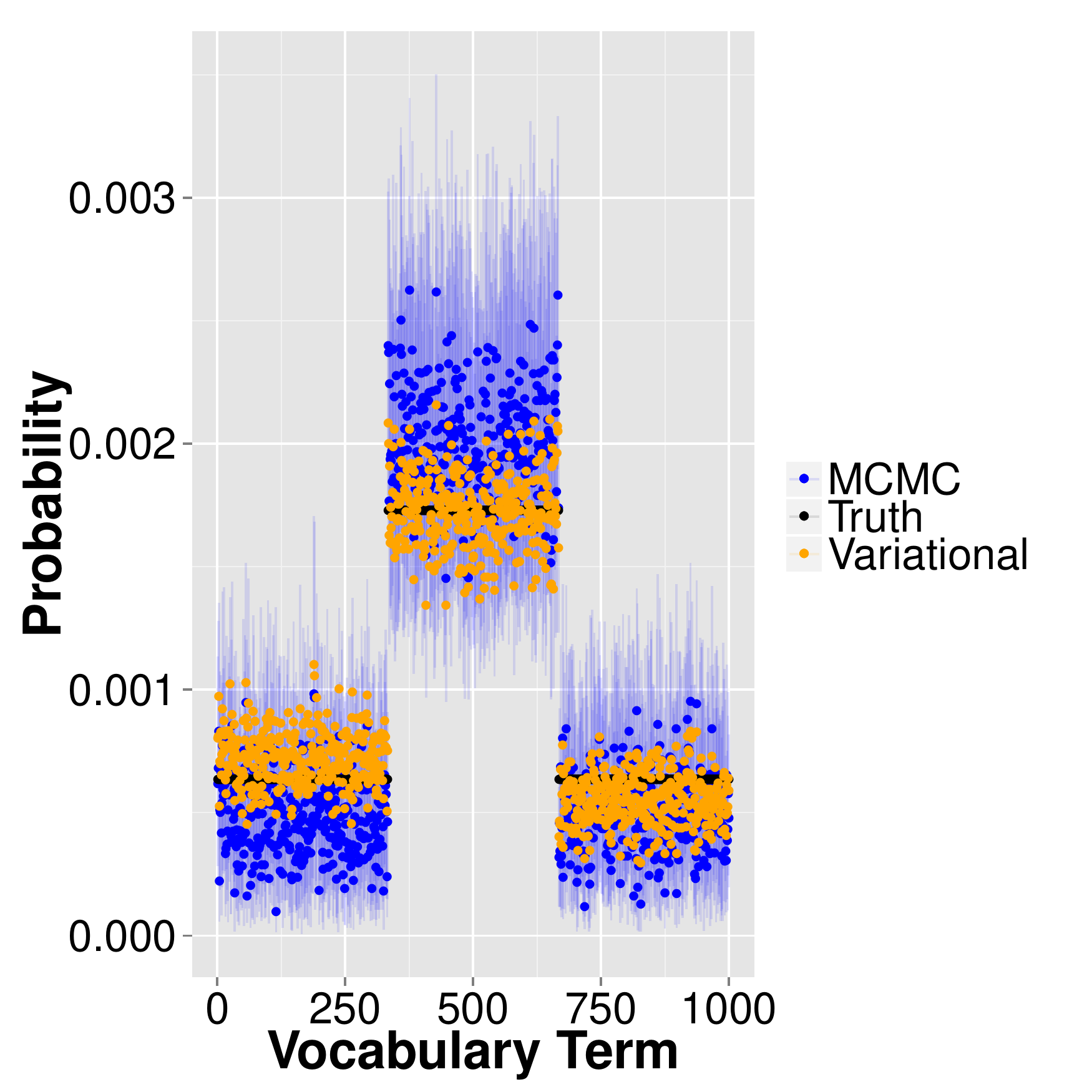}
  \caption{Topic 2}
  \label{Topic_2_Post_LL}
\end{subfigure}
\begin{subfigure}{.3\textwidth}
  \centering
  \includegraphics[width=1\textwidth]{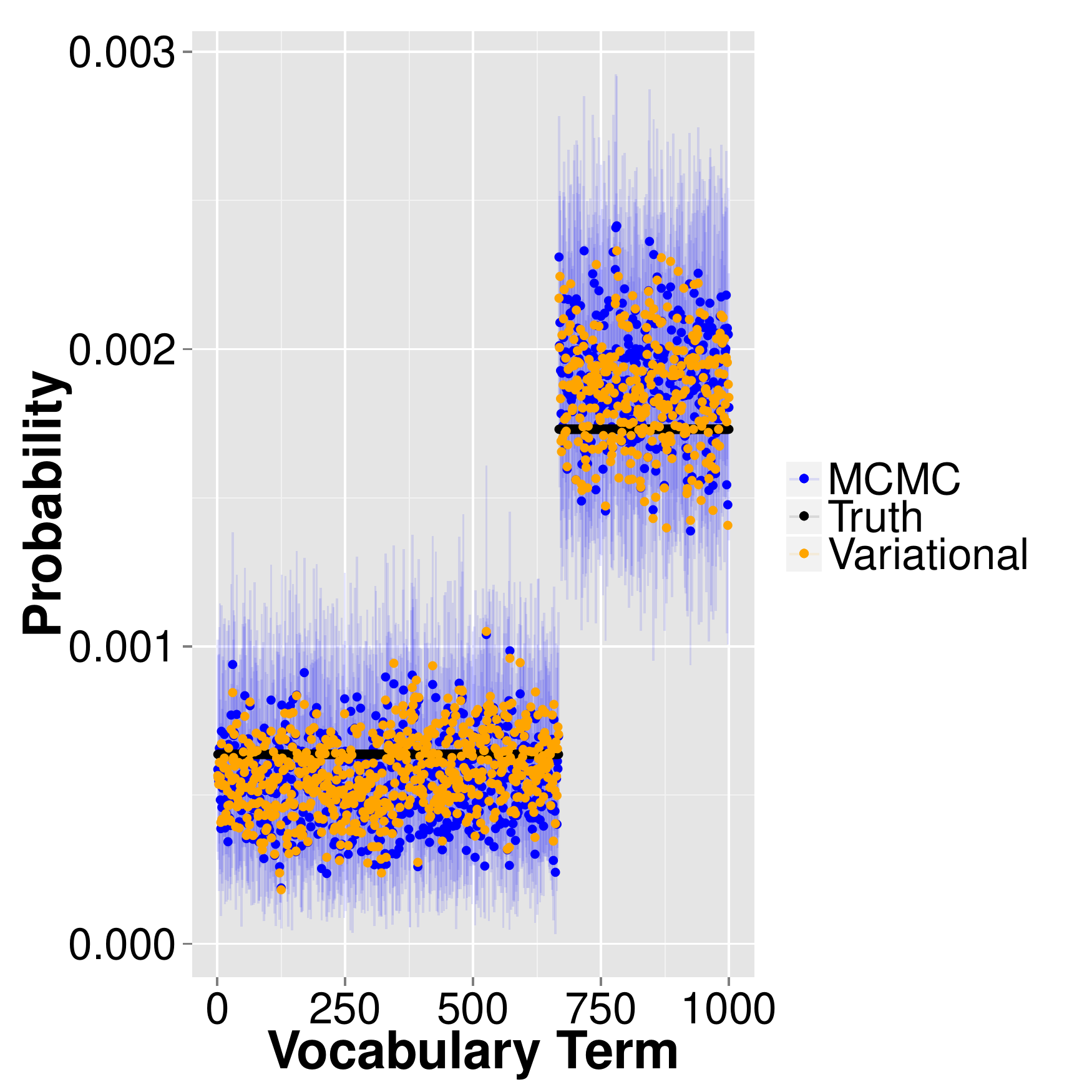}
  \includegraphics[width=1\textwidth]{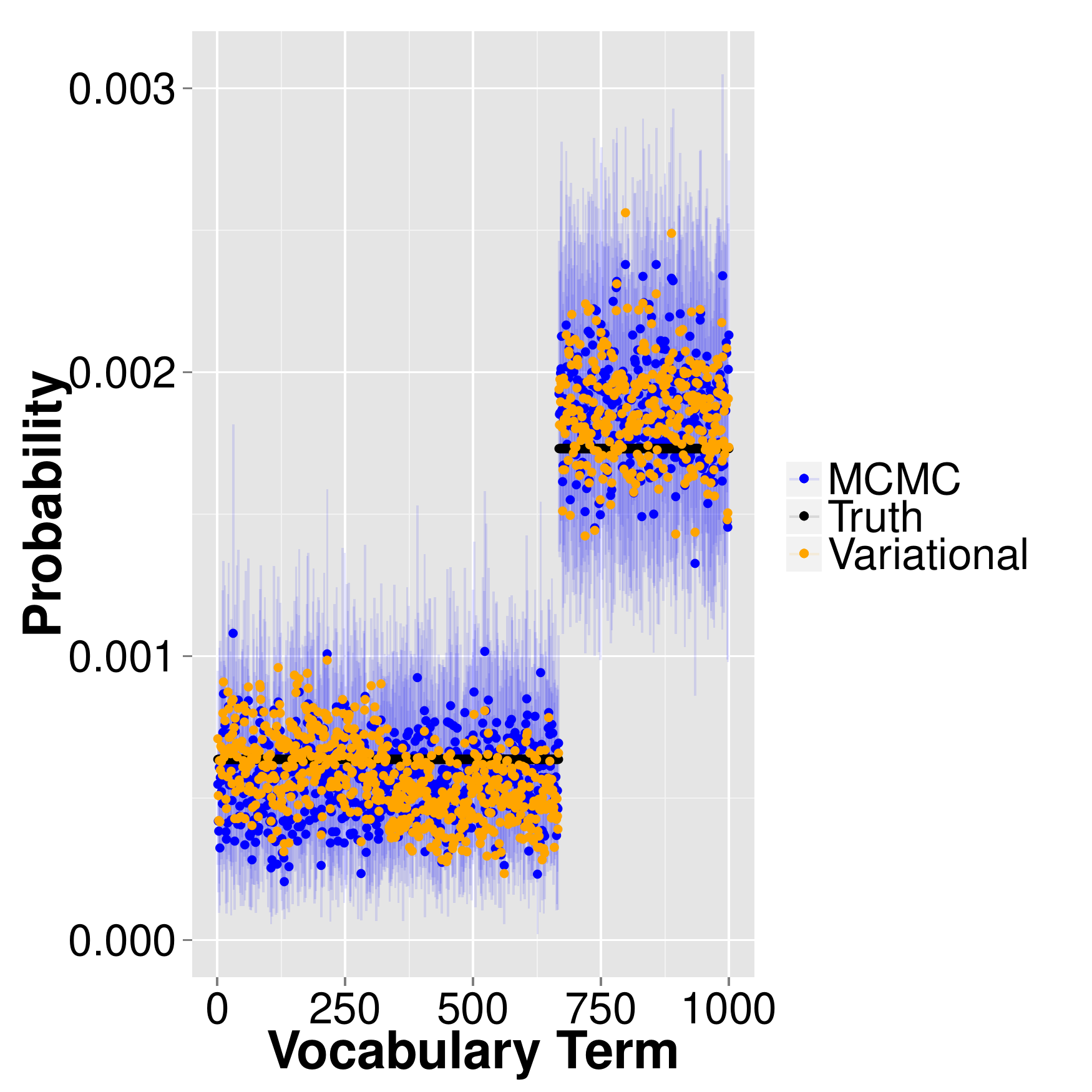}
  \includegraphics[width=1\textwidth]{Truth_Post_Mean_piecewise_501_1.pdf}
  \caption{Topic 3}
  \label{subfig:Topic_3_Post_LL}
\end{subfigure}
\end{figure}
        
\newpage
\section{Full Conditionals}
\label{App:Appendix_A}
\subsection{Full Conditional for $\beta_{k,v,t}$}
\label{subsec:FC_Beta}
In this section we compute the full conditional $\beta_{k,v,t}| \beta_{k,-v,1:t}, W_{\cdot,1:t}, Z_{\cdot,1:t}, \zeta_{k,v,1:t}$. 
 
Define $\kappa_{k,v,t}^{\beta} := y_{k,v,t} - \frac{1}{2} n_{k,t}^y$.  Also, by definition $\beta_{k,v,0} \sim N(m_{k,v,0}, \sigma_{k,v,0}^2)$.  We will focus on finding the posterior $\beta_{k,v,1}| \beta_{k,-v,1},W_{\cdot,1}, Z_{\cdot,1}, \zeta_{k,v,1}$ where $\zeta_{k,v,1} \sim PG(0, n_{k,t})$ is a Polya-Gamma random variable.  

As a result from Theorem 1 from \citet{PSW2013}, we can compute the desired posterior by augmenting the conditional likelihood with a Polya-Gamma random variable $\zeta_{k,v,t}$:  
\begin{align*}
\ell(\beta_{k,v,1} | \beta_{k,-v,1}, Z_{\cdot,1},  \zeta_{k,v,1}) \propto exp \{\kappa_{k,v,1}^{\beta} \gamma_{k,v,1} - \frac{\zeta_{k,v,1}}{2} \gamma_{k,v,1}^2 \}.
\end{align*}  
The model specifies that $\beta_{k,t}| \beta_{k,t-1} \sim N(\beta_{k,t-1}, \sigma^2 I_{V}) $.  By this covariance structure, $\beta_{k,v,t} \indep \beta_{k,j,t}$ for $v,j \in \{1,\ldots, V\}$ and $v\neq j$.  It follows that the prior for $p(\beta_{k,v,1}| \beta_{k,-v,t}) = p(\beta_{k,v,t})$ and  $\beta_{k,v,1} \sim N(m_{k,v,0}, \rho^2_{k,v,1} )$ where $\rho^2_{k,v,1} = \sigma^2_{k,v,0} + \sigma^2$.  Due to the log-quadratic form of the conditional likelihood and the Gaussian prior, the posterior at time $t=1$,  $\beta_{k,v,1} | \beta_{k,-v,1}, W_{\cdot,1},Z_{\cdot,1}, \zeta_{k,v,1}$, is Gaussian.  Below is the exact posterior computation.

Remember that $\gamma_{k,v,1} = \beta_{k,v,1} - \log \sum_{j \neq v} e^{\beta_{k,j,1}}$.  So  
\begin{align*}
& \ell(\beta_{k,v,1} | \beta_{k,-v,1}, Z_{\cdot,1}, \zeta_{k,v,1}) p(\beta_{k,v,1}) \propto \\ 
& \propto \mbox{exp} \left\{\kappa_{k,v,1}^{\beta} \gamma_{k,v,1} - \frac{ \zeta_{k,v,1} }{2}\gamma_{k,v,1}^2\ \right\} p(\beta_{k,v,1} ) \\
& \propto \mbox{exp} \left\{\kappa_{k,v,1}^{\beta} \beta_{k,v,1} - \frac{\zeta_{k,v,1}}{2} \left(\beta_{k,v,1} - \log \sum_{j\neq v} e^{\beta_{k,j,1}} \right)^2 \right\} \mbox{exp} \left\{-\frac{1}{2 \rho_{k,v,1}^2} \left(\beta_{k,v,1} - m_{k,v,0}\right)^2\right\} \\ 
& \propto \mbox{exp}\left\{\kappa_{k,v,1}^{\beta} \beta_{k,v,1} - \frac{\zeta_{k,v,1}}{2} \left( \beta_{k,v,1}^2 - 2 \beta_{k,v,1} \log \sum_{j \neq v} e^{\beta_{k,j,1} } \right) - \frac{1}{2 \rho_{k,v,1}^2} \left(\beta_{k,v,1}^2 - 2 \beta_{k,v,1} m_{k,v,0}\right) \right\} \\
& \propto \mbox{exp} \left\{-\frac{1}{2} \left(\zeta_{k,v,1} + \frac{1}{\rho_{k,v,1}^2} \right) \beta_{k,v,1}^2 + \beta_{k,v,1}\left(\kappa_{k,v,1}^{\beta} + \zeta_{k,v,1} \log \sum_{j \neq v} e^{\beta_{k,j,1}} + \frac{1}{\rho_{k,v,1}^2} m_{k,v,0}\right) \right\}
\end{align*} 
From this last line, it is clear that:
\begin{itemize}
\item $\beta_{k,v,1} | \beta_{k,-v,1}, W_{\cdot,1},Z_{\cdot,1}, \zeta_{k,v,1} \sim N(m_{k,v,1}, \sigma_{k,v,1}^2)$ 
\item $\sigma^2_{k,v,1} = \left(\zeta_{k,v,1} + \frac{1}{\rho_{k,v,1}^2} \right)^{-1}$
\item $m_{k,v,1} = \sigma^2_{k,v,1} \left( \kappa_{k,v,1}^{\beta} + \zeta_{k,v,1} \log \sum_{j\neq v} e^{\beta_{k,j,1} } + \frac{m_{k,v,0}}{\rho_{k,v,1}^2} \right)$.
\end{itemize}
 The proof for the sequential update of $\{\beta_{k,v,t}\}_{t=1}^T$ is by induction.  

Suppose that the posterior at time $t-1$ is \begin{center} $\beta_{k,v,t-1} | \beta_{k,-v,1:t-1}, W_{\cdot,1:t-1},Z_{\cdot,1:t-1} \zeta_{k,v,1:t-1}  \sim N(m_{k,v,t-1}, \sigma_{k,v,t-1}^2)$. \end{center}  Again by independence, this yields that the prior \begin{center} $\beta_{k,v,t} | \beta_{k,-v,1:t-1}, W_{\cdot,1:t-1},Z_{\cdot,1:t-1}, \zeta_{k,v,1:t-1} \sim N(m_{k,v,t-1}, \rho_{k,v,t}^2)$\end{center} where $\rho_{k,v,t}^2 = \sigma^2_{k,v,t-1} + \sigma^2$.  So   

\begin{align*}
& \ell(\beta_{k,v,t} | \beta_{k,-v,t}, Z_{\cdot,t}, \zeta_{k,v,t} ) p(\beta_{k,v,t} | \beta_{k,-v,1:t-1}, W_{\cdot,1:t-1},Z_{\cdot,1:t-1}, \zeta_{k,v,1:t-1}) \propto \\
& \propto \mbox{exp} \left\{ \kappa_{k,v,t}^{\beta} \gamma_{k,v,t} - \frac{\zeta_{k,v,t}}{2} \gamma_{k,v,t}^2\right\} p(\beta_{k,v,t} | \beta_{k,-v,1:t-1}, W_{\cdot,1:t-1},\zeta_{k,v,1:t-1} ) \\
& \propto \mbox{exp} \left\{ \kappa_{k,v,t}^{\beta} \beta_{k,v,t} - \frac{\zeta_{k,v,t}}{2} \left(\beta_{k,v,t} - \log \sum_{j \neq v} e^{\beta_{k,j,t}}\right)^2 - \frac{1}{2\rho_{k,v,t}^2} \left(\beta_{k,v,t} - m_{k,v,t-1}\right)^2 \right\}\\
& \propto \mbox{exp} \left\{ -\frac{1}{2}\left( \zeta_{k,v,t} + \frac{1}{\rho_{k,v,t}^2} \right) \beta_{k,v,t}^2 + \beta_{k,v,t} \left( \kappa_{k,v,t}^{\beta} + \zeta_{k,v,t} \log \sum_{j \neq v} e^{\beta_{k,j,t} } + \frac{m_{k,v,t-1}}{\rho_{k,v,t}^2}\right) \right\}  
\end{align*}

Again, from this last line, it is clear that:
\begin{itemize}
\item $\beta_{k,v,t} | \beta_{k,-v,t}, W_{\cdot,1:t},Z_{\cdot,1:t}, \zeta_{k,v,1:t} \sim N(m_{k,v,t}, \sigma_{k,v,t}^2)$  
\item $\sigma_{k,v,t}^2 = \left(\zeta_{k,v,t} + \frac{1}{\rho_{k,v,t}^2} \right)^{-1}$
\item $m_{k,v,t} = \sigma_{k,v,t}^2\left(\kappa_{k,v,t}^{\beta} + \zeta_{k,v,t} \log \sum_{j \neq v} e^{\beta_{k,j,t} } + \frac{m_{k,v,t-1} }{\rho_{k,v,t}^2} \right)$.
\end{itemize}

What we have computed thus far is the filtered distributions $\beta_{k,v,t} | \beta_{k,-v,1:t}, W_{\cdot,1:t}, Z_{\cdot,1:t}, \zeta_{k,v,1:t}$.  Our goal is to sample from the joint distribution $\beta_{k,v,1:T} | \beta_{k,-v,1:T}, W_{\cdot,1:T}, Z_{\cdot,1:T}, \zeta_{k,v,1:T}$.  We do this with a backward recursion.

\begin{enumerate}
\item Sample $\beta_{k,v,T} | \beta_{k,-v,1:T}, W_{\cdot,1:T}, Z_{\cdot,1:T}, \zeta_{k,v,1:T}$.
\item Sample $\beta_{k,v,T-1} | \beta_{k,v,T}, \beta_{k,-v,1:T}, W_{\cdot,1:T}, Z_{\cdot,1:T}, \zeta_{k,v,1:T}$.
\item sample continue recursively: $\beta_{k,v,t-1} | \beta_{k,v,t:T}, \beta_{k,-v,1:T}, W_{\cdot,1:T}, Z_{\cdot,1:T}, \zeta_{k,v,1:T}$.
\end{enumerate}

The first step is straightforward given the filtered distributions.  By the conditional independence structure as demonstrated in the graphical model form shown in Figure \ref{fig:Graph}, \begin{center} $p(\beta_{k,v,t-1} | \beta_{k,v,t:T}, \beta_{k,-v,1:T}, W_{\cdot,1:T}, Z_{\cdot,1:T}, \zeta_{k,v,1:T} ) = p( \beta_{k,v,t-1} | \beta_{k,v,t}, \beta_{k,-v,1:t-1}, W_{\cdot,1:t-1}, Z_{\cdot,1:t-1}, \zeta_{1:t-1} )$. \end{center} The objective now is to find this condtional distribution.

\begin{align*}
& p(\beta_{k,v,t-1} | \beta_{k,v,t}, \beta_{k,-v,1:t-1}, W_{\cdot,1:t-1}, Z_{\cdot,1:t-1}, \zeta_{k,v,1:t-1} )  \\
& \propto p(\beta_{k,v,t}, \beta_{k,v,t-1} | \beta_{k,-v,1:t-1}, W_{\cdot,1:t-1}, Z_{\cdot,1:t-1}, \zeta_{k,v,1:t-1} ) \\
& \propto p(\beta_{k,v,t} | \beta_{k,v,t-1} ) p( \beta_{k,v,t-1} | \beta_{k,-v,1:t-1}, W_{\cdot,1:t-1}, Z_{\cdot,1:t-1}, \zeta_{k,v,1:t-1} )
\end{align*} 

The desired conditional distribution factorizes up to a constant into two distributions which we have readily available: $p(\beta_{k,v,t} | \beta_{k,v,t-1})$, which is the 1-step transition distribution specified in the model, and $p(\beta_{k,v,t-1} | \beta_{k,-v,1:t-1}, W_{\cdot,1:t-1}, Z_{\cdot,1:t-1}, \zeta_{k,v,1:t-1})$, which is the filtered distribution for $\beta_{k,v,t-1}$.  

Proceeding with a standard posterior computation:
\begin{align*}
& p(\beta_{k,v,t-1} | \beta_{k,v,t}, \beta_{k,-v,1:t-1}, W_{\cdot,1:t-1}, Z_{\cdot,1:t-1}, \zeta_{k,v,1:t-1}) \propto \\
& \propto \mbox{exp} \left \{ - \frac{1}{2\sigma^2} \left( \beta_{k,v,t} - \beta_{k,v,t-1} \right)^2  \right \} \mbox{exp} \left \{ -\frac{1}{2 \sigma_{k,v,t-1}^2} \left(\beta_{k,v,t-1} - m_{k,v,t-1} \right)^2 \right \} \\
& \propto \mbox{exp} \left \{ - \frac{1}{2\sigma^2} \left( \beta_{k,v,t-1}^2 - 2 \beta_{k,v,t} \beta_{k,v,t-1} \right)  \right \} \mbox{exp} \left \{ -\frac{1}{2 \sigma_{k,v,t-1}^2} \left(\beta_{k,v,t-1}^2 -2  m_{k,v,t-1} \beta_{k,v,t-1} \right) \right \} \\
& \propto \mbox{exp} \left \{ -\frac{1}{2} \left(\frac{1}{\sigma^2} + \frac{1}{\sigma_{k,v,t-1}^2} \right) \beta_{k,v,t-1}^2 + \beta_{k,v,t-1} \left( \frac{\beta_{k,v,t}}{\sigma^2} + \frac{m_{k,v,t-1}}{\sigma_{k,v,t-1}^2} \right)  \right \}
\end{align*}
From the last line above, we can identify the parameters of a Gaussian distribution, so

\begin{itemize}
\item $\beta_{k,v,t-1} | \beta_{k,v,t}, \beta_{k,-v, 1:t-1}, W_{\cdot,1:t-1}, Z_{\cdot,1:t-1}, \zeta_{k,v,1:t-1} \sim N(\tilde{m}_{k,v,t-1}, \tilde{\sigma}_{k,v,t-1}^2)$
\item $\tilde{\sigma}_{k,v,t-1}^2 = \left( \frac{1}{\sigma^2} + \frac{1}{\sigma_{k,v,t-1}^2}  \right)^{-1}$
\item $\tilde{m}_{k,v,t-1} = \tilde{\sigma}_{k,v,t-1}^2 \left(\frac{\beta_{k,v,t}}{\sigma^2} + \frac{m_{k,v,t-1}}{\sigma_{k,v,t-1}^2} \right) $. 
\end{itemize}
To sample $\beta_{k,v,1:T} | \beta_{k,-v,1:T}, W_{\cdot,1:T}, Z_{\cdot,1:T}, \zeta_{k,v,1:T}$, just
\begin{enumerate}
\item sample $\beta_{k,v,T} | \beta_{k,-v,1:T}, W_{\cdot,1:T}, Z_{\cdot,1:T}, \zeta_{k,v,1:T}$ from the filtered distribution,
\item recursively sample $\beta_{k,v,t-1} | \beta_{k,v,t}, \beta_{k,-v,1:t-1}, W_{\cdot,1:t-1}, Z_{\cdot,1:t-1}, \zeta_{k,v,1:t-1}$
\end{enumerate}

\subsection{Full Conditional for $\zeta_{k,v,t}$}
\label{subsec:FC_Zeta}

The full conditional for $\zeta_{k,v,t}| \gamma_{k,v,t}$ follows from Theorem 1 of \citet{PSW2013}.  

\begin{align*}
\ell(\beta_{k,v,t} | \beta_{k,-v,t}, Z_{\cdot,t} ) \propto \mbox{exp} \left\{ \kappa_{k,v,t}^{\beta} \gamma_{k,v,t} \right\} \int_{0}^{\infty} \mbox{exp} \left\{ -\frac{\zeta_{k,v,t}}{2} \gamma_{k,v,t} \right\} \,d\zeta_{k,v,t}
\end{align*}

The integrand $ \mbox{exp} \left\{ -\frac{\zeta_{k,v,t}}{2} \gamma_{k,v,t} \right\}$ is the kernel of the joint density $p(\zeta_{k,v,t}, \gamma_{k,v,t})$.  Thus \begin{center} $p(\zeta_{k,v,t} | \gamma_{k,v,t}) = \frac{\mbox{exp} \left\{ -\frac{\zeta_{k,v,t}}{2} \gamma_{k,v,t} \right\}}{\int_0^{\infty}\mbox{exp} \left\{ -\frac{\zeta_{k,v,t}}{2} \gamma_{k,v,t} \right\}\,d \zeta_{k,v,t}  }$ \end{center} is a density in the Polya-Gamma family, and $\zeta_{k,v,t} | \gamma_{k,v,t} \sim PG(n_{k,t}^y, \gamma_{k,v,t})$.

\subsection{Full Conditional for $\eta_{k,t}$ Step}
\label{subsec:FC_Eta}

Find the full conditional for $\eta_{d,k,t} |Z_{\cdot,t}, \eta_{d,-k,t}, \omega_{d,k,t}, \alpha_{k,t}$.  Define $x_{d,k,t} := \sum_{n = 1}^{N_{d,t}} \mathbbm{1}_{\{z_{d,n,t} = k \} }$.
  
\begin{align*}
p(Z_{d,t} | \eta_{d,t}) & \propto \left( \frac{e^{\eta_{d,1,t}}} {\sum_{j = 1}^K e^{\eta_{d,j,t} } }  \right)^{x_{d,1,t}}  \ldots\left( \frac{e^{\eta_{d,k,t}}} {\sum_{j = 1}^K e^{\eta_{d,j,t} } }  \right)^{x_{d,k,t}} \ldots\left( \frac{e^{\eta_{d,K,t}}} {\sum_{j = 1}^K e^{\eta_{d,j,t} } }  \right)^{x_{d,K,t}}.  
\end{align*}

Again focus on a single arbitrary probability, $ \frac{ e^{\eta_{d,k,t}}  }{ \sum_{j=1}^K e^{\eta_{d,j,t}}  } $.  Using the same manipulation as in Section \ref{subsec:FC_Beta} gives

\begin{align*}
\frac{ e^{\eta_{d,k,t}}  }{ \sum_{j=1}^K e^{\eta_{d,j,t}}  } = \frac{ e^{\eta_{d,k,t}}  }{ e^{\eta_{d,k,t}}+ \sum_{j \neq k} e^{\eta_{d,j,t}}  } =  \frac{ \frac{e^{\eta_{d,k,t}}}{\sum_{j \neq k} e^{\eta_{d,j,t}}}  }{ \frac{e^{\eta_{d,k,t}}}{ \sum_{j \neq k} e^{\eta_{d,j,t}} } + \sum_{j \neq k} e^{\eta_{d,j,t}}  }.
\end{align*}

Define $\psi_{d,k,t}:= \eta_{d,k,t} - \log \sum_{j \neq k} e^{\eta_{d,j,t}}$. This allows us to re-parameterize the conditional distribution $p(Z_{d,t}|\eta_{d,t})$, so

\begin{align*}
p(Z_{d,t} | \eta_{d,k,t}, \eta_{d,-k,t} ) \propto \left( \frac{e^{\psi_{d,k,t}}}{1+ e^{\psi_{d,k,t}}} \right)^{x_{d,k,t}} \left( \frac{1}{1+e^{\psi_{d,k,t}}} \right)^{N_{d,t} - x_{d,k,t}}. 
\end{align*}
Define $\kappa_{d,k,t}^{\eta}:=x_{d,k,t} - \frac{1}{2} N_{d,t}$.  With Theorem 1 from \citet{PSW2013}, we can augment the conditional distribution $p(Z_{d,t} | \eta_{d,k,t}, \eta_{d,-k,t})$ with a Polya-Gamma random variate: 

\begin{align*}
p(Z_{d,t} | \eta_{d,k,t} , \eta_{d,-k,t}, \omega_{d,k,t}) &\propto \mbox{exp} \left\{\kappa_{d,k,t}^{\eta} \psi_{d,k,t} - \frac{\omega_{d,k,t}}{2} \psi_{d,k,t}^2 \right\} \\
&\propto  \mbox{exp} \left\{ \kappa_{d,k,t}^{\eta} \psi_{d,k,t} - \frac{\omega_{d,k,t}}{2} \left(\eta_{d,k,t} - \log \sum_{j\neq k} e^{\eta_{d,j,t}}\right)^2\right\}.
\end{align*} 
When conditioning on $\alpha_{k,t}$ and by independence of $\eta_{d,k,t} \indep \eta_{d,j,t}$ for $j \neq k$ specified in the model,  the prior $p(\eta_{d,k,t} | \eta_{d,-k,t}, \alpha_{k,t}) = p(\eta_{d,k,t} | \alpha_{k,t})$ and $\eta_{d,k,t} | \alpha_{k,t} \sim N(F_{d,k,t}' \alpha_{k,t}, a^2)$.  

The posterior computation follows:
\begin{align*}
&p(\eta_{d,k,t} | Z_{\cdot,t}, \eta_{d,-k,t}, \omega_{d,k,t}, \alpha_{k,t} ) \propto p(Z_{\cdot,t} | \eta_{d,k,t},\eta_{d,-k,t}, \omega_{d,k,t}) p(\eta_{d,k,t} | \alpha_{k,t}) \\
&\propto \mbox{exp} \left\{ \kappa_{d,k,t}^{\eta} \psi_{d,k,t} - \frac{\omega_{d,k,t}}{2} \psi_{d,k,t} ^2 \right\}  p(\eta_{d,k,t} | \alpha_{k,t} ) \\
&\propto \mbox{exp} \left\{ \kappa_{d,k,t}^{\eta} \eta_{d,k,t} - \frac{\omega_{d,k,t}}{2} \left( \eta_{d,k,t} - \log \sum_{j \neq k} e^{\eta_{d,j,t}}  \right)^2 - \frac{1}{2 a^2} \left(\eta_{d,k,t} - F_{d,k,t}'\alpha_{k,t} \right)^2  \right\}\\
&\propto \mbox{exp} \left\{ \kappa_{d,k,t}^{\eta} \eta_{d,k,t} - \frac{\omega_{d,k,t}}{2} \left( \eta_{d,k,t}^2 -2 \eta_{d,k,t} \log \sum_{j \neq k} e^{\eta_{d,j,t}}  \right) - \frac{1}{2 a^2} \left(\eta_{d,k,t}^2 -2 \eta_{d,k,t}  F_{d,k,t}'\alpha_{k,t} \right)  \right\}\\
&\propto \mbox{exp} \left\{ -\frac{1}{2} \left( \omega_{d,k,t} + \frac{1}{a^2}  \right) \eta_{d,k,t}^2 + \eta_{d,k,t} \left( \kappa_{d,k,t}^{\eta} +  \omega_{d,k,t} \log \sum_{j \neq k} e^{\eta_{d,j,t}} + \frac{ F_{d,k,t}'\alpha_{k,t}}{a^2} \right)  \right\} 
\end{align*}

From this last line, we can directly read off the mean and variance parameters of a univariate Gaussian distribution:
\begin{itemize}
\item $\eta_{d,k,t} | Z_{\cdot,t}, \eta_{d,-k,t}, \omega_{d,k,t}, \alpha_{k,t} \sim N(q_{d,k,t}, \lambda_{d,k,t}^2)$,
\item $\lambda_{d,k,t}^2 = \left(\omega_{d,k,t} + \frac{1}{a^2} \right)^{-1}$,
\item $q_{d,k,t} = \lambda_{d,k,t}^2 \left(\kappa_{d,k,t}^{\eta} +\omega_{d,k,t} \log \sum_{j\neq k} e^{\eta_{d,j,t} } + \frac{1}{a^2} F_{d,k,t}' \alpha_{k,t} \right)$. 
\end{itemize}

\subsection{Full Conditional for $\omega_{d,k,t}$ Step}
\label{subsec:FC_Omega}

The full conditional for $\omega_{d,k,t}|\psi_{d,k,t}$ is developed in near identical fashion to that of $\zeta_{k,v,t}$ in Section \ref{subsec:FC_Zeta}.  

\begin{align*}
p(Z_{d,t} | \eta_{d,k,t} , \eta_{d,-k,t}) &\propto \mbox{exp} \left\{\kappa_{d,k,t}^{\eta} \psi_{d,k,t}\right\} \int_{0}^{\infty} \mbox{exp}\left\{ - \frac{\omega_{d,k,t}}{2} \psi_{d,k,t}^2 \right\} \,d \omega_{d,k,t}
\end{align*}
The integrand $ \mbox{exp} \left\{ -\frac{\omega_{d,k,t}}{2} \psi_{d,k,t} \right\}$ is the kernel of the joint density $p(\omega_{d,k,t}, \psi_{d,k,t})$.  Thus 
\begin{align*} 
p(\omega_{d,k,t} | \psi_{k,v,t}) = \frac{\mbox{exp} \left\{ -\frac{\omega_{d,k,t}}{2} \psi_{d,k,t} \right\}}{\int_0^{\infty}\mbox{exp} \left\{ -\frac{\omega_{d,k,t}}{2} \psi_{d,k,t} \right\}\,d \omega_{d,k,t}  } 
\end{align*} 
is a density in the Polya-Gamma family, and $\omega_{d,k,t} | \psi_{d,k,t} \sim PG(N_{d,t}, \psi_{d,k,t})$.  

\subsection{Full Conditional for $\alpha_{k,t}$ Step}

The full conditional $\alpha_{k,t} | \eta_{1:D_T, k, t}$ is obtained from standard Forward Filtering Backward Sampling recursions. Suppose that $\alpha_{k,t-1} | \eta_{k,1:t-1} \sim N(m_{k,t-1}, C_{k,t-1} )$.  Then as defined in \citet{westharrison}, 

\begin{align*}
\alpha_{k,t} | \eta_{k,1:t} \sim N(m_{k,t}, C_{k,t} )
\end{align*}
where 
\begin{align*}
m_{k,t} & = a_{k,t} + A_{k,t} e_{k,t} \\
C_{k,t} & = R_{k,t} - A_{k,t} Q_{k,t} A_{k,t}' \\
A_{k,t} & = R_{k,t} F_{k,t} Q_{k,t}^{-1}\\
R_{k,t} & = \frac{1}{\delta} C_{k,t-1} \\
a_{k,t} & = G_{k,t} m_{k,t-1} \\
e_{k,t} & = \eta_{k,t} - f_{k,t} \\
f_{k,t} & = F_{k,t}' a_{k,t}.
\end{align*}

\subsection{Full Conditional for $z_{n,d,t}$ Step}
Conditionally on $\eta_{t}$ and $\beta_{t}$, posterior samples for $z_{n,d,t} = k | w_{n,d,t} = v, \beta_{k,v,t}, \eta_{d,k,t}$ can be drawn independently.  So
 
\begin{align*}
&Pr(w_{n,d,t} = v | z_{n,d,t} = k, \beta_t ) \propto \frac{e^{\beta_{k,v,t}}}{\sum_{j=1}^V e^{\beta_{k,j,t}} }\\
&Pr(z_{n,d,t} = k | \eta_{t} ) \propto e^{\eta_{d,k,t}} \\
&Pr(z_{n,d,t} = k | w_{n,d,t} = v, \beta_t, \eta_t )\propto \frac{e^{\beta_{k,v,t}}}{\sum_{j=1}^V e^{\beta_{k,j,t}} }  e^{\eta_{d,k,t} }.  
\end{align*}

\newpage
\section{Moments of Polya-Gamma Distribution}
\label{App:Appendix_B}

\subsection{Computing the Mean}
\label{subsec:PG_Approx_Mean}

Equation 6 of \citet{PSW2013} provides the Laplace Transform of a Polya-Gamma random variable, $\omega \sim PG(b,c)$.

\begin{align*}
E[e^{-\omega s} ] &= \frac{ \cosh^b\left ( \frac{c}{2} \right)}{\cosh^b\left(\sqrt{\frac{c/2 + s}{2} } \right) }
\end{align*}
It follows that $E[\omega ] = -\frac{d}{ds} E[ e^{-\omega s} ] \rvert_{s = 0}$.  To compute this expectation, we need to take the derivative $\frac{d}{ds}\left \{  \frac{1}{\cosh^b\left( \sqrt{\frac{c^2/2 + s}{2}}  \right) } \right \}$.  Taking the derivative requires applying the chain rule multiple times.  To assist with this, we define the auxiliary functions and compute their derivatives.

The core argument of the nested functions is the function $h(x)$, and  
\begin{align*}
h(x) = \sqrt{\frac{c^2/2 + x}{2} } \hspace{1cm} h'(x) = \frac{1}{4} \left( \frac{c^2+ x}{2} \right)^{-\frac{1}{2}}.
\end{align*}
We also need the derivative of the hyperbolic cosine:  
\begin{align*}
f(x) = \cosh(x) \hspace{1cm} f'(x) = \sinh(x).
\end{align*}
There is one more function wrapped around the hyperbolic cosine:  
\begin{align*}
g(x) = f(x)^{-b} \hspace{1cm} g'(x) = \frac{-b}{x^{b+1}}.
\end{align*}   
By the chain rule, $\frac{d}{ds} g(f(h(t))) = g'(f(h(s))) f'(h(s)) h'(s)$.  Computing each derivative and evaluating at $s=0$ yields the expression for the mean:
\begin{align*}
E[-\omega ] & = \frac{d}{ds} E[e^{-\omega s} ] \rvert_{s=0} \\
& = \frac{d}{ds} \left \{ \frac{\cosh^b \left( \frac{c}{2} \right)}{\cosh^b\left( \sqrt{\frac{c^2/2 + s}{2}}  \right) } \right \} \rvert_{s=0} \\ 
&= - \frac{b}{4}\frac{\cosh^b \left(\frac{c}{2} \right)}{\cosh^{b+1}\left(\sqrt{\frac{c^2/s + s}{2}} \right) } \sinh \left( \sqrt{\frac{c^2/2 + s}{2}} \right) \left( \frac{c^2/2 + s}{2} \right)^{-\frac{1}{2}} \rvert_{s=0}  \\
& = - \frac{b}{4}\frac{\cosh^b \left(\frac{c}{2} \right)}{\cosh^{b+1}\left( \frac{c}{2} \right) } \sinh \left( \frac{c}{2} \right) \left( \frac{c^2}{4} \right)^{-1\frac{1}{2}} \\
& = - \frac{2b}{4c}\frac{\sinh \left( \frac{c}{2} \right)}{\cosh \left( \frac{c}{2} \right) } \\
& =  -\frac{b}{2c} \tanh \left( \frac{c}{2} \right).
\end{align*} 
By negating, we find that $E[\omega] = \frac{b}{2c} \tanh \left( \frac{c}{2} \right)$ This corresponds to the result in \citet{PSW2013}.

\subsection{Computing the Variance}
The harder part is to find the second derivative for the variance calculation.  
Since $E[\omega^2 ] = \frac{d^2}{ds^2} E[ e^{-\omega s} ] \rvert_{s = 0}$, we will simply compute the derivative of the quantity found for the mean:

\begin{align*}
\frac{d^2}{ds^2} \left \{ \frac{\cosh^b \left( \frac{c}{2} \right)}{\cosh^b\left( \sqrt{\frac{c^2/2 + s}{2}}  \right) } \right \}  = - \frac{d}{ds}  \frac{b}{4}\frac{\cosh^b \left(\frac{c}{2} \right)}{\cosh^{b+1}\left(\sqrt{\frac{c^2/s + s}{2}} \right) } \sinh \left( \sqrt{\frac{c^2/2 + s}{2}} \right) \left( \frac{c^2/2 + s}{2} \right)^{-\frac{1}{2}}.
\end{align*}
Again, we define auxiliary functions to compute this derivative:  

\begin{align*}
F_1(s) &= \frac{1}{\cosh^{b+1} \left( \sqrt{\frac{c^2/2 + s}{2}} \right) } \\
F_2(s) &= \sinh \left(\sqrt{\frac{c^2/2 + s}{2}} \right) \\
F_3(s) &= \left(\frac{c^2/2 + s}{2} \right)^{-\frac{1}{2}}. 
\end{align*}
So we are interested in computing the derivative
\begin{align*} 
\frac{d}{ds}\left \{  -\frac{b}{4} \cosh^b\left(\frac{c}{2} \right) F_1(s) F_2(s) F_3(s) \right \}.
\end{align*}  
By the product rule, we need the quantity 
\begin{align*}
F_1'(s) F_2(s) F_3(s) + F_1(s) F_2'(s) F_3(s) + F_1(s) F_2(s) F_3'(s).
\end{align*}
That is, the second moment 
\begin{align*}
E[\omega^2 ] = -\frac{b}{4} \cosh^b\left( \frac{c}{2} \right) \left \{ F_1'(s) F_2(s) F_3(s) + F_1(s) F_2'(s) F_3(s) + F_1(s) F_2(s) F_3'(s) \right \}.
\end{align*}
We will compute each of these quantities separately.  

Begin by computing $F_1'(s)$.  We have already computed this derivative in Section \ref{subsec:PG_Approx_Mean} when finding the mean:  

\begin{align*}
F_1'(s) &= - \frac{b+1}{4} \frac{\sinh\left( \sqrt{\frac{c^2/2 + s}{2} } \right) }{\cosh^{b+2} \left( \sqrt{\frac{c^2/2 + s}{2} } \right) } \left( \sqrt{\frac{c^2/2 + s}{2} } \right)^{-\frac{1}{2}} \\ 
& = -\frac{b+1}{4} \sech \left( \sqrt{\frac{c^2/2 + s}{2} } \right) F_1(s) F_2(s) F_3(s).  
\end{align*}
As a result, $F_1'(s) F_2(s) F_3(s) = -\frac{b+1}{4} \sech \left( \sqrt{\frac{c^2/2 + s}{2} } \right) F_1(s) F_2(s)^2 F_3(s)^2$.

Next compute $F_2'(s)$:  
\begin{align*}
\frac{d}{ds} F_2(s) & = \frac{d}{ds} \sinh \left(\sqrt{\frac{c^2/2 + s}{2}} \right) \\
& = \frac{1}{4} \cosh \left(\sqrt{\frac{c^2/2 + s}{2}} \right) \left(\frac{c^2/2+s}{2} \right)^{-\frac{1}{2}} \\
& = \frac{1}{4} \cosh \left(\sqrt{\frac{c^2/2+s}{2} } \right) F_3(s).
\end{align*}  
As a result, $F_1(s) F_2'(s) F_3(s) =\frac{1}{4} \cosh\left(\sqrt{\frac{c^2/2 + s}{2}} \right) F_1(s) F_3(s)^2$.  

Next compute $F_3'(s)$:  
\begin{align*}
F_3'(s) &= \frac{d}{ds} \left( \frac{c^2/2 + s}{2}\right)^{-\frac{1}{2}} \\
& = -\frac{1}{4}  \left( \frac{c^2/2 + s}{2}\right)^{-\frac{3}{2}} \\
& = -\frac{1}{4}  F_3(s)^3
\end{align*} 
As a result, $F_1(s) F_2(s) F_3(s) = -\frac{1}{4} F_1(s) F_2(s) F_3(s)^3$.  
Putting together the various pieces: 

{\tiny
\begin{align*}
&\frac{d^2}{ds^2} E[e^{-\omega s} ]\\
& = - \frac{b}{4} \cosh^{b} \left(\frac{c}{2} \right )\left \{ -\frac{b+1}{4} \sech \left( \sqrt{ \frac{c^2/2 + s}{2} } \right) F_1(s) F_2(s)^2 F_3(s)^2 + \frac{1}{4} \cosh \left( \sqrt{ \frac{c^2/2 + s}{2} } \right) F_1(s) F_3(s)^2 - \frac{1}{4} F_1(s) F_2(s) F_3(s)^3  \right \} \\
& = - \frac{b}{4} \cosh^{b} \left(\frac{c}{2} \right ) F_1(s) F_3(s)^2 \left \{ -\frac{b+1}{4} \sech \left( \sqrt{ \frac{c^2/2 + s}{2} } \right) F_2(s)^2 + \frac{1}{4} \cosh \left( \sqrt{ \frac{c^2/2 + s}{2} } \right) -\frac{1}{4} F_2(s) F_3(s) \right \} \\
& = \frac{b}{16} \cosh^{b} \left(\frac{c}{2} \right ) F_1(s) F_3(s)^2 \left \{ (b+1) \sech \left( \sqrt{ \frac{c^2/2 + s}{2} } \right) F_2(s)^2 - \cosh \left( \sqrt{ \frac{c^2/2 + s}{2} } \right) + F_2(s) F_3(s)  \right\}.
\end{align*}
}
Now evaluate this expression at $s = 0$.  

{\tiny
\begin{align*}
E[\omega^2 ] & = \\
& = \frac{b}{16} \cosh^{b} \left(\frac{c}{2} \right ) \frac{\sech\left( \frac{c}{2} \right) }{\cosh^b \left( \frac{c}{2} \right) } \left(\frac{2}{c} \right)^2 \left \{ (b+1) \sech \left( \frac{c}{2} \right) \sinh^2 \left( \frac{c}{2} \right) - \cosh \left( \frac{c}{2} \right) +  \left(\frac{2}{c} \right) \sinh \left( \frac{c}{2} \right)  \right \} \\
& = \frac{4b}{16c^2} \sech\left( \frac{c}{2} \right) \left \{ (b+1) \sech \left( \frac{c}{2} \right) \sinh^2 \left( \frac{c}{2} \right) - \cosh \left( \frac{c}{2} \right) +  \left(\frac{2}{c} \right) \sinh \left( \frac{c}{2} \right)  \right \} \\
& = \frac{4b}{16c^2} \sech^2\left( \frac{c}{2} \right) \left \{ (b+1) \sinh^2 \left( \frac{c}{2} \right) - \cosh^2 \left( \frac{c}{2} \right) +  \left(\frac{2}{c} \right) \sinh \left( \frac{c}{2} \right) \cosh \left( \frac{c}{2} \right)  \right \}. \\
\end{align*}
}

The following identities will be useful:
\begin{align*}
\sinh^2\left( \frac{c}{2} \right) - \cosh^2 \left( \frac{c}{2} \right) = - 1 \hspace{1cm} \sinh\left( \frac{c}{2} \right) \cosh \left( \frac{c}{2} \right) = \frac{1}{2} \sinh(c).
\end{align*}
So
\begin{align*}
E[\omega^2] & = \\ 
& = \frac{4b}{16c^2} \sech^2\left( \frac{c}{2} \right) \left \{ (b \sinh^2 \left( \frac{c}{2} \right) - 1 +  \left(\frac{2}{c} \right)  \frac{1}{2} \sinh(c)  \right \} \\
& = \frac{b}{4c^3} \sech^2\left( \frac{c}{2} \right)  \left \{  \sinh(c) - c + bc \sinh^2 \left( \frac{c}{2} \right) \right \}. \\
\end{align*}

The variance follows immediately:
\begin{align*}
E[\omega^2] - E[\omega]^2 & = \\ 
& = \frac{b}{4c^3} \sech^2\left( \frac{c}{2} \right) \left \{  \sinh(c) - c \right \} + \frac{b^2}{4c^2} \frac{\sinh^2 \left( \frac{c}{2} \right)}{\cosh^2 \left( \frac{c}{2} \right) } - \left(\frac{b}{2c} \tanh\left( \frac{c}{2} \right)  \right)^2  \\
& = \frac{b}{4c^3} \sech^2\left( \frac{c}{2} \right) \left \{  \sinh(c) - c \right \}.
\end{align*}

\end{document}